\documentclass[11pt]{article}
\usepackage{fullpage}

\usepackage[utf8]{inputenc} 
\usepackage[T1]{fontenc}    
\usepackage{hyperref}       
\usepackage{url}            
\usepackage{booktabs}       
\usepackage{amsfonts}       
\usepackage{nicefrac}       
\usepackage{microtype}      
\usepackage{tcolorbox}
\usepackage{diagbox}

\usepackage{tabularx}
\usepackage{booktabs}

\usepackage{enumerate}
\usepackage[shortlabels]{enumitem}
\usepackage{algorithmic}
\usepackage{algorithm}
\usepackage{graphicx} 
\usepackage{subcaption}
\usepackage{amsmath}
\usepackage{amsthm}
\usepackage{amssymb}
\usepackage{tikz}
\usepackage{tablefootnote}
\usepackage{multirow}
\usepackage{enumerate}
\usepackage{color}
\usepackage{xcolor}
\usepackage{natbib}

\usetikzlibrary{arrows}

\allowdisplaybreaks[4]

\usepackage{mathrsfs}

\usepackage{hyperref}
\usepackage{bm,todonotes}

\allowdisplaybreaks

\newtheorem{theorem}{Theorem}[section]
\newtheorem{lemma}{Lemma}[section]

\newtheorem{asmp}{Assumption}[section]
\newtheorem{definition}{Definition}[section]

\newtheorem{remark}{Remark}[section]

\newtheorem{axiom}{Axiom}

\hypersetup{
	colorlinks=true,
	filecolor=blue,
	citecolor = blue,
	urlcolor=cyan,
}

\usepackage{caption, subcaption}
 
\title{
Uncertainty Quantification of Data Shapley via Statistical Inference
}

\author{
Mengmeng Wu\thanks{Equal contribution. Center for Intelligent Decision-Making and Machine Learning, School of Management, Xi'an Jiaotong University; email: \texttt{mengmengwu@stu.xjtu.edu.cn}. } \\
	\and
Zhihong Liu\thanks{Equal contribution. Center for Intelligent Decision-Making and Machine Learning, School of Management, Xi'an Jiaotong University; email: \texttt{lzh200829@stu.xjtu.edu.cn}. } \\
\and
Xiang Li\thanks{University of Pennsylvania; email:\texttt{lx10077@upenn.edu}. } \\
\and
Ruoxi Jia\thanks{Virginia Tech, Blacksburg; email: \texttt{ruoxijia@vt.edu}. } \\
\and
Xiangyu Chang \thanks{Corresponding author: Center for Intelligent Decision-Making and Machine Learning, School of Management, Xi’an Jiaotong University; email: \texttt{xiangyuchang@xjtu.edu.cn}. } 
}

\begin{document}

\maketitle

\begin{abstract}
As data plays an increasingly pivotal role in decision-making, the emergence of data markets underscores the growing importance of data valuation. 
Within the machine learning landscape, Data Shapley stands out as a widely embraced method for data valuation. 
However, a limitation of Data Shapley is its assumption of a fixed dataset, contrasting with the dynamic nature of real-world applications where data constantly evolves and expands. 
This paper establishes the relationship between Data Shapley and infinite-order U-statistics and addresses this limitation by quantifying the uncertainty of Data Shapley with changes in data distribution from the perspective of U-statistics. We make statistical inferences on data valuation to obtain confidence intervals for the estimations.
We construct two different algorithms to estimate this uncertainty and provide recommendations for their applicable situations. 
We also conduct a series of experiments on various datasets to verify asymptotic normality and propose a practical trading scenario enabled by this method.
\end{abstract} 

\section{Introduction}
\label{intro}

In today's data-driven landscape, the critical role of data in decision-making processes is unequivocal~\citep{barua2012measuring}. 
As data generation intensifies across multiple sectors, organizations increasingly depend on diverse datasets to shape their strategies and operations. 
This surge in demand has given rise to numerous data trading markets and platforms, exemplified by AWS Data Exchange and ICE Data Services, which facilitate the exchange of valuable data assets. 
The cornerstone of data trading lies in the valuation of data~\citep{pei2020survey}.

Our focus is on valuing the contribution of data to a machine learning model's performance rather than the influence of market factors on data pricing~\citep{2021ERATO}.
This valuation must consider two key data characteristics. 
First, data often require collaborative efforts to unleash its maximum potential, as seen in supervised learning where data are combined to train models, subsequently utilized for predictions and decision-making. 
The value of each data is intertwined with the other data trained alongside it. 
Second, data value depends on the model's output, which further depends on the underlying algorithm and model performance evaluation metrics. 
Therefore, in the context of supervised learning, data valuation is thus defined as the process of fairly assessing and quantifying each data point's contribution to model performance, given a fixed training dataset, a specific learning algorithm, and a desired performance evaluation metric~\citep{ghorbani2019,jia2019towards}. 

\begin{figure}[ht]
    \begin{center}
\centerline{\includegraphics[width=0.8\columnwidth]{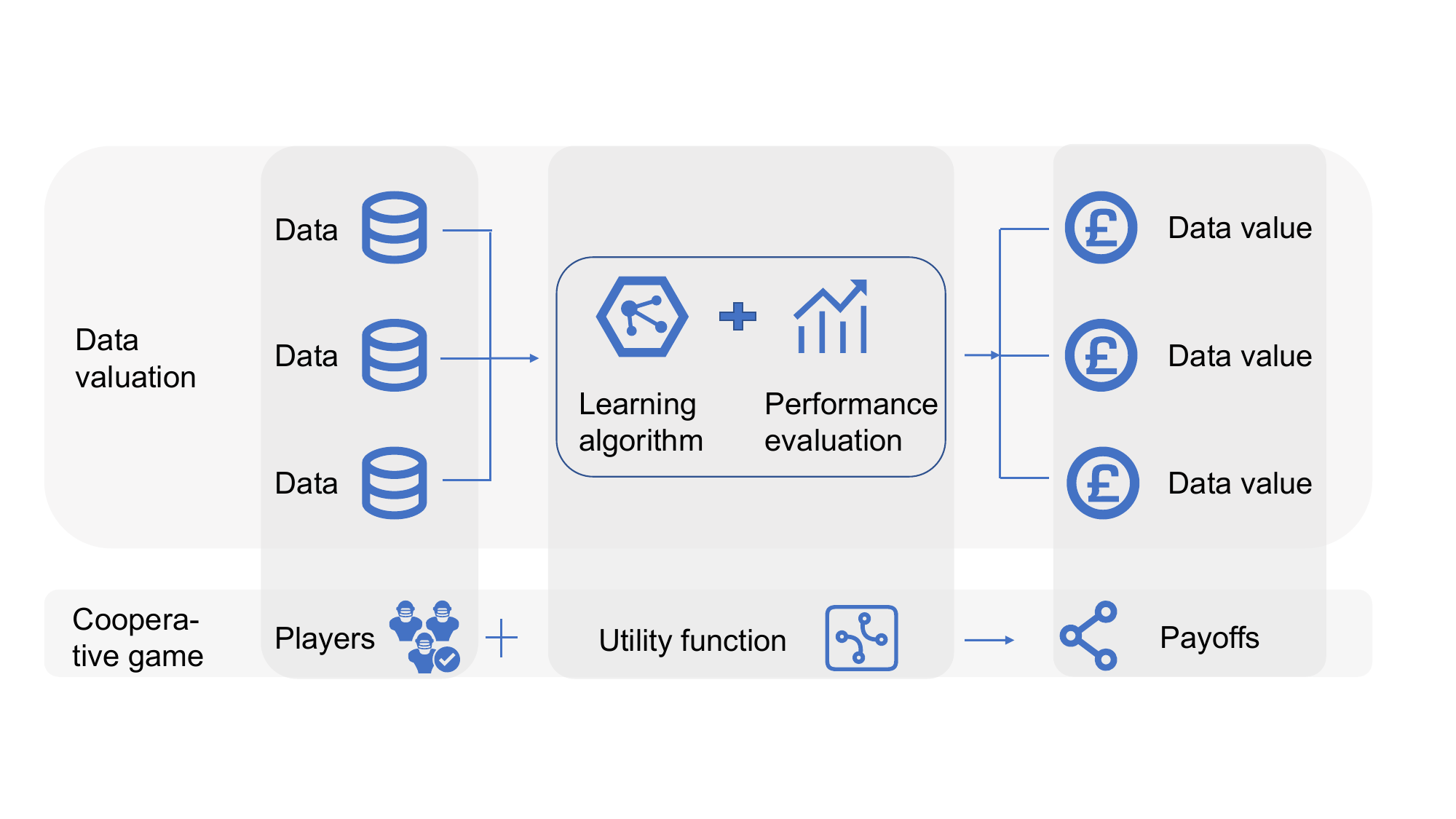}}
    \caption{This illustration represents the data valuation problem within the framework of cooperative game theory.}
\label{valuation&cooperative}
\end{center}
\vskip -0.2in
\end{figure}

The challenge of assigning fair values to data can be elegantly framed within the context of cooperative game theory. In this framework, each piece of training data is analogous to a player in a cooperative game, while the utility function represents the predictive performance of machine learning models trained on various combinations of data. The goal of assigning fair values to data mirrors the equitable distribution of payoffs to players in the game (see Figure \ref{valuation&cooperative}  for a visual representation of this analogy).
Drawing from various cooperative game theories, researchers have employed several methods for data valuation, such as the Banzhaf value~\citep{wang2023data}, leave-one-out error~\citep{ghorbani2019}, least core~\citep{yan2021if}, and the Shapley value~\citep{ghorbani2019,jia2019towards}. Among these, the Shapley value has gained widespread adoption for characterizing data value due to its \emph{unique} adherence to four desriable axioms with fairness interpretations~\citep{jia2019efficient}.

The valuation of data through the application of Shapley value is known as Data Shapley~\citep{ghorbani2019,jia2019towards}. Essentially, the Shapley value of a data point is calculated by considering its weighted marginal contributions to all subsets that do not contain it. The weights are the same among equally sized subsets, and the final value is derived by averaging across different subset sizes. 
However, Data Shapley exhibits several limitations in its practical application. 
Firstly, the computation of Data Shapley is intrinsically tied to specific datasets. Consequently, any alterations to the dataset---such as modifications or additions of individual data points due to real-time updates~\citep{hoi2021online} or partial data removal due to privacy concerns~\citep{xu2024machine}---necessitate recalculation of Data Shapley values for each data point.  
In practical applications, Monte Carlo approximation is the most commonly used estimation method. However, its asymptotic properties remain inconclusive.
Secondly, Data Shapley produces deterministic values based on a given dataset, failing to quantify the uncertainty arising from data variations~\citep{ghorbani2020distributional}. This characteristic makes it less suitable for high-risk scenarios that demand robustness, such as applications in medical and financial domains.


To address these challenges, a natural approach involves considering the inherent distribution of the data and quantifying the uncertainty of the contribution of a data point to the model's performance.
By recognizing the stochastic nature of data, we can treat the valuation of a specific data point as a random variable, transforming Data Shapley into a probabilistic framework.
Instead of computing its marginal contributions to all subsets of a fixed dataset, we consider the expectation of marginal contributions over subsets of different data samples.  Since computing the exact value of this expectation is infeasible, we focus on analyzing the statistical properties of its estimator.
As the most commonly used estimator, Monte Carlo,
if we can obtain its statistical properties, including confidence intervals---which essentially provide a measure of uncertainty or reliability regarding the contribution of the data point to model performance---it will significantly aid in determining data quality and making informed decisions.
Based on this premise, we formulate the following research questions:
\textit{(1) Does the Monte Carlo estimator of Data Shapley, exhibit specific asymptotic trends with continuous growth in dataset size? 
(2) Can we describe this trend from a statistical perspective to quantify uncertainty? 
(3) What problems can this quantification address in real-world data trading scenarios?}
To address the above issues, we demonstrate that the Monte Carlo estimator of Data Shapley can be considered as a summation of infinite-order U-statistics~\citep{frees1989infinite}.
Leveraging the statistical property of infinite-order U-statistics, we establish the asymptotic theory and confidence intervals of the Data Shapley estimator to assess the credibility of data valuation.

The remainder of the paper is organized as follows. In Section 2, we discuss related work and our contributions. Section 3 introduces the basic concepts of Data Shapley and U-statistics. Section 4 explores the understanding of Data Shapley from the perspective of U-statistics. 
Section 5 delves into the asymptotic normality of the estimator and provides a detailed description of the hypothesis testing process.
Section 6 proposes two estimation algorithms. 
Section 7 focuses on simulation studies and validates our theory through extensive experiments.
Finally, Section 8 summarizes our conclusions.

\section{Related Work and Contribution}

In recent years, the rise of data markets has heightened the focus on data pricing.  Since the marginal cost of data replication is approaching zero, traditional pricing strategies are no longer applicable to data~\citep{pei2020survey}.
The challenge of how to reasonably value and price data has become pivotal in the era of artificial intelligence, prompting extensive research by scholars~\citep{li2014pricing, 2019Towards, 2019Revenue,ghorbani2019, jia2019towards, 2021ERATO, 2020Online}.
Within the field of machine learning, data valuation is a critical problem, addressing the fair evaluation of each datum's impact on model performance. A solution to this problem has many applications, including data quality assessment~\citep{tang2021data,Wang2024AnES} and enhancing model interpretability~\citep{koh2017understanding}.

The Shapley value, originating from cooperative game theory~\citep{1953A}, is widely employed in assessing the contribution of data to model performance~\citep{ghorbani2019, jia2019towards, jia2019efficient}. Various estimation methods have been explored to improve its computational efficiency~\citep{ghorbani2019, jia2019towards, jia2019efficient, kwon2022beta, tang2021data}. 
A significant drawback of utilizing Data Shapley is its assumption of a static dataset, which yields deterministic values incapable of capturing the uncertainty arising from data changes.

Data value uncertainty can arise from stochastic data valuation estimation algorithms and data variations. 
Wu et al.~\citep{wu2022variance} introduced a stratified sampling algorithm to mitigate the uncertainty of data values arising from algorithmic uncertainty. 
The distributional Shapley value proposed by Ghorbani et al.\citep{ghorbani2020distributional} offers increased robustness in handling data variability.
Although our research also investigates uncertainty stemming from data changes, our objectives diverge from Ghorbani et al.'s work.
We are interested in studying how the estimated value of Data Shapley changes when other data or the data volume involved in model training changes. 
We have obtained the limiting variance of this estimated value as the amount of data increases. 
This variance provides an estimate of the computational complexity in calculating a data point's value, with larger variances indicating a need for more extensive sampling.
Additionally, we obtain confidence intervals for the estimated values.
These intervals serve two crucial purposes:
Firstly, they enable data providers to anticipate value changes in specific data points when other data is replaced with identically distributed data.
Secondly, they assist data buyers in evaluating the reasonableness of sellers' data pricing in data trading scenarios.

When examining the variation of estimated Data Shapley as data volume increases, we employ classical U-statistic theory introduced by Wassily Hoeffding, which is a crucial class of statistics used in non-parametric estimation~\citep{1948A}. 
Consider the U-statistic $U_n := \binom{n}{s}^{-1} \sum_{1 \leq i_1 < i_2 < \dots < i_s \leq n} h(X_{i_1},\dots,X_{i_s})$, where $\sum$ represents the summation over the $\binom {n} {s}$ combinations of $s$ distinct elements $\{i_1,\dots,i_s\}$ from $\{1,\dots,n\}$ and $h$ is a symmetric kernel of order $s$. 
Traditionally, U-statistics are applied in cases where the kernel $h$ and order $s$ are fixed and do not depend on the sample size $n$~\citep{1948A}. 
Subsequently, the class of U-statistics is expanded to include infinite-order U-statistics, where both $h$ and $s$ tend towards infinity as $n$ approaches infinity~\citep{frees1989infinite}. 
The study of inference property of random forest~\citep{mentch2016quantifying} provides sufficient conditions for the asymptotic normality of infinite-order U-statistics in advance. 
Then, \citep{song2019approximating} proposes alternative sufficient conditions based on high-order moment assumptions, while \citep{diciccio2022clt} presents a general triangular array central limit theorem for infinite-order U-statistics. 
In contrast to these studies, which focus on sufficient conditions for the asymptotic normality of U-statistics, we consider data valuation problems where the Monte Carlo estimator of Data Shapley can be viewed as the average of $n$ U-statistics from $s=0$ to $s=n-1$. 
We justify its asymptotic normality and provide statistical inferences for the estimator of Data Shapley.

Our approach allows for a more nuanced understanding of data value in dynamic environments, potentially extending the applicability of Data Shapley to scenarios where data uncertainty is a critical factor. By bridging the gap between static valuation methods and the dynamic nature of real-world data, our research contributes to the development of more adaptable and trustworthy data valuation frameworks.

The technical contributions of this paper are summarized as follows.
\begin{itemize}
\item We establish a novel connection between Data Shapley and infinite-order U-statistics. 
This connection offers a fresh perspective on understanding the uncertainty introduced by data distribution to Data Shapley, allowing us to study the asymptotic properties of Data Shapley through the lens of infinite-order U-statistics.

\item  We establish the asymptotic normality of the Data Shapley estimator and construct confidence intervals, providing tractable formulas to quantify uncertainty in data valuation. 
This sets our work apart from that of Kwon and Zou~\citep{kwon2022beta}, who examine the asymptotic convergence of variance without establishing the asymptotic normality of the Data Shapley estimator. Additionally, our study differs from previous research on the distributional Shapley value~\citep{ghorbani2020distributional}, which focuses on enhancing the robustness of data valuation against dataset variations. 
In contrast, we aim to quantify the uncertainty introduced by data distribution through statistical inference.
\item 
We propose two algorithms to estimate the asymptotic variance of the Data Shapley estimator. 
The uncertainty of Data Shapley can be quantified through this variance. 
The two algorithms, called Double Monte Carlo and Pick-and-Freeze, each have distinct advantages. 
On the one hand, the Double Monte Carlo algorithm can achieve higher accuracy than Pick-and-Freeze by increasing the number of samples, leading to higher computational costs. 
On the other hand, the Pick-and-Freeze algorithm reduces computation costs by elaborately designing the Monte Carlo processing but has less accuracy than the Double Monte Carlo.
Additionally, we conduct a series of comprehensive experiments to validate the asymptotic normality of the Data Shapley estimator and present a significant case study to demonstrate its practical applicability.
\end{itemize}





\section{Preliminaries}\label{sec:pre}

This section presents the notation used throughout this paper. 
Subsequently, we provide an overview of Data Shapley and fundamental concepts related to U-statistics.

Define $\mathcal{Z} = \mathcal{X} \times \mathcal{Y}$, where $\mathcal{X} \subseteq \mathbb{R}^d$ denotes the input space, and $\mathcal{Y}$ denotes discrete or continuous output space.
Let $\mathcal{D}$ represent a data distribution supported on $\mathcal{Z}$,
and 
$\mathcal{Z}^*$ represents the power set of $\mathcal{Z}$.
Define the function $V: \mathcal{Z}^* \rightarrow [0,1]$, representing a performance metric or a utility function.
$V$ maps any data set $B \subseteq \mathcal{Z}^*$ to the corresponding performance of a model trained on $B$.
For example, in classification problems, $V(B)$ typically represents the prediction accuracy of a model trained on $B$.
Note that the utility function depends on the learning algorithm $\mathcal{A}$. Here, we omit this dependency for notational simplicity.
Define $n$ as a positive integer, and let $B \sim \mathcal{D}^n$ represent a collection of $n$ data sampled i.i.d. from $\mathcal{D}$.
In this context, $|B|$ denotes the cardinality of $B$, $[n] = \{1,2,\dots,n\}$, and $k \sim [n]$ represents a uniformly random sample from $[n]$.

Our aim is to quantify the contribution of individual data to the model's performance. 
Training a model with a dataset is analogous to collaborative teamwork, where each data point functions as a participant, with the collective goal of achieving high prediction accuracy.
The prediction accuracy serves as the measure of the team's effectiveness.
This collaborative aspect mirrors participants working together to complete a task.
Given this cooperative framework, previous studies have approached this problem through the lens of cooperative game theory. The Shapley value, in particular, has gained favor for data valuation due to its adherence to various desirable axioms~\citep{ jia2019towards,ghorbani2019}.
The application of the Shapley value to data valuation is referred to as ``Data Shapley.''


\subsection{Data Shapley}


\begin{definition}[Data Shapley]
Given a utility function $V$, a dataset $B \subseteq \mathcal{Z}^*$, where $|B|=n$, then the Data Shapley of a data point $z \in B$ is defined as
\begin{equation}\label{Shapley}
    \nu(z;V,B):=\frac{1}{n}\sum_{S\subseteq B_z,|S|=s}\frac{1}{\tbinom{n-1}{s}}[V(S\cup \{z\})-V(S)],
\end{equation}
where $B_z=B\backslash\{z\}$.
\end{definition}

The Data Shapley is a weighted average of the marginal contributions of the data to all subsets that do not include it.
The weighting is such that each possible cardinality is weighted equally, and subsets with the same cardinality have the same weight.
What sets the Data Shapley apart is its unique status as the only solution that satisfies a set of desirable axioms~\citep{1953A}. These axioms, detailed as follows, which include properties such as efficiency, symmetry, linearity, and null player, provide a strong theoretical foundation for fair data valuation~\citep{ghorbani2019,jia2019towards}.

\begin{axiom}{(Efficiency Axiom)}
$\sum_{z\in B}\nu(z;V,B)=V(B)$.  
\end{axiom}

\begin{axiom}{(Symmetry Axiom)}
For data $z_i$ and $z_j$, if $V(S\cup \{z_i\})=V(S\cup \{z_j\})$ holds for all $S$, where $S\subseteq B$ and $z_i,z_j\notin S$, then $\nu(z_i;V,B)=\nu(z_j;V,B)$.  
\end{axiom}

\begin{axiom}{(Dummy Axiom)}
 If $V(S\cup \{z\})=V(S)$ holds for all $S$, where $S\subset B$ and $z\notin S$, then $\nu(z;V,B)=0$.
\end{axiom}

\begin{axiom}{(Additivity Axiom)}
 For any pair of games $(B,V)$ and $(B,W)$, if $(V+W)(S)=V(S)+W(S)$ holds for all $S$, then $\nu(z;V+W,B)=\nu(z;V,B)+\nu(z;W,B)$. 
\end{axiom}



The efficiency axiom suggests that the utility of the entire dataset should be distributed among the data points, ensuring comprehensive value allocation. The symmetry axiom guarantees equal value for data points with the same marginal contributions to all coalitions excluding them, promoting fairness in valuation. The dummy axiom states that a data point with a marginal contribution of zero to all sets without it should have a value of zero. The additivity axiom enables the decomposition of a utility function into a sum of independent utility functions.

Despite its conceptual soundness, the practical use of Data Shapley faces computational challenges.
The process involves computing the marginal contributions of a data point to all $2^{n-1}$ subsets that do not contain it. Moreover, each marginal contribution requires training the model twice, resulting in a computational complexity of $\mathcal{O}(2^n)$.

In practical applications, especially when dealing with large datasets, Monte Carlo approximation is often employed for estimation.
Monte Carlo is a stochastic simulation method that derives approximate solutions to problems through statistical simulation or sampling. 
The Monte Carlo approximation formula for Data Shapley, denoted as $\hat{\nu}(z;V,B)$, is given by:
\begin{equation}
\hat{\nu}(z;V,B):=\frac{1}{m}\sum_{i=1}^m[V(S_i\cup {z})-V(S_i)],
\label{mc}
\end{equation}
where $m$ is the number of sampled subsets of $B\setminus \{z\}$.
This approach utilizes the average of the marginal contributions to 
$m$ randomly sampled subsets to approximate Data Shapley.

\subsection{U-statistics}

Revisiting the concept of U-statistics is valuable for understanding the statistical properties of Data Shapley. 


Let $X_1,\dots,X_n$ be i.i.d. from an population $\mathcal{P}$. 
In a broad class of problems, the parameters to be estimated are of the form:
\begin{equation*}
    \theta=\mathbb{E}_\mathcal{P}[h(X_1,\dots,X_s)]
\end{equation*}
with a positive integer $s$ and a symmetric Borel function $h$ satisfying
$\mathbb{E}_\mathcal{P}|h(X_1,\dots,X_s)| < \infty$.
The symmetry of $h$ indicates that the output of $h$ depends solely on the input variables, not their order.

A symmetric unbiased estimator for $\theta$ is given by
\begin{equation}
    U_n:=\binom {n} {s}^{-1}\sum_{1\leq i_1< i_2 < \dots i_s \leq n} h(X_{i_1},\dots,X_{i_s}),
    \label{U_statistic}
\end{equation}
where $\sum$ represents the summation over the $\binom {n} {s}$ combinations of $s$ distinct elements $\{i_1,\dots,i_s\}$ from $\{1,\dots,n\}$. 
We can now formally define U-statistics as follows.

\begin{definition}[U-statistics~\citep{1948A}]~\label{def:u-stats}
    The statistic $U_n$ in Eq.(\ref{U_statistic}) is called a U-statistic with kernel $h$ of order $s$.
\end{definition}

Initially applied to cases where both $s$ and $h$ are independent of $n$, the concept of U-statistics was later expanded to situations where both $h$ and $s$ depend on $n$, addressing applications in spatial statistics~\citep{1983Central,frees1989infinite}.

\begin{definition}[Infinite-Order U-statistics]\label{def:ious}
When $s \rightarrow \infty$ as $n \rightarrow \infty$,
    we call $U_n$ an infinite-order U-statistic (IOUS).
\end{definition}

A notable challenge in calculating $U_n$ accurately arises from the need to sample $\binom{n}{s}$ times. 
For large $n$, this incurs significant computational costs.
An effective approximation, termed \textit{Incomplete Infinite Order U-statistics}~\citep{1984The}, involves sampling a subset of $\binom{n}{s}$-tuples of indices and averaging them.


\begin{definition}[Incomplete Infinite-Order U-statistics]\label{def:iious}
We call
   \begin{equation}
\hat{U}_n:=\frac{1}{m}\sum_{j=1}^m   h(X_{i_1^j},\dots,X_{i_s^j}),
\end{equation}
an incomplete IOUS, where $\sum$ represents the summation over the $m$ randomly sampled combinations of $s$ distinct elements $\{i_1^j,\dots,i_s^j\}$ from $\{1,\dots,n\}$, with $m \le \binom{n}{s}$ and 
$s \rightarrow \infty$ as $n \rightarrow \infty$.
\end{definition}

The following Section \ref{sec:Usta} will show the connection between the Data Shapley and IOUS, then Section \ref{sec:asy} will utilize the property of incomplete IOUS to justify the asymptotic behavior of the estimator of Data Shapley.

\section{From Data Shapley to Infinite-Order U-statistics}
~\label{sec:Usta}


This section aims to explicitly demonstrate the connection between Data Shapley and IOUS.
To this end, we reemphasize that $Z_i\in B, i\in[n]$ are i.i.d. sampled from distribution $\mathcal{D}$. 
Thus, we could consider $B\sim \mathcal{D}^n$ and $B_z=B\backslash\{z\}\sim\mathcal{D}^{n-1}$.
For a fixed cardinality $s$, we define
\begin{equation} \nu_s(z;B):=\frac{1}{\tbinom{n-1}{s}}\sum_{S\subseteq B_z,|S|=s}[V(S \cup \{z\})-V(S)],
\label{nu_s}
\end{equation}
then the Data Shapley $\nu(z;B)=\frac{1}{n}\sum_{s=0}^{n-1}\nu_s(z;B)$.
For simplicity, we denote the marginal contribution of $z$ to a set $S$ with cardinality $s$ as $h_s(S;z) := V(S\cup {z}) - V(S)$.
Thus,
\begin{equation}
    \nu_s(z;B)=\frac{1}{\tbinom{n-1}{s}}\sum_{S\subseteq B_z,|S|=s}h_s(S;z).
\end{equation}

Notably, $h_s(S;z)$ should be a random variable whose randomness stems from the distribution of $S$, as $S=\{Z_{i_1},\dots,Z_{i_s}\}$ is a subset of $B_z$, i.e., $S\sim \mathcal{D}^s$.
From this perspective, $\nu_s(z;B)$ is an unbiased estimate of $\mathbb{E}_{\mathcal{D}^s}[h_s(S;z)]$.
Therefore, ${\nu}_s(z;B)$ is the U-statistic of $\mathbb{E}_{\mathcal{D}^s}[h_s(S;z)]$ with kernel $h_s$ of order $s$ based on the Definition \ref{def:u-stats}.

Moreover, for computing Data Shapley in Eq.\eqref{Shapley}, $s$ ranges from $0$ to $n-1$, implying $s \rightarrow \infty$ as $n \rightarrow \infty$.
Thus, ${\nu}_s(z;B)$ is an IOUS of $\mathbb{E}_{\mathcal{D}^s}[h_s(S;z)]$ according to the Definition \ref{def:ious}.

\begin{remark}
Let us denote:
\begin{equation}
    \nu(z;n):=\frac{1}{n}\sum_{s=0}^{n-1}\mathbb{E}_{\mathcal{D}^{s}}[\nu_s(z;B)].
\end{equation}
This expression aligns with the distributional Shapley value proposed by~\citep{ghorbani2020distributional}, primarily introduced for two key considerations. 
Firstly, for Data Shapley, any changes, additions, or deletions to the dataset necessitate recalculation of all data values, incurring significant computational costs.
Secondly, computing the expected values of Data Shapley across diverse datasets reduces variance, enhancing the stability of the values.
However, our focus extends beyond this. We aim to analyze the asymptotic behavior of the infinite-order U-statistic in data valuation, providing a credible valuation interval as data volume $n$ approaches infinity.
\end{remark}

It is crucial to note that accurately calculating ${\nu}_s(z;B)$ requires enumerating all $\binom{n-1}{s}$ subsets. 
In practical applications, especially with large $n$, this incurs substantial computational costs. 
An effective surrogate is to approximate it by a finite-sample average, i.e., to sample $m_s$ batches and compute:
\begin{equation}
\hat{\nu}_s(z;n,m_s):=\frac{1}{m_s}\sum_{j=1}^{m_s} h_s(Z_{i_1^j},\dots,Z_{i_s^j};z)
\label{apo},
\end{equation}
where $\sum$ represents the summation over the $m_s$ combinations of $s$ distinct elements 
sampled from $\{1,\dots,n\}$.  
This approximation represents an incomplete IOUS of $\mathbb{E}_{\mathcal{D}^s}[h_s(S;z)]$ according to Definition \ref{def:iious}.
This incomplete IOUS representation can be further viewed as a Monte Carlo approximation of $\nu_s(z;B)$ (see Eq.\eqref{mc} and Eq.\eqref{nu_s}).
To show a more distinct road map between the U-statistics and Data Shapley, we present a summary of symbols in Table \ref{symbols}.


\begin{table*}[t]
    \caption{Summary of Symbols Corresponding to Data Shapley and U-statistics}
    \label{symbols}
    \centering
    \renewcommand{\arraystretch}{1.5}
    \begin{tabularx}{\textwidth}{p{3.3cm}p{7cm}p{7cm}} 
        \toprule
        Notations & U-statistic & Data Shapley \\
        \midrule
        Aim & $\theta = \mathbb{E}_\mathcal{P}[h(X_1, \dots, X_s)]$ & $\mathbb{E}_{\mathcal{D}^s}[h_s(Z_{i_1}, \dots, Z_{i_s}; z)]$ \\
        Statistic & \scriptsize $U_n = \binom{n}{s}^{-1} \sum_{1 \leq i_1 < i_2 < \dots < i_s \leq n} h(X_{i_1}, \dots, X_{i_s})$ & \scriptsize $\nu_s(z; B) = \binom{n-1}{s}^{-1} \sum_{S \subseteq B_z, |S| = s} h_s(S; z)$ \\
        Kernel & $h(X_{i_1}, \dots, X_{i_s})$ & $h_s(Z_{i_1}, \dots, Z_{i_s}; z)$\\
        Order & $s$ & $s$ \\
        Infinite Order & $s \rightarrow \infty$ & $s \rightarrow \infty$ \\
        Incomplete Infinite Order & $\hat{U}_n$ & $\hat{\nu}_s(z; n, m_s)$ \\
        \bottomrule
    \end{tabularx}
\end{table*}

\section{Asymptotic normality and  statistical inference of Data Shapley}~\label{sec:asy}

{We initiate our analysis by establishing the asymptotic normality of $\hat{\nu}_s(z;n,m_s)$ and subsequently applying it to derive the asymptotic normality of the aggregated value $\hat{\nu}(z;n,m)=\frac{1}{n}\sum_{s=0}^{n-1}\hat{\nu}_s(z;n,m_s)$, where $\sum_{s=0}^{n-1}m_s=m$.
The derivation uses the notion of deletion stability and relies on a small deletion stability assumption (i.e., Assumption~\ref{assump}).
Before presenting the main results, we first introduce its formal definition.}


\subsection{Deletion stability}
\begin{definition}
For the utility function $V$, and non-increasing $\beta: \mathbb{N}\rightarrow [0,1]$, $V$ is $\beta(s)$-deletion stable 
if for all $s \in \mathbb{N}$ and $S \in \mathcal{Z}^{s}$ and for all $z\in \mathcal{Z}$,
 \begin{equation}
 \left|V(S\cup z)-V(S)\right|\leq\beta(s).
    \end{equation}  
\end{definition}
This concept of deletion stability describes the relationship between the change in the corresponding utility function when a single data point is removed and the amount of data.
We introduce the following assumption.
\begin{asmp}
\label{assump}
    Let $h_s(S;z) := V(S\cup {z}) - V(S)$, we assume that $h_s$ is $\beta(s)$-deletion stable, and further, $\beta(s)= \mathcal{O} (\frac{1}{s+1})$.
\end{asmp}

\begin{remark}
This assumption has been widely utilized in the data valuation literature \citep{ghorbani2020distributional,wu2022variance} to facilitate theoretical analysis.
In machine learning applications, the utility function of a set $S$ is often defined as the loss of the model trained over $S$ for predicting a test point $z^ \prime$, i.e., $V(S) = l(\mathcal{A}(S),z ^ \prime)$, where  $\mathcal{A}$ is the underlying learning algorithm mapping a dataset to a model.
Hence, $\left|V(S\cup z)-V(S)\right|=\left|l(\mathcal{A}(S\cup z),z ^ \prime)-l(\mathcal{A}(S),z ^ \prime)\right|$.
Prior work~\citep{bousquet2002stability,Shalev2010Learnability} has defined that an algorithm $\mathcal{A}$ has the $\beta'$-uniform stability with respect to the loss function $f$ if $\forall S$ and $\forall z$,  $\left|l(\mathcal{A}(S\cup z),.)-l(\mathcal{A}(S),.)\right| \leq \beta'$.
Obviously, if 
algorithm $\mathcal{A}$ has $\beta'$-uniform stability,
$h_s(S;z)$ is $\beta=\beta'/2$-deletion stable.
\citep{bousquet2002stability,Shalev2010Learnability} have justified that a number of machine learning algorithms satisfy the uniform stability with $\beta'=\mathcal{O}(1/(1+s))$.
These results further support the rationality of Assumption \ref{assump}. 
\end{remark}

\subsection{Asymptotic normality}
To explore asymptotic normality, we must first define the variance term. 
Let 
\begin{align}\label{eq:vszn}
    h_{k,s}(Z_1,\cdots,Z_k;z) 
    :=& \mathbb{E}\big(h_s(Z_1,\cdots,Z_s;z \mid Z_1,\cdots,Z_k)\big) - \nu_s(z;n),
\end{align}
for $1\leq k\leq s$ and $\nu_s(z;n):=\mathbb{E}_{\mathcal{D}^s}[h_s(S;z)]$. 
Eq.\eqref{eq:vszn} denotes the conditional expectation of marginal contribution given $k$ variables when the cardinality of the set is $s$. 
It is obvious that \begin{equation}
        \mathbb{E}(h_{k,s}(Z_1,\cdots,Z_k;z))=0.
    \end{equation}
We define by
    \begin{equation}
        \zeta_{k,s}(z):=\mathrm{Var}(h_{k,s}(Z_1,\cdots,Z_k;z))
        \label{zeta}
    \end{equation}
its variance. $\zeta_{k,s}(z)$ plays a crucial role in analyzing asymptotic normality, especially $\zeta_{1,s}(z)$.
\citep{Lee} shows that the variance of $\sqrt{n}U_n$ tends
to $s^2\zeta_{1,s}(z)$, which has provided significant inspiration for studying asymptotic normality.

\begin{theorem}
    \label{thm:layer_asymp}
     Let $Z_1,Z_2,\dots \overset{i.i.d}{\sim}\mathcal{D}$,
    and $\hat{\nu}_s(z;n,m_s)=\frac{1}{m_s}\sum_{1\leq i_1< i_2 < \dots i_s \leq n} h_s(Z_{i_1},\dots,Z_{i_s};z)$ be an incomplete 
 IOUS with kernel $h_s$ that satisfies assumption \ref{assump}.
    Let $\nu_s(z;n)=\mathbb{E}h_s(Z_{i_1},Z_{i_2},\dots,Z_{i_s};z)$ such that $\mathbb{E}h_s^2 < \infty$ and let $\lim\limits_{n\rightarrow\infty}{n/m_s}=0$ then for $s\sim n^{\gamma}, \gamma\in(0,1/2)$ and 
    $\zeta_{1,s}(z)> C \neq 0$,
  we have
\begin{equation}
        \frac{\sqrt{n-1}\left( \hat{\nu}_s(z;n,m_s)-\nu_s(z;n)\right)}{\sqrt{s^2\zeta_{1,s}(z)}}\stackrel{d}{\rightarrow}\mathcal{N}(0,1).
    \end{equation}
\end{theorem}

Theorem~\ref{thm:layer_asymp} states that when $s$, the cardinality of the set, is fixed, as long as the number of samples $m_s$ is sufficiently large compared to $n$, the estimated value $\hat{\nu}_s(z;n,m_s)$ follows a limit normal distribution with a mean of $\nu_s(z;n)$ and a variance of $\frac{s^2\zeta_{1,s}(z)}{n-1}$.
The proof process of the theorem is shown in~\ref{appx51}.

Two important points should be emphasized. Firstly, it is essential that $s$ remains fixed and $\frac{s}{\sqrt{n}}$ tends to 0 as $n$ approaches infinity. This requirement ensures the convergence of the variance. Secondly, we assume that the variance $\zeta_{1,s}(z)$ is greater than a non-zero constant. 
This assumption is justified as the variance of any unbiased estimator must be greater than or equal to the Cramér-Rao bound~\citep{1989MUSIC}.

\begin{remark}
The results presented here are partially supported by \citep{kwon2022beta}. Let $\zeta_s(z):=\mathrm{Var}(h_s(S;z))$. \citep{kwon2022beta} prove that when $s=\mathcal{O}(n^{1/2})$, and assuming $\lim\limits_{s\rightarrow \infty} \zeta_s(z)/(s\zeta_{1,s}(z))$ is bounded, then $(s^2\zeta_{1,s}(z)/n)^{-1}\mathrm{Var}(\nu_s(z;n))\rightarrow 1$.
We expand their conclusion in two aspects. 
Firstly, we prove the variance of incomplete IOUS $\hat{\nu}_s(z;n,m_s)$, which is different from their complete U-statistics. 
Secondly, we further demonstrate that it follows an asymptotic normal distribution rather than just the limit form of variance.
\end{remark}

The Monte Carlo estimate of Data Shapley is the average of $\hat{\nu}_s(z;n,m_s)$ for different cardinals $s$. We further derive the asymptotic properties of the Monte Carlo estimation of Data Shapley based on Theorem~\ref{thm:asy}.

\begin{theorem}
Suppose the conditions in Theorem~\ref{thm:layer_asymp} hold
and $\gamma \in (1/3,1/2)$, let $\hat{\nu}(z;n,n^\gamma,m)=\frac{1}{n^\gamma}\sum_{s=0}^{n^\gamma-1} \hat{\nu}_s(z;n,m_s)$, where 
$m=\sum_{i=0}^{n^\gamma-1}m_s$,
then 
\label{thm:asy}
        \begin{equation} 
        \frac{\sqrt{n-1}(\hat{\nu}(z;n,n^\gamma,m)- \nu(z;n))}{\sqrt{\frac{1}{(n^\gamma)^2}\sum_{s=1}^{n^\gamma-1}s^2\zeta_{1,s}(z)}}\stackrel{d}{\rightarrow}\mathcal{N}(0,1).
        \label{limiting distribution}
\end{equation}
\end{theorem}
{Theorem~\ref{thm:asy} introduces a new parameter $\gamma$, which characterizes the conditions under which the estimated $\hat{\nu}(z;n,n^\gamma,m)$ satisfies asymptotic normality. Specifically, this occurs when the maximum value of $s$ is at least $n^\gamma$. As $n$ approaches infinity, $s$ also tends towards infinity, with its range originally spanning from 0 to $n$. The value $n^\gamma$ effectively divides the range of $s$ into two non-overlapping intervals: $(0, n^\gamma)$ and $(n^\gamma, \infty)$. The approximation on $\sum_{s=0}^{n^\gamma}$ satisfies asymptotic normality, while the approximation on $\sum_{s=n^\gamma}^{\infty}$ converges to 0.
Our proof process reflects this division. For the interval $(0, n^\gamma)$, we first establish the satisfaction of the Lindeberg condition \citep{billingsley1986probability}. We then apply the Lindeberg-Feller Central Limit Theorem \citep{hunter2014notes} to demonstrate asymptotic normality. For $s > n^\gamma$, which tends towards positive infinity, we show that the limit converges to 0. We conclude the proof by applying Slutsky's theorem to confirm asymptotic normality. The complete proof of Theorem~\ref{thm:asy} is provided in~\ref{appx52}.
Theorem~\ref{thm:asy} demonstrates that for $\gamma \in (1/3,1/2)$, as $n$ approaches infinity, $\hat{\nu}(z;n,n^{\gamma},m)$ follows a normal distribution with mean $\nu(z;n)$ and variance $\frac{1}{(n-1)(n^{\gamma})^2}\sum_{s=1}^{n^\gamma-1}s^2\zeta_{1,s}(z)$.
}


\subsection{Inference Procedure}

After establishing the asymptotic normality of the Monte Carlo estimate of Data Shapley, we proceed to statistical inference, which is crucial when we apply Data Shapley to practical problems. 
In this section, we focus on confidence intervals and hypothesis testing for uncertainty quantification and assess the significance of Data Shapley.

Confidence intervals are a powerful tool for quantifying the uncertainty surrounding Data Shapley values.
By estimating the range within which the Data Shapley value is likely to lie, confidence intervals provide valuable insights into the reliability and stability of data valuation. 
We construct confidence intervals based on the asymptotic normality of the Monte Carlo estimate of Data Shapley, capturing the inherent variability introduced by data distribution. 
According to Theorem~\ref{thm:asy}, the confidence interval is denoted as [LB, UB], where the lower and upper bounds, LB and UB, represent the $\alpha/2$ and $1-\alpha/2$ quantiles, respectively, of the normal distribution with mean ${\nu}(z,n)$ and variance $\frac{1}{(n^\gamma)^2}\sum_{s=0}^{n^\gamma-1}\frac{s^2\zeta_{1,s}(z)}{n-1}$.

Hypothesis testing enables us to evaluate the significance of Data Shapley and investigate potential deviations from expected outcomes.  
We formulate hypotheses regarding Data Shapley and employ statistical tests to assess their validity.  
By comparing observed Data Shapley with expected values under null hypotheses, we determine whether observed differences are statistically significant, shedding light on the effectiveness and reliability of data valuation techniques.
This confidence interval addresses hypotheses formulated as follows:
\begin{align*} &H_0:\nu(z;n)=c\ &H_1:\nu(z;n)\neq c \end{align*}

Formally, the Wald test statistic is defined as: \begin{equation} W_n:=\frac{\sqrt{n-1}(\hat{\nu}(z;n,n^\gamma,m)-c)}{\sqrt{\frac{1}{(n^\gamma)^2}\sum_{s=0}^{n^\gamma-1}s^2\hat{\zeta}_{1,s}(z)}}.
\end{equation}
Here, $n$ is the sample size, and we reject $H_0$ if $|W_n|$ exceeds the $1-\alpha/2$ quantile of the standard normal distribution. 
This implies a test with a type 1 error rate of $\alpha$, ensuring: \begin{align*}
    &P[\mbox{reject } H_0\mid H_0 \mbox{ is true}]\\&=P[|W_n|>1-\frac{\alpha}{2}\mid\hat{\nu}(z;n,n^\gamma,m)=c]\\&=\alpha.
\end{align*}  
However, this testing procedure is equivalent to checking whether $c$ falls within the calculated confidence interval. If $c$ is within the confidence interval, we fail to reject the hypothesis that the true mean prediction is equal to $c$; otherwise, we reject it.
In conclusion, the Wald confidence interval of $\nu(z;n)$ is $\hat{\nu}(z;n,n^\gamma,m)\pm U_{1-\alpha/2}\sqrt{\frac{1}{(n^\gamma)^2}\sum_{s=0}^{n^\gamma-1}\frac{s^2\hat{\zeta}_{1,s}(z)}{n-1}}$.

To summarize, our inference process includes the following steps:
\begin{itemize}
    \item 
    Estimation of Data Shapley: We can use all existing Monte Carlo methods to estimate Data Shapley, such as permutation sampling \citep{jia2019towards,ghorbani2019} and stratified sampling~\citep{wu2022variance}, which instantiate $m_s$ on Monte Carlo framework in different ways.
    \item
Construction of Confidence Intervals: Utilizing the asymptotic normality, we construct confidence intervals to quantify uncertainty. Here, we employ two algorithms to estimate variance, which will be discussed in  Section \ref{sec:estvar}.
\item
Hypothesis Formulation: We formulate null and alternative hypotheses regarding the mean value, setting the stage for hypothesis testing.
\item Statistical Testing: Employing the Wald test, we assess the significance of observed Data Shapley and validate our hypotheses.
\item Interpretation: We interpret the results of confidence intervals and hypothesis tests in the context of data valuation, providing insights into the reliability and robustness of Data Shapley estimates.
For example, we can estimate the difficulty of calculating data values based on variance and determine how many samples are needed for accurate estimation. It can also be anticipated how the value of data will change if other data are replaced with data from the same distribution. Moreover, similar to the detailed case study in the experiment section, by evaluating whether the selling price given by the seller falls within the confidence interval calculated by the buyer, the credibility of the seller's valuation of the data can be assessed.

\end{itemize}

\section{Estimating Algorithms}~\label{sec:estvar}

The limiting distribution (see Eq.\eqref{limiting distribution}) depends on the unknown mean parameter ${\nu}(z,n)$ as well as the unknown variance parameters $\zeta_{1,s}(z)$.
To use these quantities for statistical inference in practice, we must establish consistent estimators for these parameters.
While we can use the sample mean as a consistent estimate of ${\nu}(z,n)$, determining an appropriate variance estimate is less straightforward.
In this section, we introduce two methods for estimating the variance $\zeta_{1,s}(z)$.

\subsection{Double Monte Carlo}

The Double Monte Carlo method (DMC) involves two layers of Monte Carlo simulations to estimate $\zeta_{1,s}(z)$.
The inner layer generates Monte Carlo samples to estimate the conditional expectation, while the outer layer performs simulations to estimate the variances.
To estimate $\zeta_{1,s}(z)$, the outer Monte Carlo randomly selects a fixed data point from the dataset that does not contain the point of interest $z$, repeating this process $T$ times.
Within each iteration, the inner layer randomly selects subsets of data points (excluding $z$) of a fixed size $s-1$, repeating $T_i$ times. 
For each inner loop $t$ (assuming the corresponding outer loop is $i$), the marginal contribution of the point $z$ is calculated, obtaining $h_{i,t}$. 
For each outer loop, we calculate the mean $\bar{h}_i = \frac{1}{T_i}\sum_{t=1}^{T_i} h_{i,t}$ and then calculate the average of all loops $h = \frac{1}{T}\sum_{i=1}^{T}\bar{h}_i$.
Finally, the estimator $\hat{\zeta}_{1,s}(z)$ is computed using the sample variance:

\begin{equation}
\hat{\zeta}_{1,s}(z) := \frac{1}{T-1}\sum_{i=1}^{T}(h - \bar{h}_i)^2.
\end{equation}


The procedure for DMC is provided in Algorithm \ref{zeta_{1,s} estimation procedure}.
\begin{algorithm}[!ht]
	\caption{Double Monte Carlo estimator of  $\zeta_{1,s}(z)$}
	\label{zeta_{1,s} estimation procedure}
	\begin{algorithmic}[1]
	\REQUIRE
 Dataset that follows the distribution $\mathcal{D}$, learning algorithm $\mathcal{A}$, performance score $V$, sampling $T$ initial fixed points, sampling $T_i$ datasets for $i$th fixed point. \\
    \ENSURE $\zeta_{1,s}(z)$\\
    \FOR{$i\in \{1,\dots,T\}$}
	    \STATE 
	    Sample $1$ initial fixed data point from $\mathcal{D}$ which does not contain $z$;\\

	    \FOR{$t\in \{1,\dots,T_i\}$}
	    \STATE 
	    Sample $s-1$ data from $\mathcal{D}$ which do not contain $z$ to form $I_{i,t}$;\\
             \STATE
             $h_{i,t}:=V(I_{i,t}\cup \{z\})-V(I_{i,t})$;\\
            \ENDFOR
            \STATE 
            $\Bar{h}_{i}:=\frac{1}{T_i}\sum_{t=1}^{T_i}h_{i,t}$;\\
            \ENDFOR
            \STATE 
	     $h:=\frac{1}{T}\sum_{i=1}^{T}\Bar{h}_{i}$\\
      \STATE
      $\hat{\zeta}_{1,s}(z):=\frac{1}{T-1}\sum_{i=1}^{T}(h-\Bar{h}_{i})^2$.\\
\end{algorithmic}
\end{algorithm}

While DMC provides a systematic and straightforward way to estimate $\zeta_{1,s}(z)$,
it requires extensive computational resources due to the two levels of Monte Carlo simulations.  This motivates us to study alternative estimation methods with lighter computational complexity.

\subsection{Pick-and-Freeze}

The Pick-and-Freeze estimator (PF) is based on the following Lemma~\ref{cov}~\citep{Sobol2001}.
This lemma facilitates the Monte Carlo process by converting the variance of the conditional expectation into the covariance of two components, effectively
simplifying the double-layer Monte Carlo into a single-layer Monte Carlo.
\begin{lemma}
    Let $W=(X,Z)$. 
    If $X$ and $Z$ are independent, then
    \begin{equation*}
        \mathrm{Var}(\mathbb{E}(Y|X))=\mathrm{Cov} (Y,Y^X)
    \end{equation*}
    with $Y^X=f(X,Z^{\prime})$ and $Y=f(X,Z)$ where $Z^{\prime}$ is an independent copy of $Z$ and $f$ is any real-valued function.
    \label{cov}
\end{lemma}


According to Lemma~\ref{cov}, $\zeta_{1,s}(z)$ can be expressed as
\begin{equation}
\label{zeta_1s}
\zeta_{1,s}(z)=\mathrm{Cov}(h_s(Z_1,Z_2,\dots,Z_s;z),h_s(Z_1,Z_2^{'},\dots,Z_s^{'};z))
\end{equation}
where $Z_2^{'},\dots,Z_s^{'}\sim \mathcal{D}$ and are replications of $Z_2,\dots,Z_s$.
For simplicity, we denote $Z=(Z_2,Z_3,\dots,Z_s)$ and $Z^{'}=(Z_2^{'},\dots,Z_s^{'})$.
The Pick-and-Freeze estimator involves taking $2T$ samples as $(Z_1^{1},Z^{1}),(Z_1^{1},$
\\${Z^{'}}^{1}),\dots,(Z_1^{T},Z^{T}),(Z_1^{T},{Z^{'}}^{T})$,
and computing
\begin{align*}
\hat{\zeta}_{1,s}(z):=&\frac{1}{T}\sum_{i=1}^{T}h_s(Z_1^{i},Z^{i};z)h_s(Z_1^{i},{Z^{'}}^{i};z)\\
-&\left(\frac{1}{T}\sum_{i=1}^{T}h_s(Z_1^{i},Z^{i};z)\right)\left(\frac{1}{T}\sum_{i=1}^{T}h_s(Z_1^{i},{Z^{'}}^{i};z)\right).
\end{align*}
The procedure for Pick-and-Freeze estimator is provided in Algorithm \ref{Pick-and-Freeze}.

\begin{algorithm}[!ht]
	\caption{Pick-and-Freeze estimator of $\zeta_{1,s}(z)$}
	\label{Pick-and-Freeze}
	\begin{algorithmic}[1]
	\REQUIRE
	Dataset that follows the distribution $\mathcal{D}$, learning algorithm $\mathcal{A}$, performance score $V$, number of samples $T$.\\
    \ENSURE $\zeta_{1,s}(z)$\\
    \FOR{$i\in \{1,\dots,T\}$}
	    \STATE 
	    Sample $1$ initial fixed data from $\mathcal{D}$ which do not contain $z$ as $Z_1^{i}$;\\
        Sample $s-1$ data from $\mathcal{D}$ which do not contain $z$ as $Z^{i}$;\\
        Calculate and save $h_s(Z_1^i,Z^i;z)$;\\
        Sample $s-1$ data from $\mathcal{D}$ which do not contain $z$ as ${Z^{'}}^{i}$;\\
        Calculate and save $h_s(Z_1^i,{Z^{'}}^{i};z);$
    \ENDFOR
             \STATE\begin{align*}
\hat{\zeta}_{1,s}(z):=&\frac{1}{T}\sum_{i=1}^{T}h_s(Z_1^{i},Z^{i};z)h_s(Z_1^{i},{Z^{'}}^{i};z)\\
-&\left(\frac{1}{T}\sum_{i=1}^{T}h_s(Z_1^{i},Z^{i};z)\right)\left(\frac{1}{T}\sum_{i=1}^{T}h_s(Z_1^{i},{Z^{'}}^{i};z)\right).
\end{align*}
\end{algorithmic}
\end{algorithm}


\begin{remark}
Comparing the two algorithms, DMC and PF, we find that DMC first samples $T$ fixed initial data points and then samples $T_i$ datasets, each composed of $s-1$ data points for each initial data point. 
In contrast, PF samples $T$ fixed initial data points and then samples two datasets, each composed of $s-1$ data points for each initial data point. According to Lemma \ref{cov}, when the data are completely independent, the two estimators are equivalent. 
Specifically, when both algorithms set the same $T$ and DMC sets $T_i$ to 2, the estimated values for the same samples should be the same. 
However, in practice, complete independence of the data is not guaranteed.
DMC, as a direct estimation method, has the advantage of higher accuracy as $T_i$ increases but at the cost of higher computational time. 
PF converts the two-layer nested sampling of DMC into a single-layer sampling, effectively reducing computation time. 
Thus, if the time cost is acceptable, the DMC algorithm can be chosen. 
If the time cost is a critical concern, PF can be considered as an alternative.
\end{remark}


\section{Experiments}
Our experiments mainly focus on the following aspects. Firstly, we visually demonstrate the asymptotic normality of Data Shapley estimates through histograms and fitted normal curves. We compare these visualizations across varying amounts of data to illustrate how the distribution approaches normality as the sample size increases.
Secondly, we evaluate and compare the strengths and limitations of the two proposed algorithms for estimating $\zeta_{1,s}(z)$. This comparison provides insights into the performance and applicability of each method under different conditions. 
Thirdly, we present the process and results of hypothesis testing for data valuation using real-world data. 
We demonstrate the construction and interpretation of confidence intervals, showcasing their practical application in assessing the reliability of Data Shapley estimates.
Fourthly, we propose and analyze a realistic scenario in data trading where confidence intervals can be effectively applied. This scenario illustrates the practical utility of our statistical inference approach in real-world data valuation contexts.

\subsection{Experimental Setting}

We validate the previous results using multiple datasets, as shown in Table \ref{Data_sets} of \ref{exp}. 
For the main text, we focus on two illustrating examples: the Covertype dataset (a tabular dataset) and FashionMNIST (an image dataset). Results for other datasets are provided in the supplementary materials. 
The Covertype dataset involves the classification of seven forest cover types based on attributes such as elevation, slope direction, slope, shade, and soil type. 
Various classification tasks are performed on this dataset.
FashionMNIST consists of $60,000$ grayscale images ($28\times28$ pixels) across $10$ fashion categories, along with a test set of $10,000$ images. 

As our focus is on data valuation, which is not directly tied to a specific machine learning algorithm, we opt for the classical logistic regression.
Model performance is measured by prediction accuracy. 
Our evaluation includes datasets with sizes ranging from $50$ to $2000$ samples. To assess its normality, each experiment is randomly tested $30$ times.

\subsection{Experimental Results}
\begin{figure}[ht]
\begin{center}
\centerline{\includegraphics[width=0.7\columnwidth]{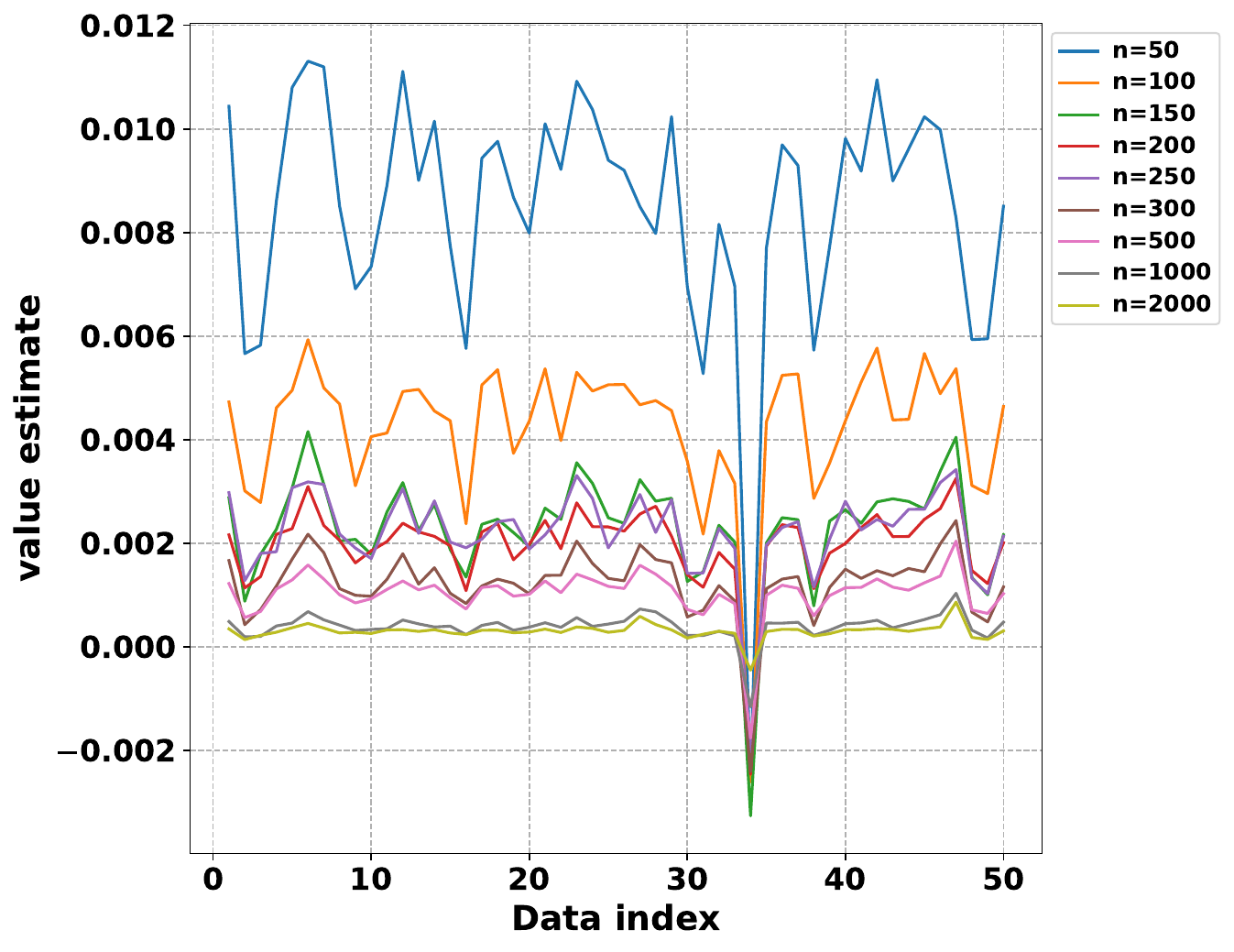}}
\caption{Estimation of values for 50 samples in the FashionMNIST dataset using the Logistic Regression learning algorithm, with the prediction accuracy as the utility function. 
}
\label{f11}
\end{center}
\vskip -0.2in
\end{figure}

\textbf{Trend of Estimated Data Shapley with Increasing Data size}
We explore the variation in data values as the data size increases. Figure \ref{f11} illustrates the estimated values for the first 50 points in the dataset using data sizes ranging from 50 to 2000 points. It is evident that as the data size increases, the valuations across different data points tend to stabilize.

\begin{figure*}[htbp]
    \centering
    \begin{minipage}{\textwidth}
        \begin{minipage}{0.33\textwidth}
            \centering
            \includegraphics[width=\linewidth]{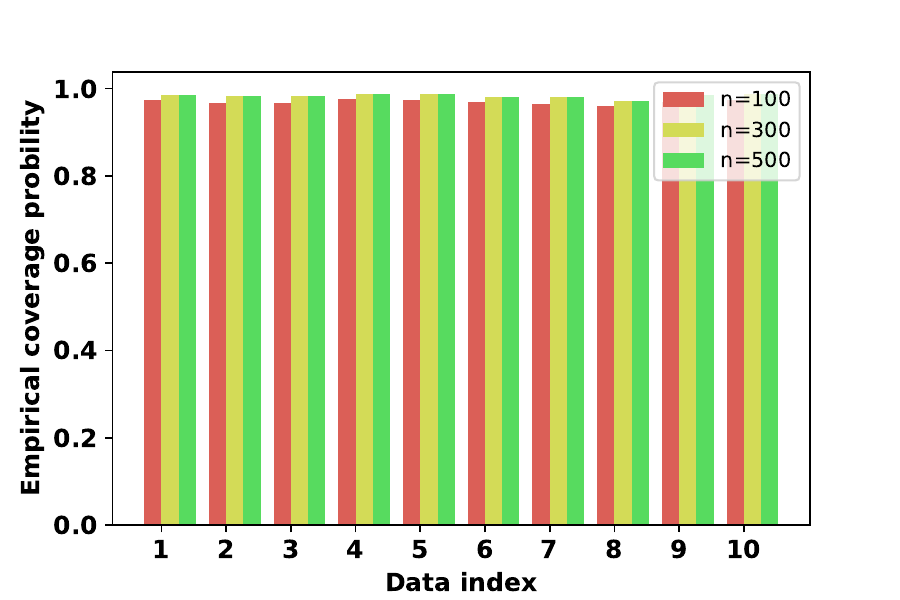}
         \subfloat[]{\label{f2}}
        \end{minipage}%
        \hfill 
        \begin{minipage}{0.33\textwidth}
            \centering
            \includegraphics[width=\linewidth]{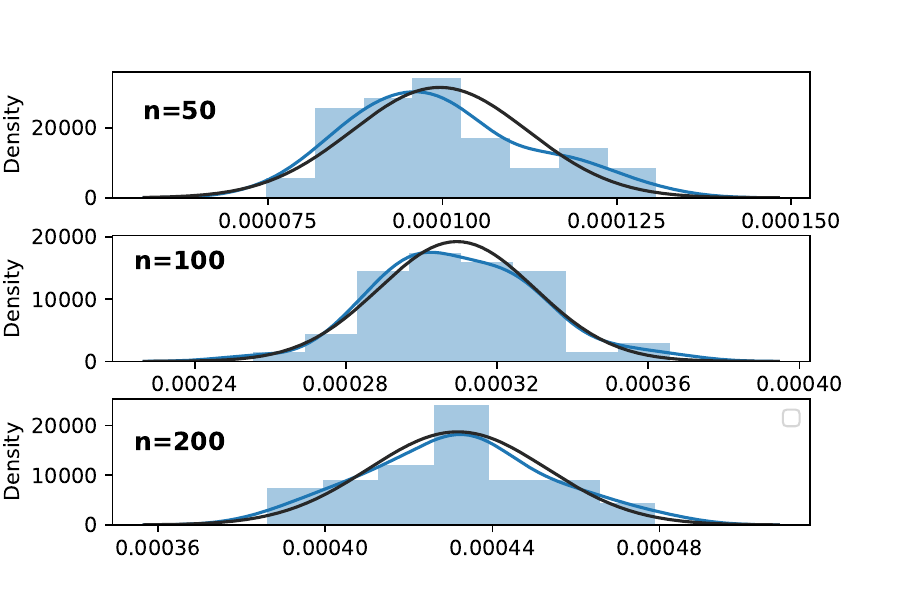}
            \subfloat[]{\label{f12}}
        \end{minipage}
        \hfill
        \begin{minipage}{0.33\textwidth}
            \centering
            \includegraphics[width=\linewidth]{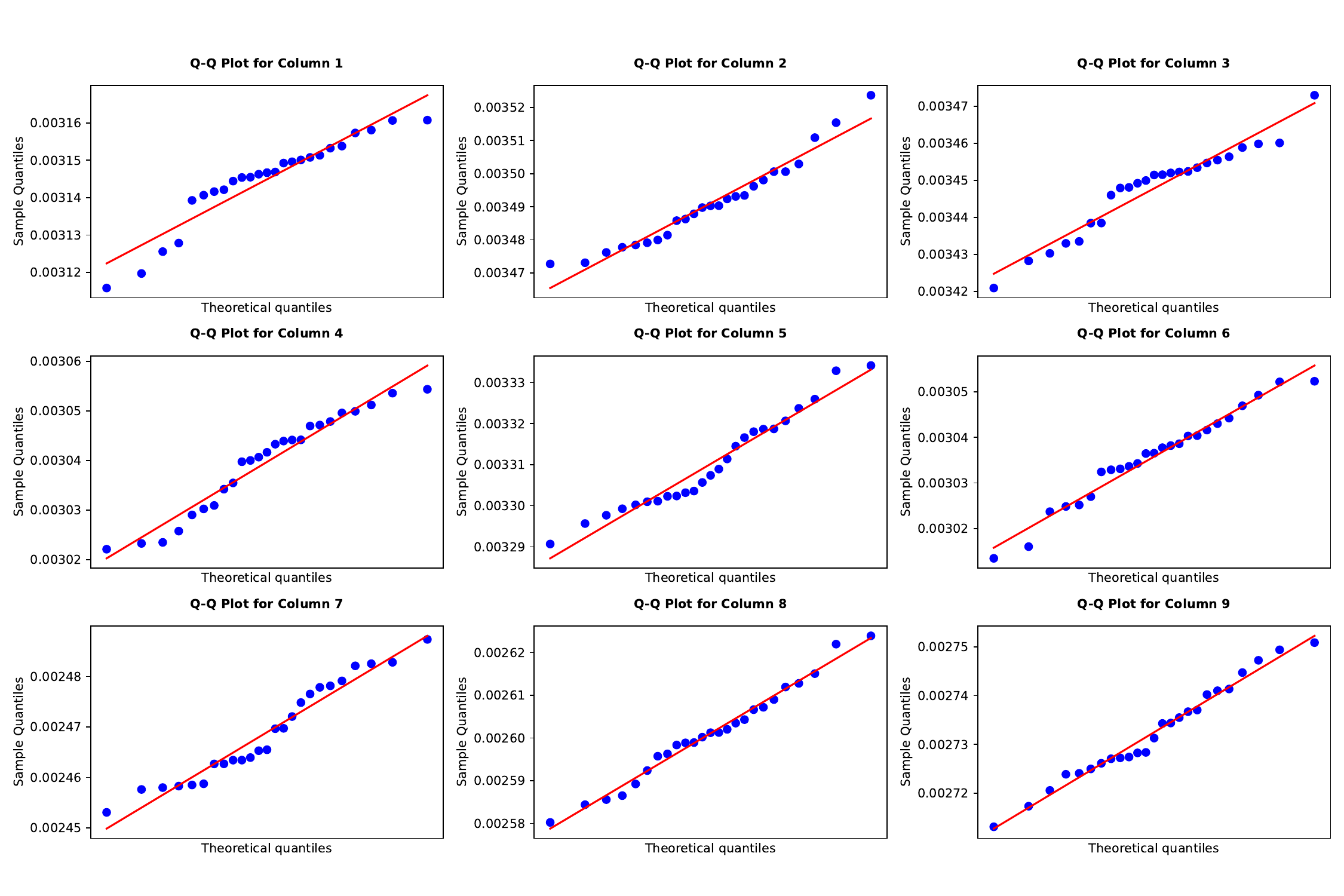}
            \subfloat[]{\label{f22}}
        \end{minipage}
    \end{minipage}
    \caption{The behavior of estimated Data Shapley values.
Figure \ref{f2} presents the empirical coverage of 100 randomly selected points, showing the proportion of sample data values falling within their intervals as the data volume $n$ are set to 100, 300, and 500.
Figure \ref{f12} displays histograms of the estimated values for a randomly selected sample, with experiments repeated 50 times; the blue line represents the fitted curve, while the black line indicates a normal distribution curve with the same mean and variance. 
Figure \ref{f22} shows the Q-Q plot of 9 samples.}
    \label{fig:asy}
\end{figure*}

\textbf{Asymptotic Normality}
We set the dataset sizes 
$n$ to 100, 300, and 500, and compute the Monte Carlo estimates of Data Shapley values along with their corresponding 95\% confidence intervals. Subsequently, we determine the proportion of data values falling within these intervals,
which is defined as empirical coverage rate.
Figure \ref{f2} shows the empirical coverage rate of $100$ randomly selected points under different 
$n$ conditions. 
Specific values can be found in Table~\ref{tab0}.
As the data size
$n$ increases, the empirical coverage rate steadily rises.

Additionally, we present histograms of the estimated values for a randomly selected sample when training with different data sizes in Figure \ref{f12}. 
The experiments are repeated 50 times. The blue line in the figure represents the fitted curve, while the black line represents a normal distribution curve with the same mean and variance. 
It is observable that as the data size increases, the estimated values tend towards a normal distribution.
Figure \ref{f22} shows the Q-Q plot of 9 samples.
It is observable that the estimated values tend towards a normal distribution.

\begin{table*}[t]
    \caption{Empirical coverage.}
    \label{tab0}
    \centering
    \begin{tabularx}{\linewidth}{l|XXXXXXXXXX} 
        \toprule
        Data volume & \multicolumn{10}{c}{Data index} \\
        \cmidrule(lr){2-11}
        & 1 & 2 & 3 & 4 & 5 & 6 & 7 & 8 & 9 & 10 \\
        \midrule
        m=100 & 0.973 & 0.966 & 0.966 & 0.976 & 0.975 & 0.970 & 0.964 & 0.961 & 0.970 & 0.973 \\
        m=300 & 0.985 & 0.984 & 0.983 & 0.987 & 0.989 & 0.981 & 0.981 & 0.972 & 0.986 & 0.987 \\
        m=500 & 0.990 & 0.986 & 0.985 & 0.991 & 0.991 & 0.988 & 0.985 & 0.980 & 0.987 & 0.988 \\
        \bottomrule
    \end{tabularx}
\end{table*}

\begin{figure}[ht]
\vskip 0.2in
\begin{center}
\centerline{\includegraphics[width=0.7\columnwidth]{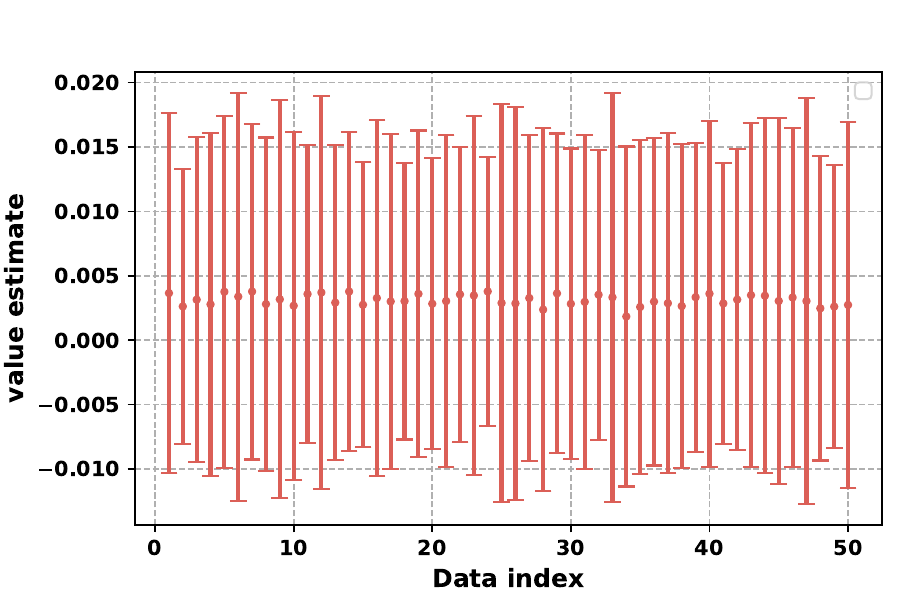}}
\caption{The estimated values and confidence intervals for the top 50 points on the FashionMNIST data, with the data volume set to 100.
}
\label{f21}
\end{center}
\vskip -0.2in
\end{figure}

\textbf{Confidence Interval and Hypothesis Testing}
{
We calculate confidence intervals for select data points in all datasets mentioned in Table \ref{Data_sets} and conduct hypothesis testing on the data values. 
Figure \ref{f21} shows the confidence intervals constructed for the top 50 points on the FashionMNIST data, with the data size set to 100. 
The process involves first estimating the Data Shapley value using the Monte Carlo algorithm and then estimating $\zeta_{1,s}$ using the Double Monte Carlo algorithm. 
Next, we set the null hypothesis that the estimated Data Shapley values are equal to the true values and apply the Wald test for hypothesis testing to construct 95\% confidence intervals. 
The proportion of estimated data values falling within their confidence interval, i.e., empirical coverage rate, is then tested, with results shown in Figure \ref{f2}. 
Results for other datasets can be found in \ref{exp}.
}

\begin{figure*}[htbp]
    \centering
    \begin{minipage}{\textwidth}
        \begin{minipage}{0.45\textwidth}
            \centering
         \includegraphics[width=\linewidth]{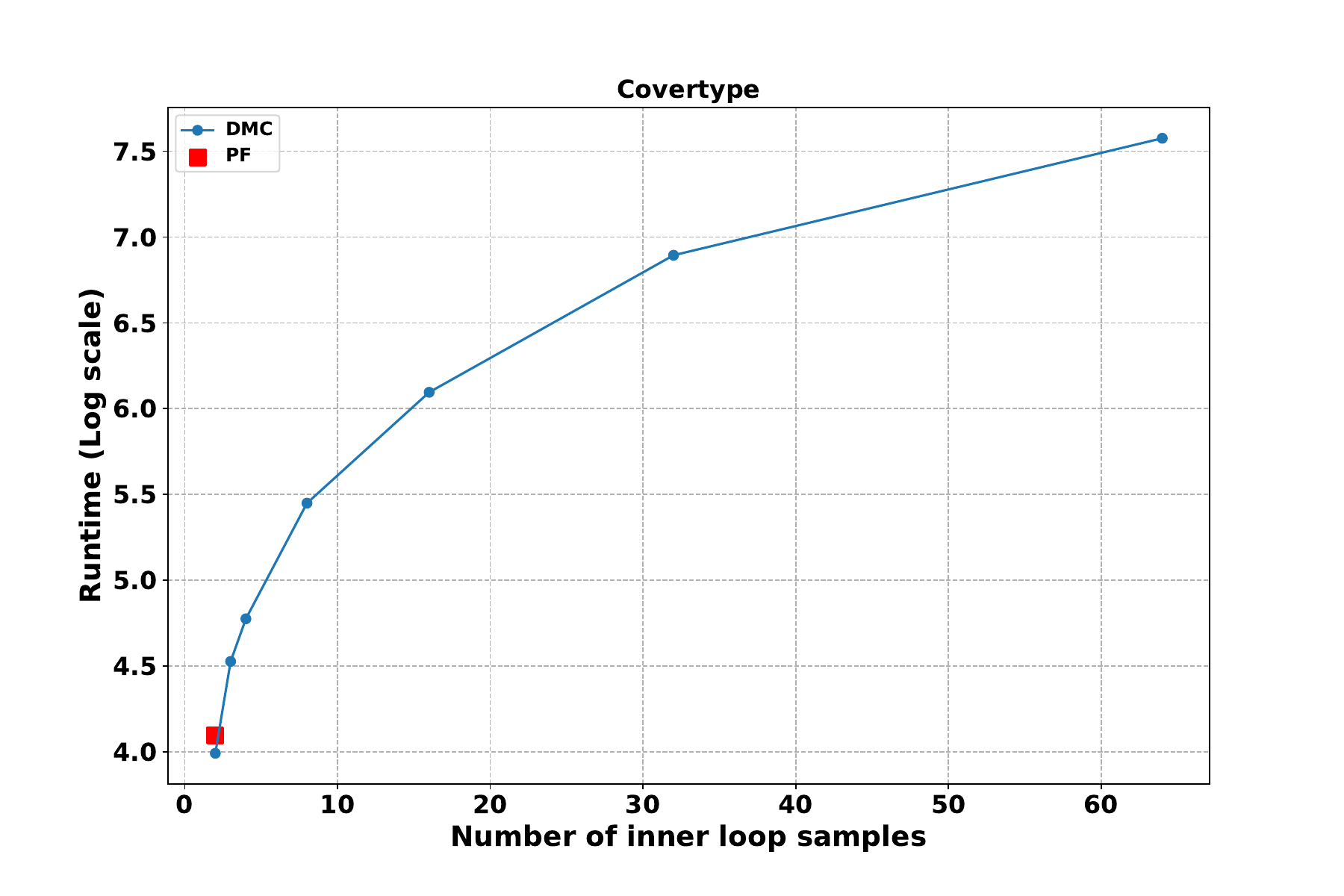}
         \subfloat[]{\label{il1}}
        \end{minipage}%
        \begin{minipage}{0.45\textwidth}
            \centering
            \includegraphics[width=\linewidth]{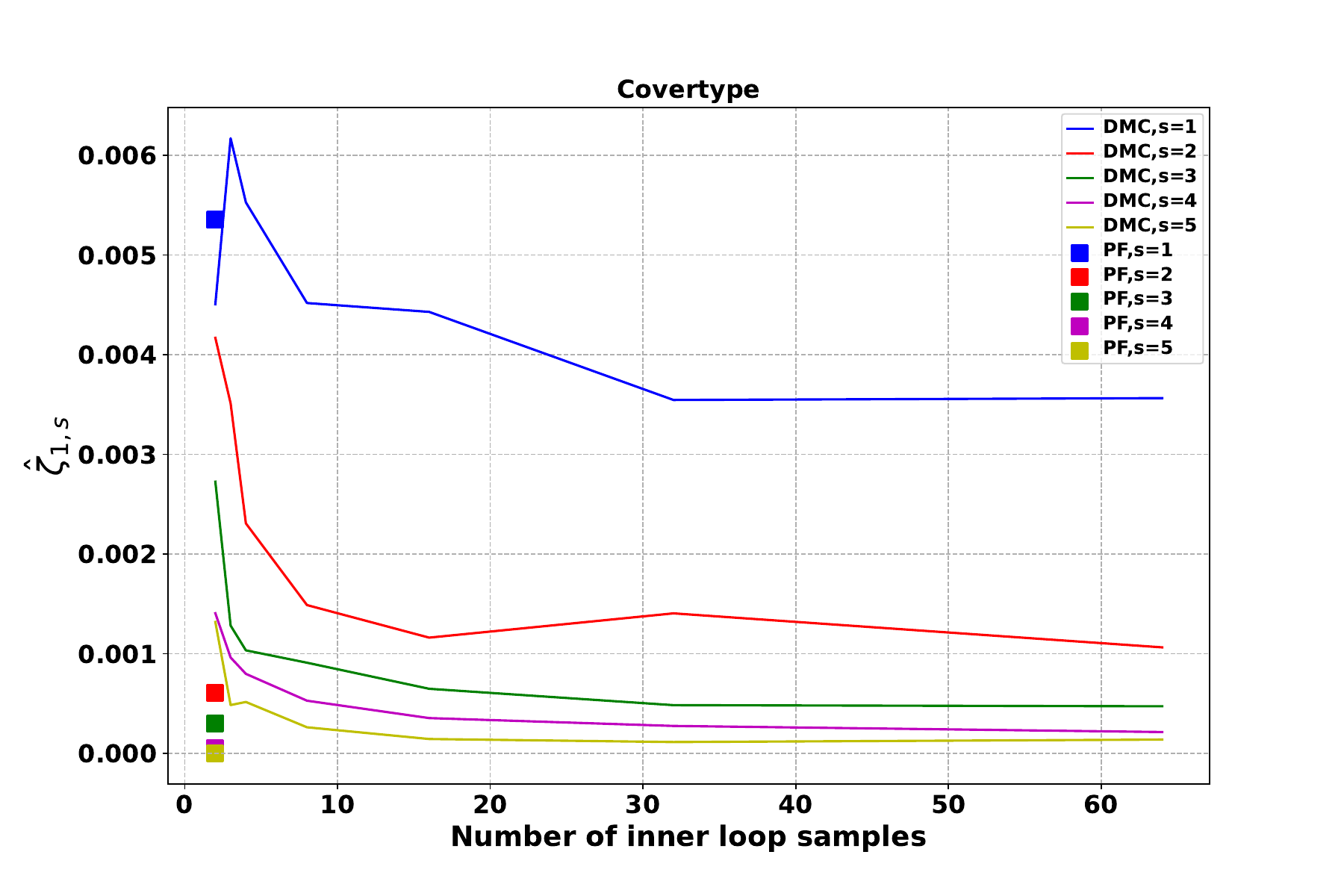}
            \subfloat[]{\label{il2}}
        \end{minipage}
    \end{minipage}
    \caption{Comparison of DMC and PF algorithms on the Covertype dataset. The left panel (Figure \ref{il1}) shows the runtime comparison, with $T=100$ for both algorithms and varying inner loop counts for DMC. The running time of DMC increases rapidly with the number of inner loops, far exceeding that of PF (vertical axis represents the logarithm of time). The right panel (Figure \ref{il2}) shows the estimated values of $\zeta_{1,s}$ using DMC with varying $T_i$. As $T_i$ increases, the estimated values become more stable and accurate.}
    \label{fig:variance}
\end{figure*}

\textbf{Limiting Variance }
{
We estimate $\zeta_{1,s}$ using both DMC and PF algorithms on the FashionMNIST and Covertype datasets and compare their runtime and accuracy. 
Figure \ref{il1} shows the time comparison on the Covertype dataset. With $T=100$ for both algorithms and the number of inner loops in DMC exceeding 2, the running time of DMC increases rapidly (the vertical axis represents the logarithm of time), far surpassing that of PF. 
Figure \ref{il2} shows the estimated values of $\zeta_{1,s}$. 
It can be observed that as $T_i$ increases, the estimated values tend to become more stable, indicating increased accuracy. Therefore, if accuracy is the primary concern, DMC is preferred, whereas if time cost is a significant factor, PF is recommended. The results on the FashionMNIST dataset can be found in \ref{exp}.
}

\textbf{Data Addition}
Leveraging our theory to quantify the uncertainty of data valuation, we propose a new method that performs well in data addition tasks under specific circumstances.
This method is based on the three-sigma rule, which means that 99.7\% of the estimated data falls within three standard deviations from the mean.
It combines Data Shapley with confidence intervals to estimate data value using the formula $\nu(z;n)$ is $\hat{\nu}(z;n,n^\gamma,m) +\lambda \cdot std$, where $std = \sqrt{\frac{1}{(n^\gamma)^2}\sum_{s=0}^{n^\gamma-1}\frac{s^2\hat{\zeta}_{1,s}(z)}{n-1}}$ and $\lambda \in [-3,3]$ is a tuning parameter.
When $\lambda=0$, it corresponds to the Data Shapley value.

The data addition task involves adding data to the model in ascending order of values and observing changes in model performance.
A good data valuation method can accurately identify the quality of data, allowing for the quick and effective addition of high-quality data, thereby resulting in a rapid increase in model performance.
We compare three data valuation methods for data addition tasks: Data Shapley combined with confidence interval, Data Shapley, and random selection.

\begin{figure*}[htbp]
    \centering
    \begin{minipage}{\textwidth}
        \begin{minipage}{0.45\textwidth}
            \centering
        \includegraphics[width=\linewidth]{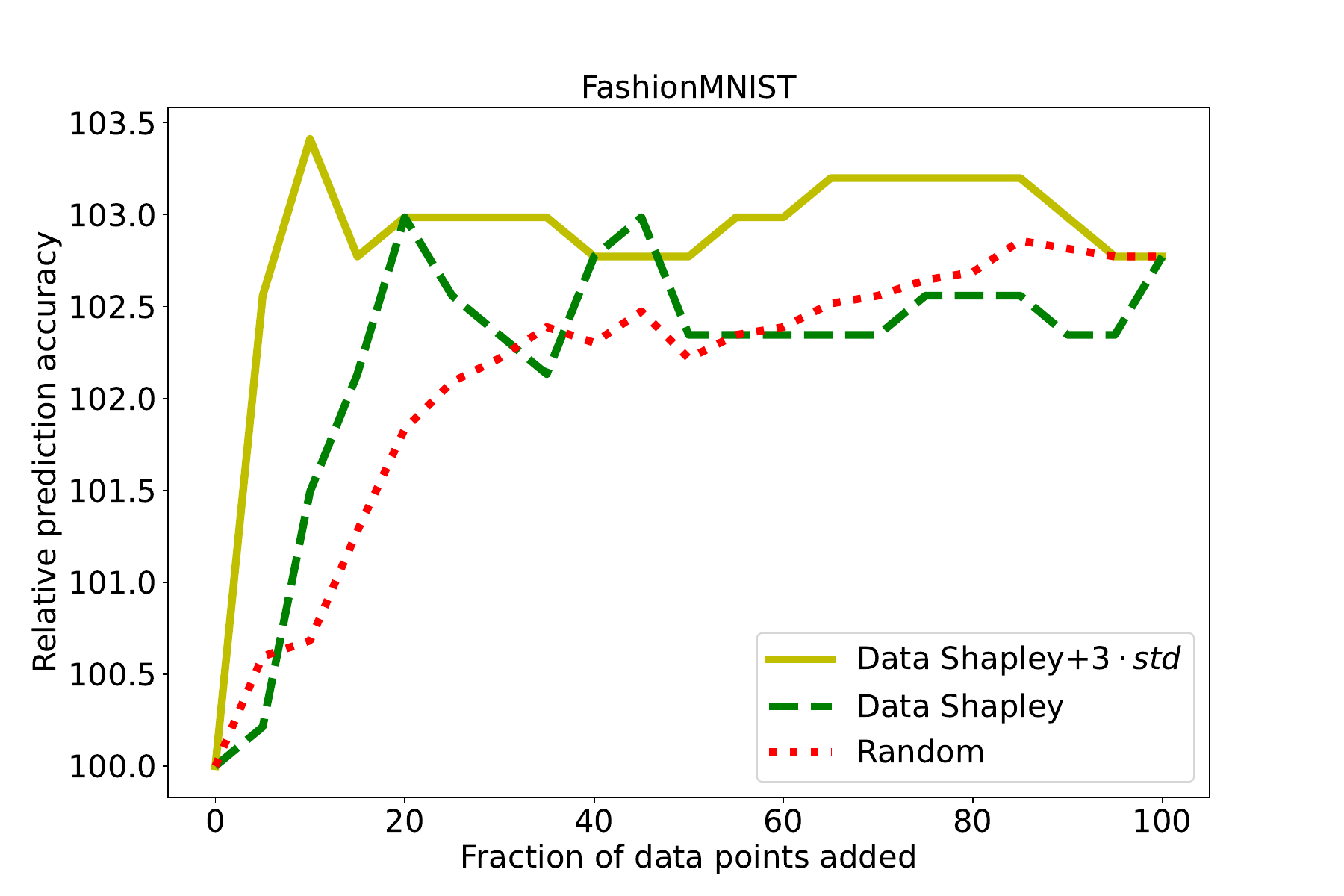}
         \subfloat[]{\label{adft}}
        \end{minipage}%
        \begin{minipage}{0.45\textwidth}
            \centering
            \includegraphics[width=\linewidth]{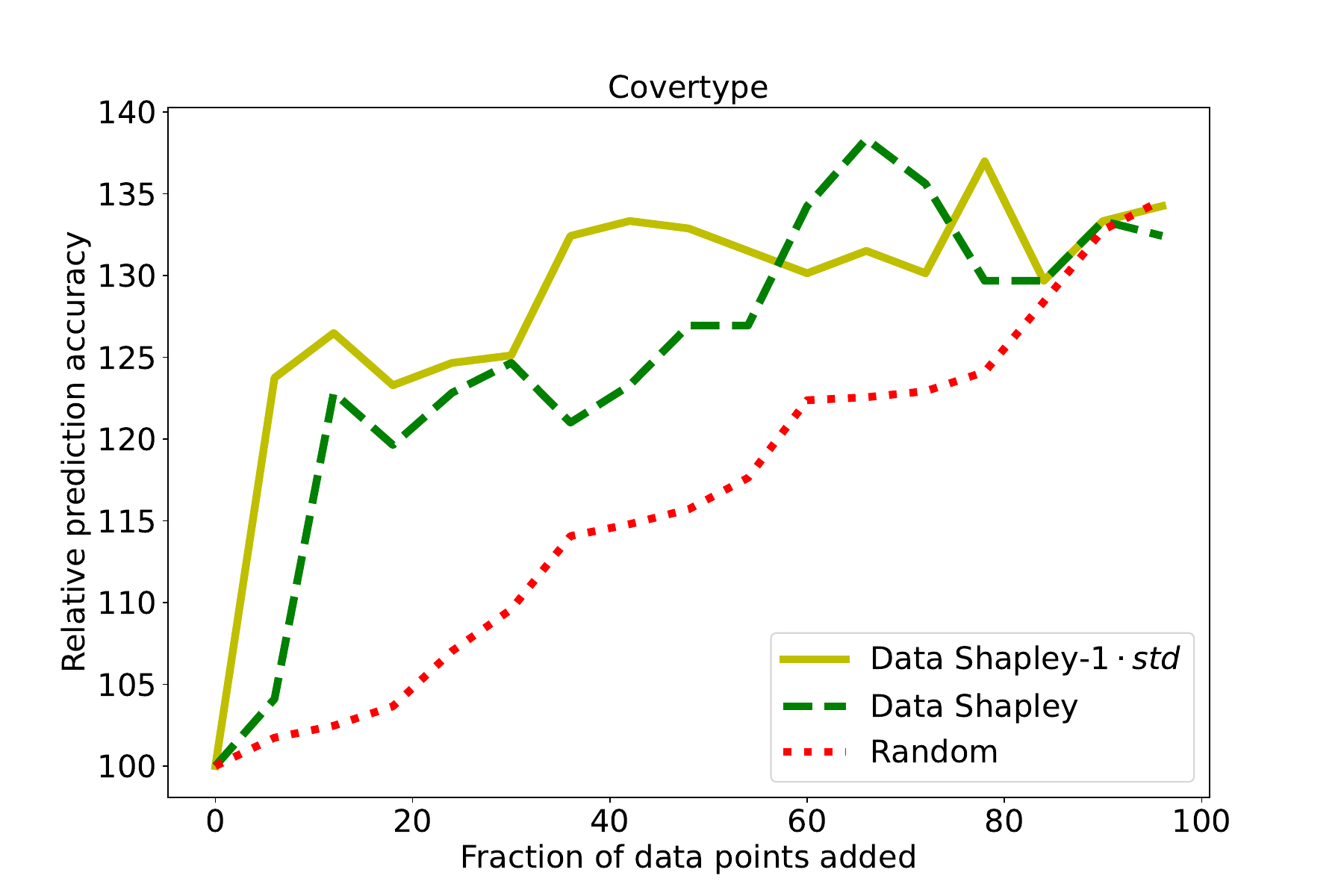}
            \subfloat[]{\label{adcov}}
        \end{minipage}
        \hfill
    \end{minipage}
    \caption{Performance comparison of different data valuation methods in data addition tasks on the FashionMNIST and Covertype datasets. 
    The figure shows the relative prediction accuracy as additional data points are added to the model in ascending order of their estimated values. 
    Three methods are compared: Data Shapley combined with confidence interval, Data Shapley, and random allocation. 
    The y-axis represents the ratio of the current prediction accuracy to the initial prediction accuracy (with 5 data points in the training dataset).
    On the FashionMNIST dataset, $Data Shapley + 3 \cdot std$ approach picks out high-value points faster, while on the Covertype dataset, $Data Shapley - 1 \cdot std$ approach performs better, where $\gamma =2/3$ and $n=100$.}
    \label{add}
\end{figure*}

Figure \ref{add} shows the experimental results of data addition tasks conducted on the FashionMNIST and Covertype datasets. 
At the beginning of the training, we place 5 pieces of data in the training set, sort the other 50 data points according to the data value, and add them to the dataset in ascending order. 
Different valuation methods will assign different values to the data, resulting in different rankings.
We compare the changes in prediction accuracy when data is sequentially added to the training set under different valuation methods.
The relative prediction accuracy on the vertical axis refers to the ratio of the current prediction accuracy to the initial prediction accuracy (with 5 data points in the training dataset).

We observe that Data Shapley method is not optimal. 
When we combine Data Shapley with the confidence intervals and proper tuning parameter $\lambda$ setting to valuation, it quickly selects valuable data, resulting in a rapid increase in the performance curve at the beginning.
It should be noted that we present two different values of $\lambda$. 
In fact, $\lambda$ can take any value within $[-3,3]$, and the optimal value of $\lambda$
can be searched for within this continuous interval. 
A principled method for searching the best $\lambda$ will be left for future discussion.




\subsection{Case Study}

\begin{figure}[ht]
    \begin{center}
\centerline{\includegraphics[width=0.8\columnwidth]{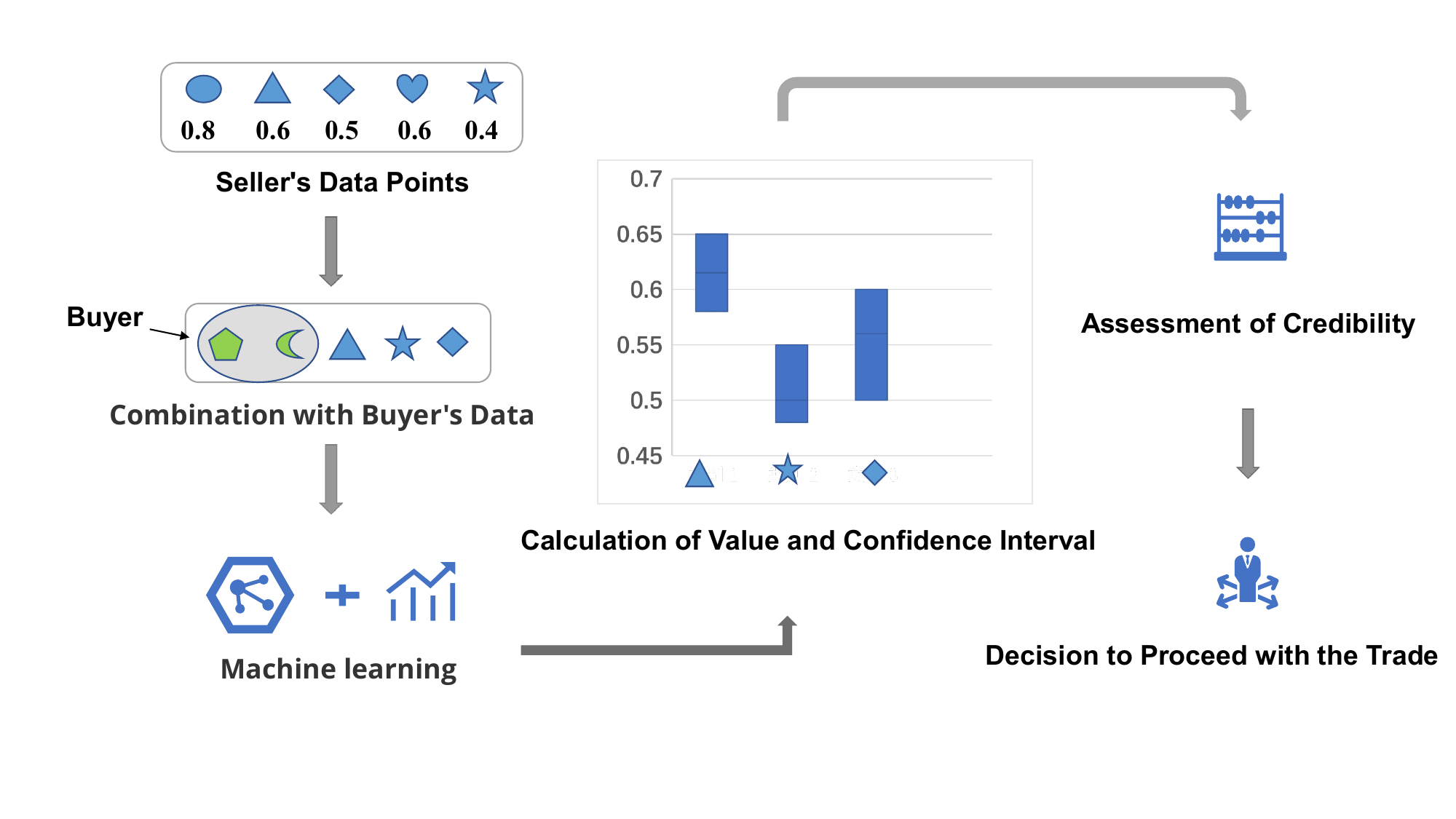}}
\vspace{-10mm}
    \caption{Transaction process dependent on confidence intervals.}
\label{case}
\end{center}
\end{figure}

Our data valuation approach is not limited to a specific dataset. It assumes data follows a distribution, making it highly applicable in data trading markets. 
Consider the following case: A seller has 200 data points for sale and can determine the value of these points as a basis for setting prices. 
When a buyer intends to purchase data but is uncertain about its accuracy and value as claimed by the seller, the buyer can randomly select a subset of the seller's data for evaluation. 
In our experiment, we set the size of this subset to 30. 
The buyer combines this subset with his existing data to calculate the value of the seller's data and establish confidence intervals.
This approach assumes that both the buyer's and seller's data originate from the same population or environmental conditions, thus following the same distribution. 
It is important to note that in this scenario, we equate data value with price. 
While this equivalence may not hold in all practical scenarios, our case study serves as a proof of concept and provides groundwork for value verification. 
The intricacies of mapping values to prices are reserved for future research.
By examining how many of the 30 data points have prices within the buyer's calculated confidence interval, the buyer can assess the credibility of the data value provided by the seller and decide whether to proceed with the trade.
The visual process description is shown in Figure \ref{case}.

\begin{figure*}[htbp]
    \centering
    \begin{minipage}{\textwidth}
        \begin{minipage}{0.33\textwidth}
            \centering
            \includegraphics[width=\linewidth]{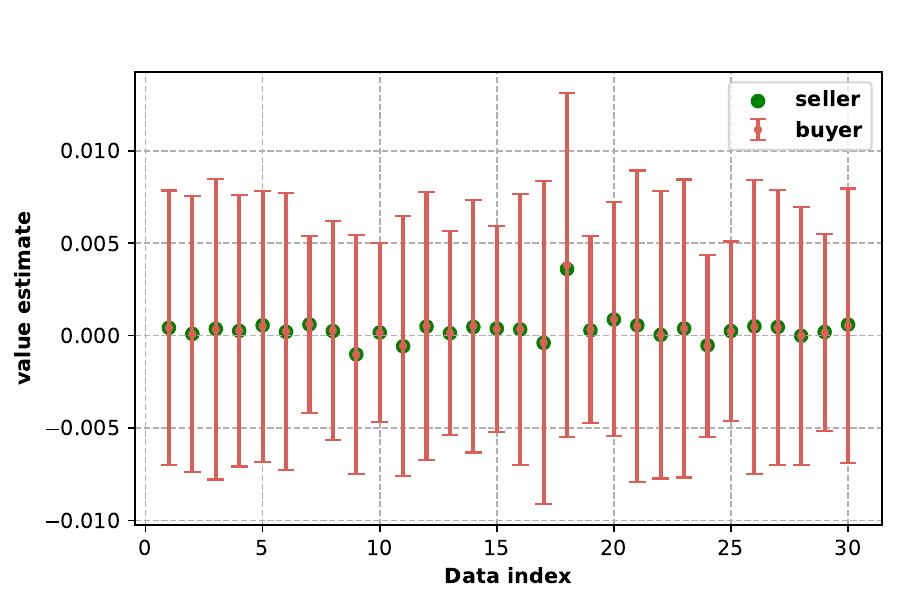}
         \subfloat[]{\label{f41}}
        \end{minipage}%
        \hfill 
        \begin{minipage}{0.33\textwidth}
            \centering
            \includegraphics[width=\linewidth]{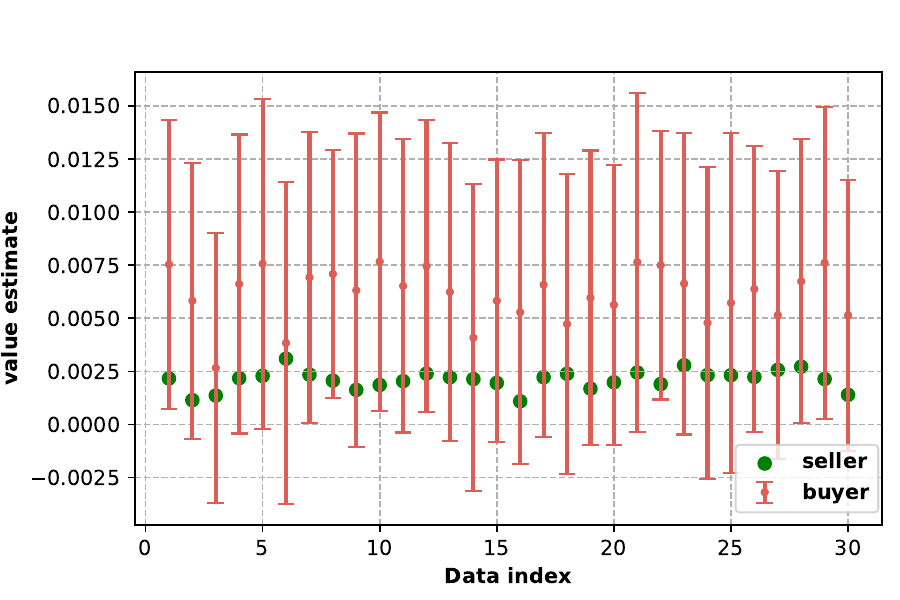}
            \subfloat[]{\label{ftm}}
        \end{minipage}
        \hfill
        \begin{minipage}{0.33\textwidth}
            \centering
            \includegraphics[width=\linewidth]{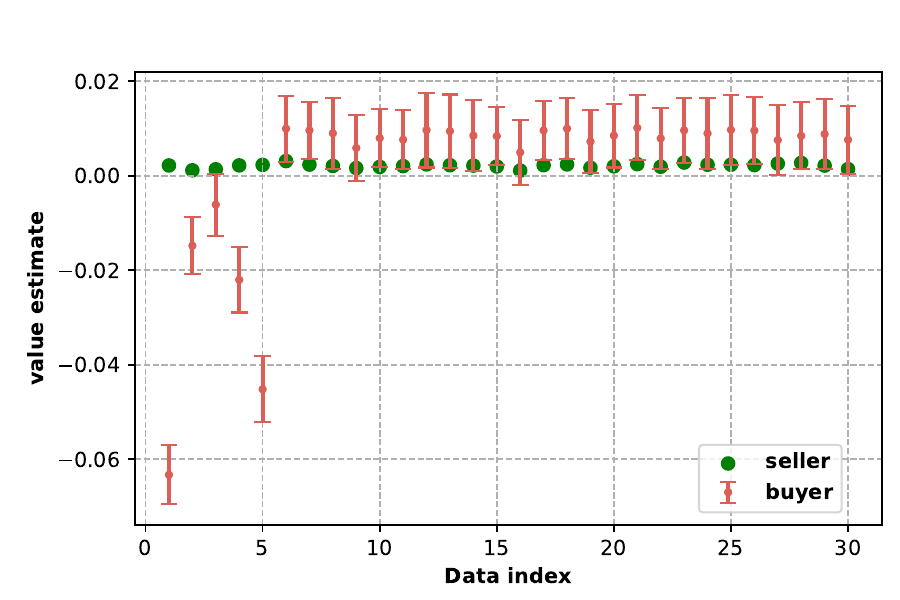}
            \subfloat[]{\label{ftmc}}
        \end{minipage}
    \end{minipage}
    \caption{The data value provided by the seller and the buyer.
    Figure \ref{f41} shows the results on the Covertype dataset.
    Figure \ref{ftm} presents the results on the FashionMNIST dataset.
    In both figures, because real data is used, the seller's estimates fall within the confidence interval estimated by the buyer.
    Figure \ref{ftmc} illustrates that when the labels of the first five data points are flipped, the seller's estimates no longer fall within the confidence interval estimated by the buyer.
}
    \label{fig:case}
\end{figure*}

As illustrated in Figure \ref{f41}, using the Covertype dataset in our experiment, all of the seller's pricing falls within the confidence interval of the buyer's estimate. Furthermore, the seller's valuation is remarkably close to the buyer's valuation, making it challenging to distinguish their differences visually.


In our investigation using the FashionMNIST dataset, the seller provides 200 data points, while the buyer possesses 70 data points. By blending the seller's 30 data points with their own set of 30 data points, the buyer computes the value and confidence interval for the seller's data. It is observed that the seller's provided valuation fell within the buyer's calculated confidence interval, as depicted in Figure \ref{ftm}. This outcome is expected, considering that our dataset comprises real FashionMNIST data.


{To verify the validity of the method for noisy data, we conduct an additional experiment. 
We select the top 5 of the 30 data passed from the seller to the buyer and flip their labels. 
Specifically, we change the label of the data with the correct label ``T-Shirt'' to ``Shirts'' and the label of the data with the correct label ``Shirts'' to ``T-Shirt.'' 
The experiment justifies that the confidence interval of the estimated values of these data with modified labels no longer encompasses the seller's valuation, and the buyer's valuation was significantly lower than the seller's valuation, as illustrated in Figure \ref{ftmc}. 
This experiment yields two significant insights. 
Firstly, buyers can assess the accuracy of the seller's valuation by determining whether it falls within their calculated confidence interval. 
Secondly, buyers can evaluate the quality of the data based on their own valuation of the seller's data. 
These findings suggest that confidence intervals provide an effective mechanism for evaluating the accuracy of the seller's valuation.
}


So far, our investigation has encompassed the variations in estimated values with sample size, the asymptotic normality of values, scrutiny of confidence intervals, and the execution of hypothesis testing, as well as the estimation of limit variance for Data Shapley.
Lastly, we delve into a practical application case, revealing that furnishing a confidence interval for the value enhances the credibility of the valuation.

\section{Conclusion}
In this study, we analyze Data Shapley from the perspective of U-statistics and derive the asymptotic normality of its estimated values. Utilizing this property, we conduct statistical inference and obtain confidence intervals for data value estimation. We propose two methods for estimating limiting variance and provide recommendations for their applicability. Finally, we conduct simulations on real datasets and present a case of trust trading based on confidence intervals in a data trading market. In future work, we aim to extend these valuations to more real-world trading scenarios, thereby enhancing the credibility and feasibility of transactions.


\bibliographystyle{plainnat}

\newpage

\section*{Appendix}\label{sec:appendix}

\renewcommand{\thesubsection}{Appendix~\Alph{subsection}}

\subsection{Proof of Lindeberg condition}
\setcounter{equation}{0}
\renewcommand{\theequation}{A.\arabic{equation}}

\begin{lemma}
    Let $Z_1,Z_2,\dots \overset{i.i.d}{\sim}\mathcal{D}$,    $\nu_s(z;n)=\mathbb{E}h_s(Z_{i_1},\dots,Z_{i_s};z)$, assuming $h_s \leq \beta(s)= \mathcal{O} (\frac{1}{s+1})$  and $h_{1,s}(Z;z)=\mathbb{E}h_s(Z,Z_{i_2},\dots,Z_{i_s};z)-{\nu}_s(z,n)$.
    Then for all $\delta>0$, and $s\sim n^{\gamma}, \gamma\in(0,1/2)$,
\begin{equation}   \lim_{n\rightarrow\infty}\frac{1}{\zeta_{1,s}(z)}\int_{|h_{1,s}(Z;z)|\geq \delta\sqrt{(n-1)\zeta_{1,s}(z)}}h^2_{1,s}(Z;z)dP=0.
\end{equation}
\label{lindeberg}
\end{lemma}

This condition is used to control the predicted tail behavior and help us obtain the conclusion of the Theorem~\ref{thm:layer_asymp}.

\textbf{Proof} \quad According to Hoeffding's inequality and $h_s \leq \beta(s)$, we have
  \begin{equation}
    \mathbb{P}\left(\left|h_{1,s}(Z;z)-0\right|\geq \delta\sqrt{(n-1)\zeta_{1,s}(z)}\right)\leq 2e^{-\frac{c\delta^2n\zeta_{1,s}(z)}{\beta^2(s)}}.
    \label{P}
\end{equation}
Then since $\beta(s)= \mathcal{O} (\frac{1}{s+1})$,
\begin{equation*}
    \frac{\delta^2n\zeta_{1,s}(z)}{\beta^2(s)} \geq c_1 \delta^2n\zeta_{1,s}(z)(s+1)^2.
\end{equation*}
According to the Cramer Rao lower bound inequality, we assume the existence of a constant $c_2$ such that $\zeta_{1,s}\geq c_2$.
Then since $s\sim n^{\gamma}, \gamma\in(0,1/2)$,
we have
$\frac{\delta^2n\zeta_{1,s}}{\beta^2(s)} \geq c\delta^2n^{1+2\gamma}$.
So this probability in Eq.\ref{P} is less than an exponential rate with respect to some positive order of $n$.
Then since $h^2_{1,s}(Z;z)$ is bounded, we have 
\begin{align*}
    &\lim_{n \to \infty} \frac{1}{\zeta_{1,s}(z)} \int_{|h_{1,s}(Z;z)| \geq \delta \sqrt{(n-1) \zeta_{1,s}(z)}} h_{1,s}^2(Z;z)dP \\
    &\leq \lim_{n \to \infty} \frac{2M}{c_2} e^{-c \delta^2 n^{1 + 2\gamma}} \\
    &= 0.    
\end{align*}

\subsection{Proof of Theorem \ref{thm:layer_asymp}}\label{appx51}
\setcounter{equation}{0}
\renewcommand{\theequation}{B.\arabic{equation}}

Define the Hajek projection of $\hat{\nu}_s(z;n,m_s)-\nu_s(z;n)$ as
\begin{equation}
    \begin{split}
      \Tilde{\nu}_s(z;n,m_s)&=\sum_{i=1}^{n-1}\mathbb{E}(\hat{\nu}_s(z;n,m_s)-\nu_s(z;n)|Z_i)-(n-2)\mathbb{E}(\hat{\nu}_s(z;n,m_s)-\nu_s(z;n))\\
    &=\sum_{i=1}^{n-1}\mathbb{E}(\hat{\nu}_s(z;n,m_s)-\nu_s(z;n)|Z_i)\\
    &=\sum_{i=1}^{n-1}\mathbb{E}\left(\frac{1}{m_s}\sum_c h_s(Z_{i_1},\dots,Z_{i_s};z)-\nu_s(z;n)|Z_i\right)\\
    &=\sum_{i=1}^{n-1}\frac{1}{m_s}\sum_c \mathbb{E}(h_s(Z_{i_1},\dots,Z_{i_s};z)-\nu_s(z;n)|Z_i))\\
    &=\sum_{i=1}^{n-1}\frac{1}{m_s}\sum_c h_{1,s}(Z_i;z). 
    \end{split}
    \label{proj}
\end{equation}

The first equation is based on Van der Vaart (2000)~\citep{vande2000}, and the second equation is based on $\mathbb{E}(\hat{\nu}_s(z;n,m_s)-\nu_s(z;n))=0$, and the fourth equation is based on the linearity of expectation.
$\sum_c$ denotes the summation over the $\binom {n-1} {s}$ combinations of $s$ distinct elements $\{Z_{i_1},\dots,Z_{i_s}\}$ from $\{Z_1,\dots,Z_n\}\setminus{z}$.
Let $W$ be the number of subsamples that contain $i$.
Since we assume that the subsamples are selected uniformly at random with replacement,
\begin{equation*}
    W \sim Binom \left(m_s, \frac{\binom{n-2}{s-1}}{\binom{n-1}{s}}\right).
\end{equation*}

So we can rewrite Eq.(\ref{proj}) as 
\begin{equation}
    \begin{split}
         &\sum_{i=1}^{n-1}\frac{1}{m_s}\sum_c \mathbb{E}(h_{1,s}(Z_i;z)|W)\\
    &=\sum_{i=1}^{n-1}\frac{1}{m_s}\mathbb{E}(W h_{1,s}(Z_i;z))\\
    &=\sum_{i=1}^{n-1}\frac{1}{m_s}\left( m_s \left( \frac{\binom{n-2}{s-1}}{\binom{n-1}{s}}\right)h_{1,s}(Z_i;z)\right)\\
    &=\frac{s}{n-1}\sum_{i=1}^{n-1}h_{1,s}(Z_i;z).
    \end{split}
    \label{tildenu}
\end{equation}

Define the triangular array
\begin{equation*}
	\begin{matrix}	
		s_1h_{1,s_1}(Z_1;z) & \dots & s_1h_{1,s_1}(Z_{s_1};z) \\	
		s_2h_{1,s_2}(Z_1;z) & \dots & \dots & s_2h_{1,s_2}(Z_{s_2};z) \\	
		\vdots & & & &\ddots\\
		s_kh_{1,s_k}(Z_1;z) & \dots  &\dots &\dots  & s_kh_{1,s_k}(Z_{s_k};z).\\		
	\end{matrix}
\end{equation*}

For each variable in the array, we have
\begin{equation*}
    \mathbb{E}(s h_{1,s}(Z_i;z))=s\mathbb{E}(h_{1,s}(Z_i;z))=0,
\end{equation*}
and
\begin{equation*}
    \mathrm{Var}(s h_{1,s}(Z_i;z))=s^2 \mathrm{Var}(h_{1,s}(Z_i;z))=s^2 \zeta_{1,s}(z).
\end{equation*}


For $\delta \geq 0$, the Lindeberg condition is 
\begin{equation}
    \begin{split}
        &\lim_{n\rightarrow\infty}\sum_{i=1}^{n-1}\frac{1}{(n-1)s^2\zeta_{1,s}(z)}\left[\int_{|h_{1,s}(Z;z)|\geq \delta\sqrt{(n-1)\zeta_{1,s}(z)}}s^2h^2_{1,s}(Z_i;z)dP\right]\\
        &=\lim_{n\rightarrow\infty}\sum_{i=1}^{n-1}\frac{1}{(n-1)\zeta_{1,s}(z)}\left[\int_{|h_{1,s}(Z;z)|\geq \delta\sqrt{(n-1)\zeta_{1,s}(z)}}h^2_{1,s}(Z_i;z)dP\right]\\
        &\leq \lim_{n\rightarrow\infty} \max_{1\leq i \leq (n-1)}\frac{1}{\zeta_{1,s}(z)}\left[\int_{|h_{1,s}(Z_i;z)|\geq \delta\sqrt{(n-1)\zeta_{1,s}(z)}}h^2_{1,s}(Z_i;z)dP\right]\\
        &=\lim_{n\rightarrow\infty}\frac{1}{\zeta_{1,s}(z)}\int_{|h_{1,s}(Z;z)|\geq \delta\sqrt{(n-1)\zeta_{1,s}(z)}}h^2_{1,s}(Z;z)dP\\
        &=0.
    \end{split}
\end{equation}

The last equation is due to Lemma \ref{lindeberg} and thus the Lindeberg condition is satisfied.
By the Lindeberg-Feller central limit theorem,
\begin{equation*}
    \frac{\sum_{i=1}^{n-1} s h_{1,s}(Z_i;z)}{s \sqrt{(n-1)\zeta_{1,s}(z)}}\stackrel{d}{\rightarrow}\mathcal{N}(0,1).
\end{equation*}
According to Eq.(\ref{tildenu}), 
\begin{equation*}
    s\sum_{i=1}^{n-1}h_{1,s}(Z_i;z)=(n-1)\Tilde{\nu}_s(z;n,m_s).
\end{equation*}
That is 
\begin{equation}
    \frac{\sqrt{n-1}\Tilde{\nu}_s(z;n,m_s)}{\sqrt{s^2\zeta_{1,s}(z)}}\stackrel{d}{\rightarrow}\mathcal{N}(0,1).
\end{equation}

Next, we compare the limiting variance ratio of $\hat{\nu}_s(z;n,m_s)$ and its projection
$\Tilde{\nu}_s(z;n,m_s)$.

From the conclusion of Blom(1976) which shows the relationship of variance of the incomplete U-statistic consisting of $m_s$ subsamples selected uniformly at random with replacement and the variance of the complete U-statistic analog, we obtain 
\begin{align*}
    \mathrm{Var} (\hat{\nu}_s(z;n,m_s))&= \mathrm{Var} \left(\frac{1}{m_s}\sum h(Z_{i_1},\dots,Z_{i_s};z)\right)\\
    &=\frac{\zeta_{s,s}(z)}{m_s}+\left(1-\frac{1}{m_s}\right)\mathrm{Var}(\nu_s(z;B)),
\end{align*}
where 
\begin{align*}  
& \mathrm{Var}(\nu_s(z;B))\\
&=\sum_{c=1}^{s}\frac{s!^2}{c!(s-c)!}\frac{(n-s)(n-s-1)\dots(n-2s+c+1)}{n(n-1)\dots(n-s+1)}\zeta_{c,s}.
\end{align*}

Comparing the limiting variance ratio, we have
\begin{align*}
    &\lim_{n\rightarrow\infty}\left(\frac{\mathrm{Var} (\hat{v}_s(z;n,m_s))}{\mathrm{Var} (\Tilde{v}_s(z;n,m_s))}\right)\\
    &=\lim_{n\rightarrow\infty}\left(\frac{\frac{\zeta_{s,s}(z)}{m_s}+\left(1-\frac{1}{m_s}\right)\mathrm{Var}(\nu_s(z;B))}{\frac{s^2}{n}\zeta_{1,s}(z)}\right)\\
    &=\lim_{n\rightarrow\infty}\left(\frac{n\zeta_{s,s}(z)}{m_ss^2\zeta_{1,s}(z)}\right)+\lim_{n\rightarrow\infty}\left(1-\frac{1}{m_s}\right)\lim_{n\rightarrow\infty}\left(\frac{n \thinspace \mathrm{Var}(\nu_s(z;B))}{s^2\zeta_{1,s}(z)}\right)\\
    &=\lim_{n\rightarrow\infty}\left(\frac{n \thinspace \mathrm{Var}(\nu_s(z;B))}{s^2\zeta_{1,s}(z)}\right)\\
    &=\lim_{n\rightarrow\infty}\left ( \frac{n\sum_{c=1}^{s}\frac{s!^2}{c!(s-c)!}\frac{(n-s)(n-s-1)\dots(n-2s+c+1}{n(n-1)\dots(n-s+1)}\zeta_{c,s}}{s^2\zeta_{1,s}(z)}\right)\\
    &=\lim_{n\rightarrow\infty}\left(\frac{\frac{s!^2}{(s-1)!^2}\frac{(n-s)(n-s-1)\dots(n-2s+1+1)}{(n-1)\dots(n-s+1)}\zeta_{1,s}(z)}{s^2\zeta_{1,s}(z)}\right)\\
    &=\lim_{n\rightarrow\infty}\frac{(n-s)(n-s-1)\dots(n-2s+1+1)}{(n-1)\dots(n-s+1)}\\
    &=1,
\end{align*}
where the third equal sign is because $\lim_{n\rightarrow\infty}\frac{n}{m_s}=0$, 
and $\zeta_{1,s}(z)>C\neq 0$.
The fifth and sixth equal sign is because for $s\sim n^{\gamma}, \gamma\in(0,1/2)$, we have $\lim \frac{s}{n}=0$.

By Slutsky's Theorem and Theorem 11.2 in Van der Vaart~\citep{vande2000}, we have
\begin{align*}
    &\frac{\sqrt{n-1}\left( \hat{\nu}_s(z;n,m_s)-\nu_s(z;n)\right)}{\sqrt{s^2\zeta_{1,s}(z)}}\\&=\frac{\sqrt{n-1}(\hat{\nu}_s(z;n,m_s)-\nu_s(z;n)+\Tilde{\nu}_s(z;n,m_s)-\Tilde{\nu}_s(z;n,m_s))}{\sqrt{s^2\zeta_{1,s}(z)}}\\
    &=\frac{\sqrt{n-1}(\hat{\nu}_s(z;n,m_s)-\nu_s(z;n)-\Tilde{\nu}_s(z;n,m_s))}{\sqrt{s^2\zeta_{1,s}(z)}}+\frac{\sqrt{n-1}\Tilde{\nu}_s(z;n,m_s)}{\sqrt{s^2\zeta_{1,s}(z)}},
\end{align*}
where 
\begin{align*}
    \frac{\sqrt{n-1}(\hat{\nu}_s(z;n,m_s)-\nu_s(z;n)-\Tilde{\nu}_s(z;n,m_s))}{\sqrt{s^2\zeta_{1,s}(z)}}\stackrel{P}{\rightarrow} 0
\end{align*}
and
\begin{align*}
    \frac{\sqrt{n-1}\Tilde{\nu}_s(z;n,m_s)}{\sqrt{s^2\zeta_{1,s}(z)}}\stackrel{d}{\rightarrow} \mathcal{N}(0,1),
\end{align*}
Therefore,
\begin{equation*}
    \frac{\sqrt{n-1}\left( \hat{\nu}_s(z;n,m_s)-\nu_s(z;n)\right)}{\sqrt{s^2\zeta_{1,s}(z)}}\stackrel{d}{\rightarrow} \mathcal{N}(0,1).
\end{equation*}

\subsection{Proof of Theorem \ref{thm:asy}}\label{appx52}
\setcounter{equation}{0}
\renewcommand{\theequation}{C.\arabic{equation}}

We already know
\begin{equation*}
    \sqrt{n-1}( \hat{\nu}_s(z;n,m_s)-\nu_s(z;n))\stackrel{d}{\rightarrow}\mathbb{N}(0,s^2\zeta_{1,s}(z)).
\end{equation*}
Since the sampling of each layer $s$ is independent, the variance of $\nu(z;n,n^\gamma)=\frac{1}{n^\gamma}\sum_{s=0}^{n^\gamma-1}\mathbb{E}_{\mathcal{D}^{s}}[\nu_s(z;B)]$ is
\begin{equation*}
\mathrm{Var}(\nu(z;n,n^\gamma))=\frac{1}{(n^\gamma)^2}\sum_{s=0}^{n^\gamma-1}s^2\zeta_{1,s}(z).
\end{equation*}
Hence,
\begin{align*}
    &\quad\frac{\sqrt{n-1}}{\sqrt{\frac{1}{(n^\gamma)^2}\sum_{s=0}^{n^\gamma-1}s^2\zeta_{1,s}(z)}}(\hat{\nu}(z;n,n^\gamma,m)-\nu(z;n))\\
    &=\frac{\sqrt{n-1}}{\sqrt{\frac{1}{(n^\gamma)^2}\sum_{s=0}^{n^\gamma-1}s^2\zeta_{1,s}(z)}}(\hat{\nu}(z;n,n^\gamma,m)- \nu(z;n,n^\gamma))+\frac{\sqrt{n-1}}{\sqrt{\frac{1}{(n^\gamma)^2}\sum_{s=0}^{n^\gamma-1}s^2\zeta_{1,s}(z)}}(\nu(z;n,n^\gamma)-\nu(z;n))\\
    &=\frac{\sqrt{n-1}}{\sqrt{\frac{1}{(n^\gamma)^2}\sum_{s=0}^{n^\gamma-1}s^2\zeta_{1,s}(z)}}\frac{1}{n^\gamma}\sum_{s=0}^{n^\gamma-1}( \hat{\nu}_s(z;n,m_s)-\nu_s(z;n))+\frac{\sqrt{n-1}}{\sqrt{\frac{1}{(n^\gamma)^2}\sum_{s=0}^{n^\gamma-1}s^2\zeta_{1,s}(z)}}(\nu(z;n,n^\gamma)-\nu(z;n)).
\end{align*}
Now we check the Lindeberg condition. That is, we want to check whether the following equation holds
for any $\epsilon\geq 0$.
\begin{align*} 
   &\lim_{n\rightarrow \infty}\frac{1}{\sum_{s=0}^{n^\gamma-1}s^2\zeta_{1,s}(z)}\sum_{s=0}^{n^\gamma-1}\mathbb{E}((\sqrt{n-1}(\hat{\nu}_s-\nu_s))^2  I\{\left|\sqrt{n-1}(\hat{\nu}_s-\nu_s)\right|\geq \epsilon\sqrt{\sum_{s=0}^{n-1}s^2\zeta_{1,s}(z)}\})=0.
\end{align*} 
Since $\nu_s(z;n)$ is bound, we assume $\nu_s(z;n) \leq \delta$, then based on Hoeffding's inequality, we have
\begin{align*}
  &\quad\mathbb{P}\left(|\sqrt{n-1}(\hat{\nu}_s-\nu_s)|\geq \epsilon\sqrt{\sum_{s=0}^{n-1}s^2\zeta_{1,s}(z)}\right)\\
    &\leq 2 \exp \left(-\frac{\epsilon^2\sum_{s=0}^{n-1}s^2\zeta_{1,s}(z)}{2\delta^2(n-1)}\right).
\end{align*}
Since $\zeta_{1,s}\geq C\neq 0$, we know
\begin{equation*}
    \sum_{s=0}^{n-1}s^2\zeta_{1,s}\geq c'n^3, \ c'>0.
\end{equation*}
Then the left side of Lindeberg condition can be scaled as
\begin{equation*}
     %
     \frac{1}{n^2}\exp (-c''\epsilon^2n^2)\rightarrow 0,\ c''>0.
\end{equation*}
So the Lindebeg condition is satisfied.

According to the Lindeberg-Feller Central Limit Theorem,
\begin{equation}
\frac{\sum_{s=0}^{n^\gamma-1}\sqrt{n-1}( \hat{\nu}_s(z;n,m_s)-\nu_s(z;n))}{\sqrt{\sum_{s=0}^{n^\gamma-1}s^2\zeta_{1,s}(z)}}
    \stackrel{d}{\rightarrow}\mathcal{N}(0,1).
    \label{Fnormal}
\end{equation}

For the second part, we have

\begin{equation}
    \begin{split}
    &\frac{\sqrt{n-1}}{\sqrt{\frac{1}{(n^\gamma)^2}\sum_{s=0}^{n^\gamma-1}s^2\zeta_{1,s}(z)}}(\nu(z;n,n^\gamma)-\nu(z;n))\\
    &=\frac{\sqrt{n-1}}{\sqrt{\frac{1}{(n^\gamma)^2}\sum_{s=0}^{n^\gamma-1}s^2\zeta_{1,s}(z)}}\left|\frac{1}{n^\gamma}\sum_{s=0}^{n^\gamma-1}\nu_s(z;n)-\frac{1}{n}\sum_{s=0}^{n-1}\nu_s(z;n)\right|\\
    &=\frac{\sqrt{n-1}}{\sqrt{\frac{1}{(n^\gamma)^2}\sum_{s=0}^{n^\gamma-1}s^2\zeta_{1,s}(z)}}\left|\sum_{s=0}^{n^\gamma-1}\left(\frac{1}{n^\gamma}-\frac{1}{n}\right)\nu_s(z;n)-\frac{1}{n}\sum_{s=n^\gamma}^{n-1}\nu_s(z;n)\right|.
    \end{split}
    \label{nuP}
\end{equation}

Since $\nu_s(z;n)=\mathbb{E}h_s(Z_{1_1},\dots,Z_{i_s};z)\leq \beta(s)= O(\frac{1}{s+1})$, we have
\begin{equation}
\begin{split}
     &\frac{\sqrt{n-1}}{\sqrt{\frac{1}{(n^\gamma)^2}\sum_{s=0}^{n^\gamma-1}s^2\zeta_{1,s}(z)}}(\nu(z;n,n^\gamma)-\nu(z;n))\\
     &\leq \left|\frac{c\sqrt{n-1}(n-n^r)}{\sqrt{\frac{1}{(n^\gamma)^2}\sum_{s=0}^{n^\gamma-1}s^2\zeta_{1,s}(z)}n^\gamma n}\sum_{s=0}^{n^\gamma-1}\frac{1}{s+1}\right|+\left|\frac{c\sqrt{n-1}}{n\sqrt{\frac{1}{(n^\gamma)^2}\sum_{s=0}^{n^\gamma-1}s^2\zeta_{1,s}(z)}}\sum_{s=n^\gamma}^{n-1}\frac{1}{s+1}\right|.
\end{split}
\end{equation}

We note that $\sum_{j=0}^{n-1}\frac{1}{j+1}\le \log n+1$, and assume $ \zeta_{1,s}\geq c\neq 0$, then $\sqrt{\frac{1}{(n^\gamma)^2}\sum_{s=0}^{n^\gamma-1}s^2\zeta_{1,s}} \sim O((n^\gamma)^{\frac{1}{2}})$, so

\begin{equation}
\begin{split}
     &\frac{\sqrt{n-1}}{\sqrt{\frac{1}{(n^\gamma)^2}\sum_{s=0}^{n^\gamma-1}s^2\zeta_{1,s}(z)}}(\nu(z;n,n^\gamma)-\nu(z;n))\\
     &\leq \left|\frac{c\sqrt{n-1}(n-n^\gamma)(\log n^\gamma+1)}{n^{\frac{\gamma}{2}}n^\gamma n}\right|+\left|\frac{c\sqrt{n-1}(\log n- \log n^\gamma)}{n^{\frac{\gamma}{2}}n}\right|.
\end{split}
\end{equation}

As long as $\gamma\geq \frac{1}{3}$, 
\begin{equation}
    \frac{\sqrt{n-1}}{\sqrt{\frac{1}{(n^\gamma)^2}\sum_{s=0}^{n^\gamma-1}s^2\zeta_{1,s}(z)}}(\nu(z;n,n^\gamma)-\nu(z;n))\stackrel{P}{\rightarrow} 0.
    \label{zeroexp}
\end{equation}


Since Eq.(\ref{Fnormal}) and Eq.(\ref{zeroexp}), according to Slutsky's theorem, we have 
\begin{equation*}
    \frac{\sqrt{n-1}(\hat{\nu}(z;n,n^\gamma,m)- \nu(z;n))}{\sqrt{\frac{1}{(n^\gamma)^2}\sum_{s=0}^{n^\gamma-1}s^2\zeta_{1,s}(z)}}\stackrel{d}{\rightarrow}\mathcal{N}(0,1).
\end{equation*}

\subsection{Additional Experimental Results}\label{exp}
We corroborate our theoretical findings through empirical validation across diverse datasets, as detailed in Table \ref{Data_sets}.
\begin{table*}[t]
    \caption{Data sets}
    \label{Data_sets}
    \centering 
    \begin{tabular*}{\textwidth}{@{\extracolsep{\fill}}p{3cm} p{10cm}} 
        \toprule
        Data set & Reference \\
        \midrule
        Covertype & \citep{misc_covertype_31} \\
        FashionMNIST & \citep{xiao2017fashion} \\
        Creditcard & \citep{yeh2009comparisons} \\
        Vehicle & \citep{duarte2004vehicle} \\
        Apsfail & \url{https://archive.ics.uci.edu/ml/datasets/IDA2016Challenge} \\
        Phoneme & \url{https://sci2s.ugr.es/keel/dataset.php?cod=105} \\
        Wind & \url{https://www.openml.org/search?type=data&status=any&id=847} \\
        Pol & \url{https://www.openml.org/search?type=data&status=any&id=722} \\
        Cpu & \url{https://www.openml.org/search?type=data&status=any&id=796} \\
        2Dplanes & \url{https://www.openml.org/search?type=data&active&id=727} \\
        \bottomrule
    \end{tabular*}
\end{table*}

We verify the normality of the values supported by Theorem~\ref{thm:asy} across multiple datasets. 
On eight different datasets, we set the training set size to $100$, with $\gamma=2/3$, and estimate the values. 
We use Double Monte Carlo simulation to calculate the $\zeta_{1,s}$ for the first 10 data points and construct 95\% confidence intervals. Additionally, we check the empirical coverage of $1000$ estimated values and find that the empirical coverage probability is nearly 100\%. 
The first column of Figures \ref{othersets1} and \ref{othersets2} shows the confidence interval graphs for the corresponding datasets, the second column displays the Q-Q plots for the estimated values, and the third column presents the empirical coverage probabilities for these points.

\begin{figure*}[htbp]
    \centering
    \begin{minipage}{\textwidth}
        \begin{minipage}{0.45\textwidth}
            \centering
         \includegraphics[width=\linewidth]{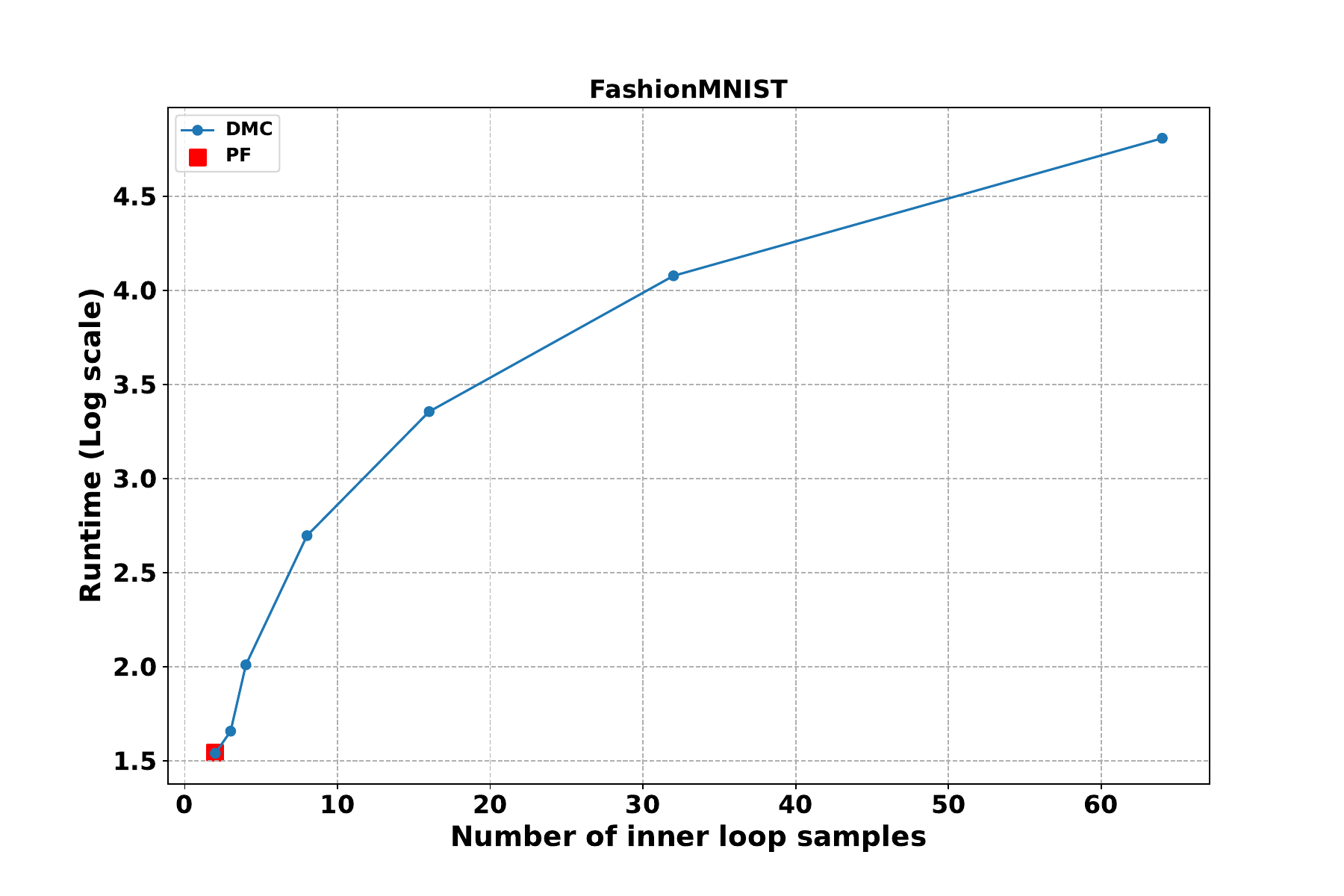}
         \subfloat[]{\label{ilfm1}}
        \end{minipage}%
        \begin{minipage}{0.45\textwidth}
            \centering
            \includegraphics[width=\linewidth]{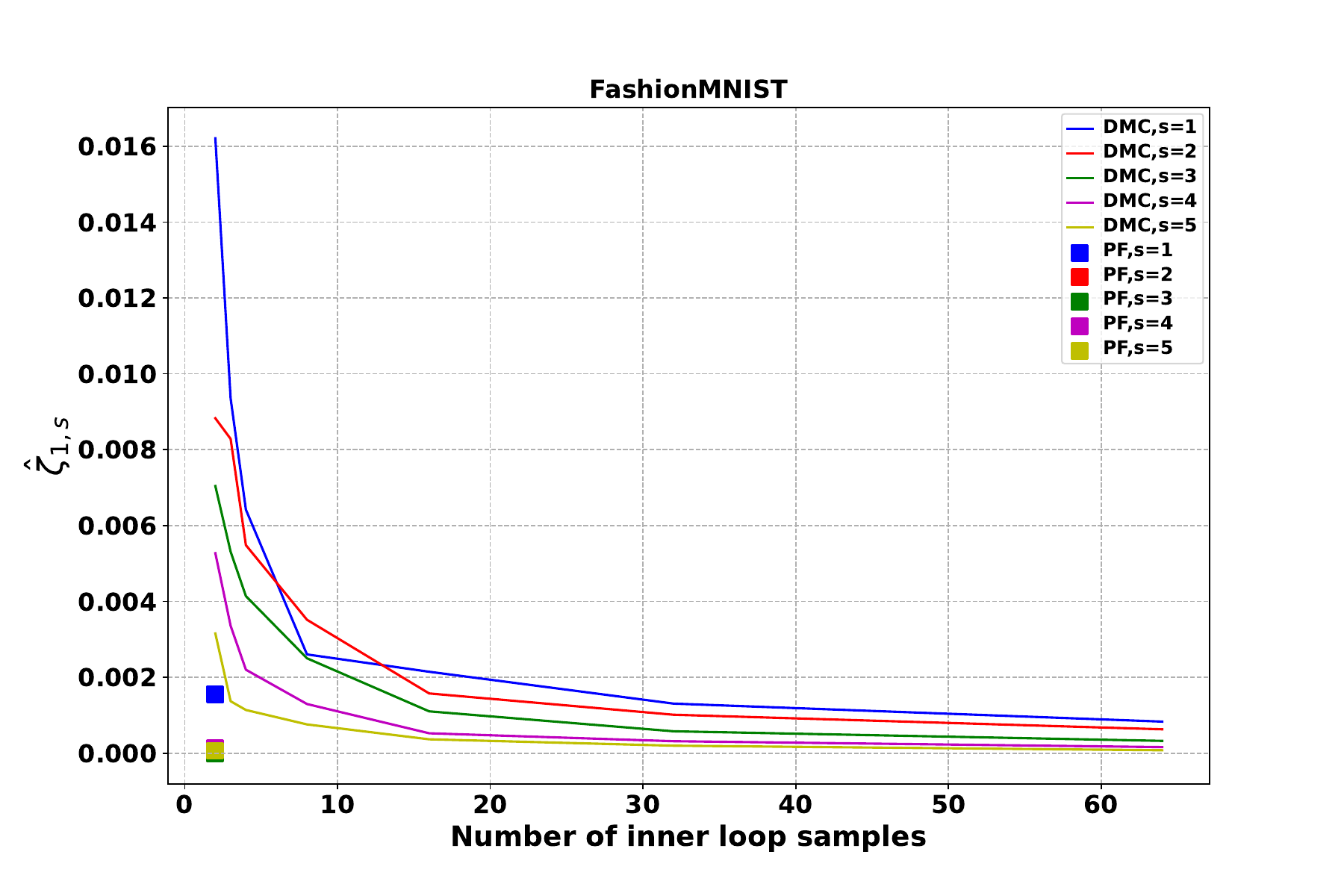}
            \subfloat[]{\label{ilfm2}}
        \end{minipage}
    \end{minipage}
    \caption{Comparison of DMC and PF algorithms on the FashionMNIST dataset. The left panel (Figure \ref{ilfm1}) shows the runtime comparison, with $T=100$ for both algorithms and varying inner loop counts for DMC. The running time of DMC increases rapidly with the number of inner loops, far exceeding that of PF (vertical axis represents the logarithm of time). The right panel (Figure \ref{ilfm2}) shows the estimated values of $\zeta_{1,s}$ using DMC with varying $T_i$. As $T_i$ increases, the estimated values become more stable and accurate.}
    \label{fig:combined}
\end{figure*}

We estimate $\zeta_{1,s}$ using both DMC and PF algorithms on the FashionMNIST datasets and compare their runtime and accuracy. 
Figure \ref{ilfm1} shows the time comparison on the Covertype dataset. 
When $T=100$ for both algorithms and the number of inner loops in DMC is greater than 2, the running time of DMC increases rapidly (the vertical axis in the figure represents the logarithm of time), far exceeding that of PF. 
Figure \ref{ilfm2} shows the estimated values of $\zeta_{1,s}$. 
It can be observed that as the value of $T_i$ increases, the estimated values tend to become more stable, indicating increased accuracy.


\begin{figure*}[htbp]
\centering
\subfloat{
\begin{minipage}[b]{.32\linewidth}
\centering
\includegraphics[width=\linewidth]{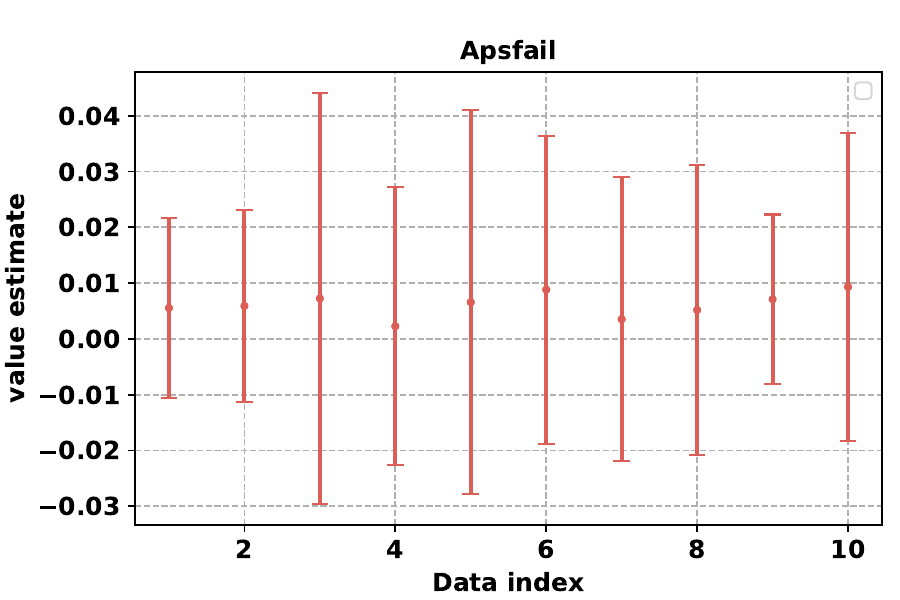}
\end{minipage}
}
\subfloat{
\begin{minipage}[b]{.30\linewidth}
\centering
\includegraphics[width=\linewidth]{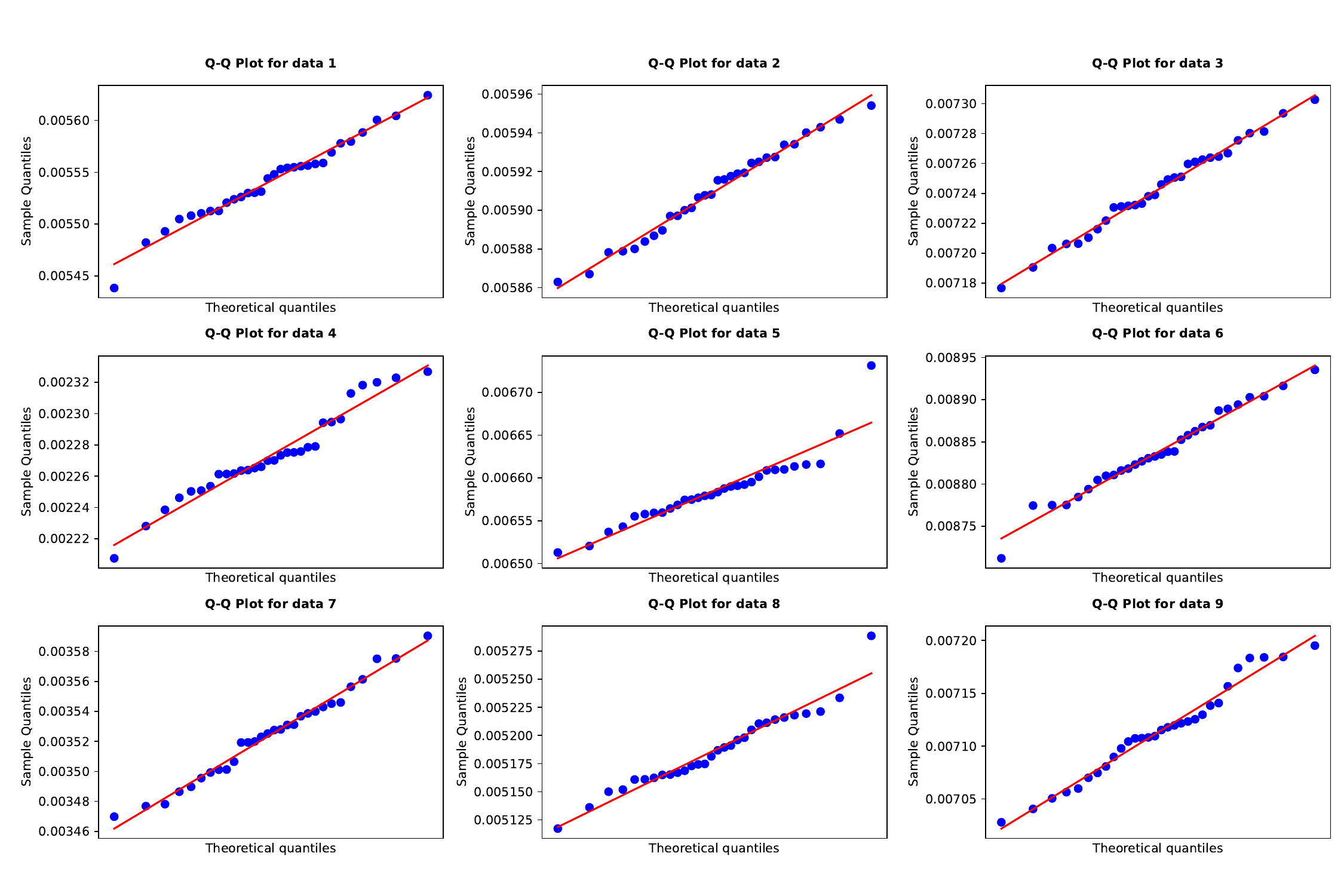}
\end{minipage}
}
\subfloat{
\begin{minipage}[b]{.32\linewidth}
\centering
\includegraphics[width=\linewidth]{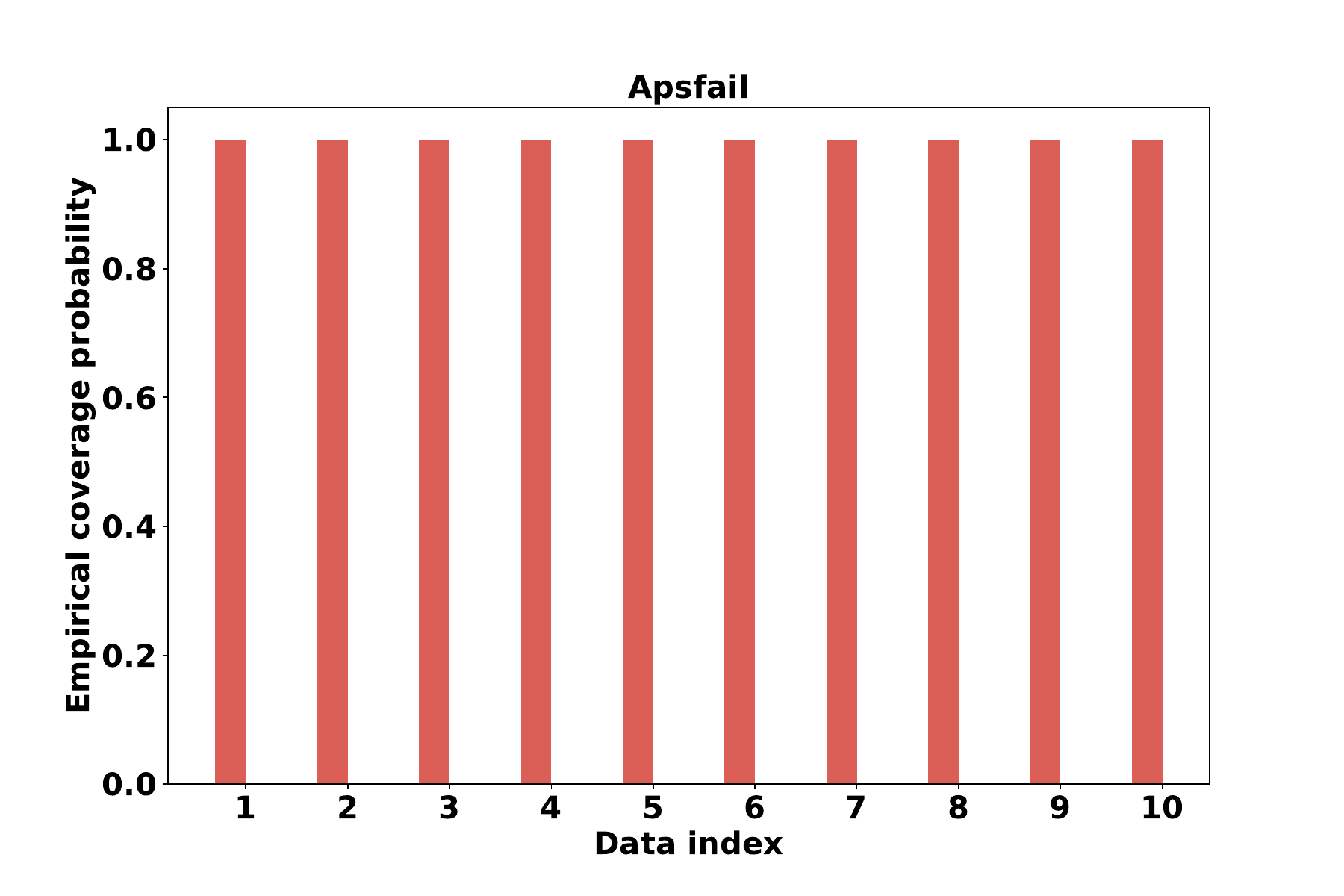}
\end{minipage}
}

\subfloat{
\begin{minipage}[b]{.32\linewidth}
\centering
\includegraphics[width=\linewidth]{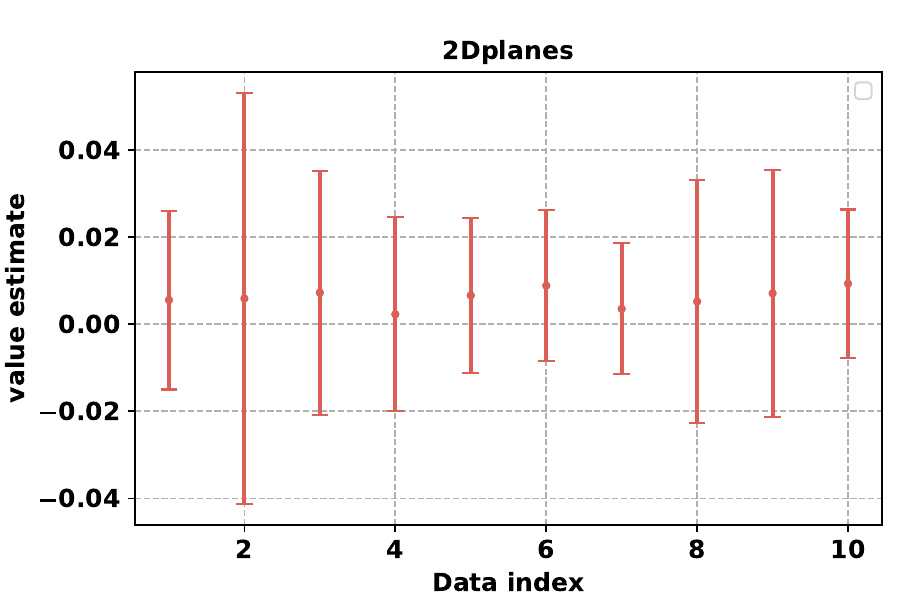}
\end{minipage}
}
\subfloat{
\begin{minipage}[b]{.30\linewidth}
\centering
\includegraphics[width=\linewidth]{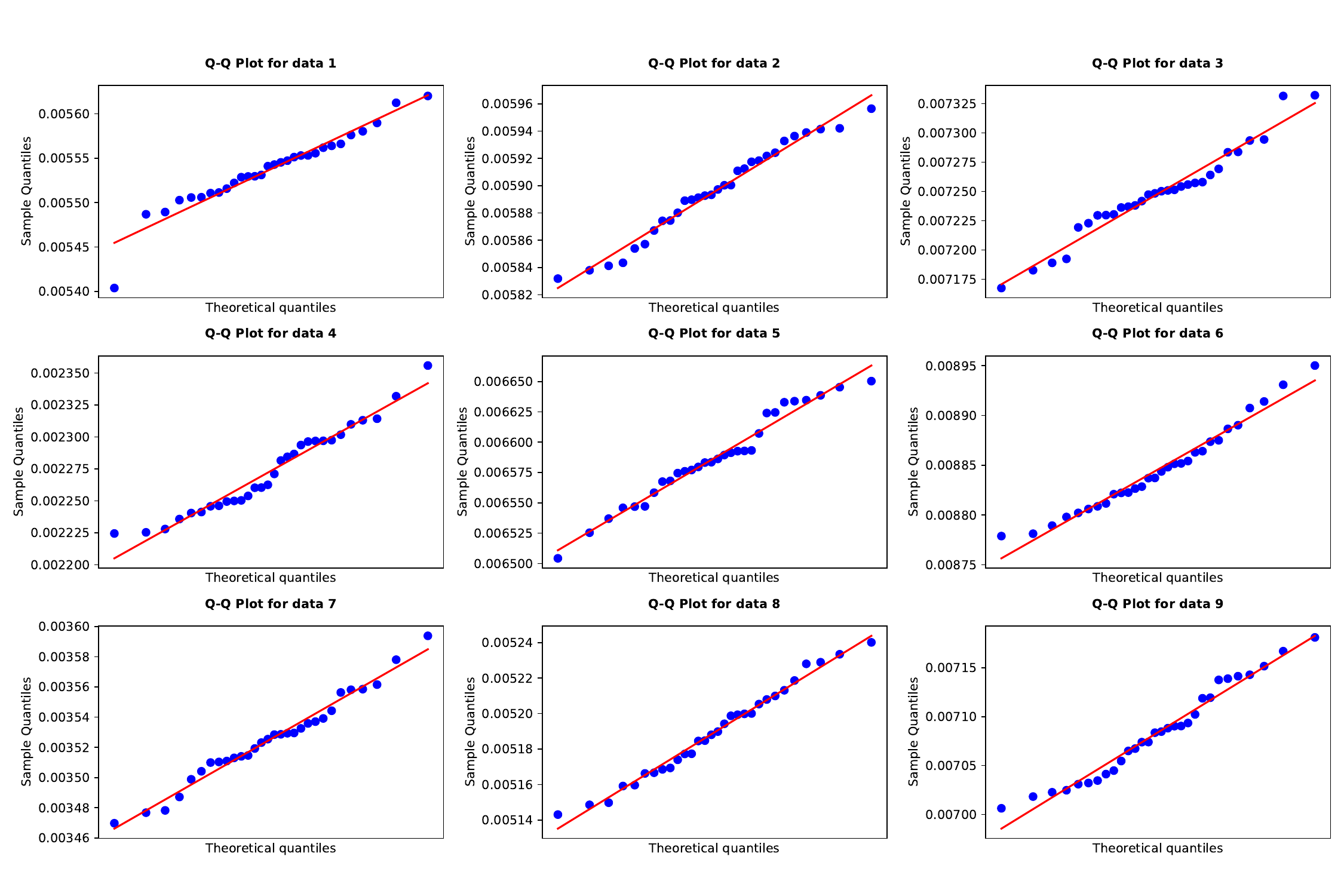}
\end{minipage}
}
\subfloat{
\begin{minipage}[b]{.32\linewidth}
\centering
\includegraphics[width=\linewidth]{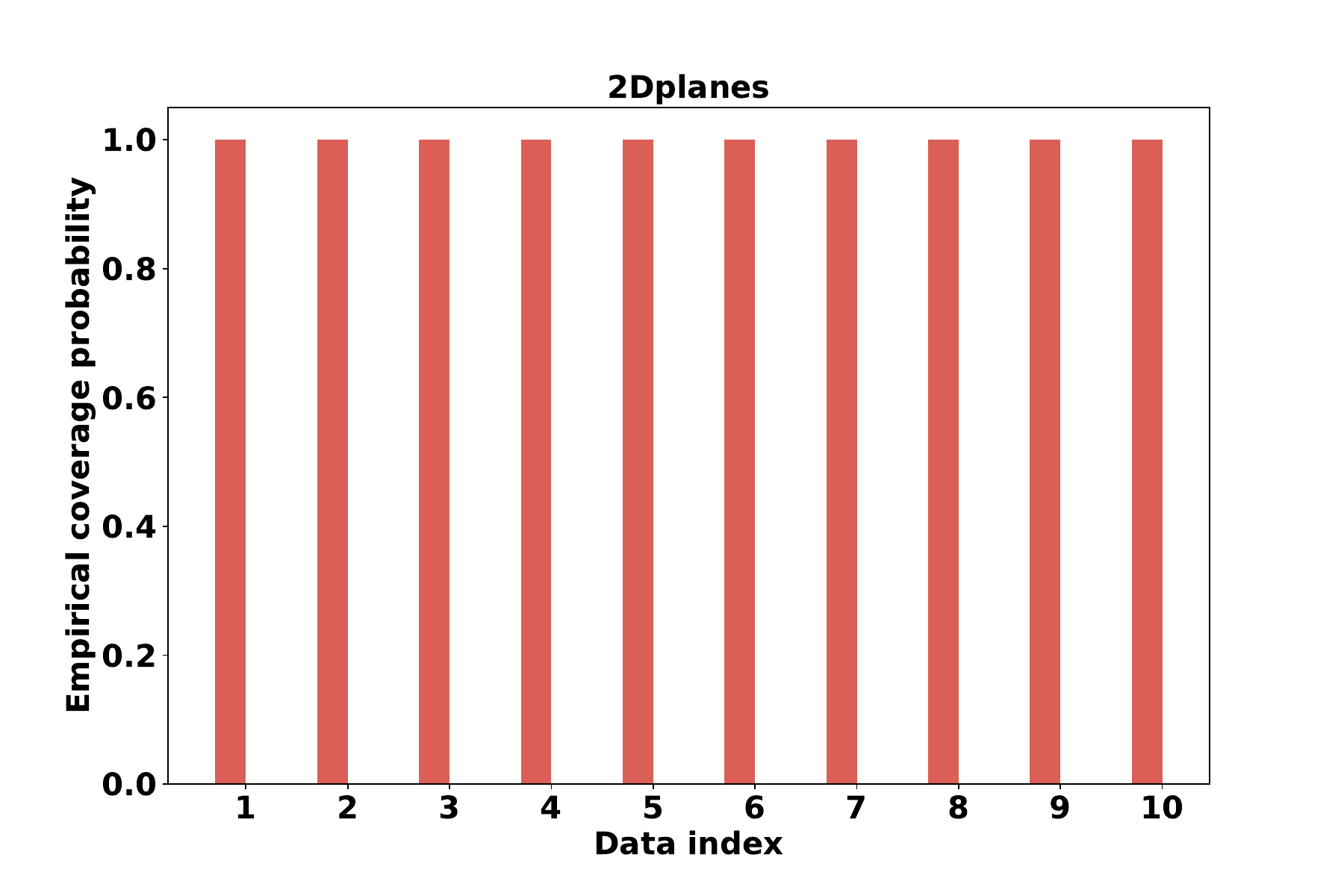}
\end{minipage}
}

\subfloat{
\begin{minipage}[b]{.32\linewidth}
\centering
\includegraphics[width=\linewidth]{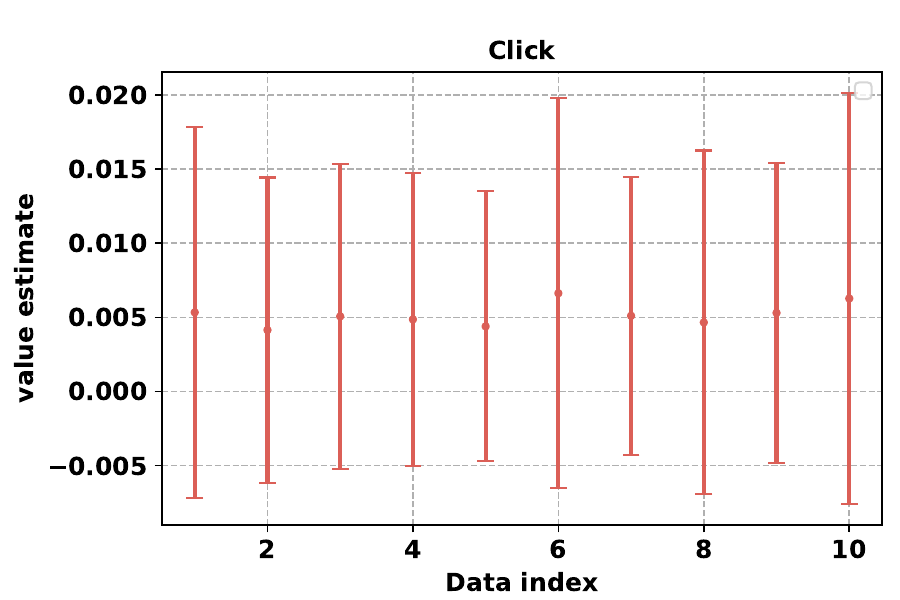}
\end{minipage}
}
\subfloat{
\begin{minipage}[b]{.30\linewidth}
\centering
\includegraphics[width=\linewidth]{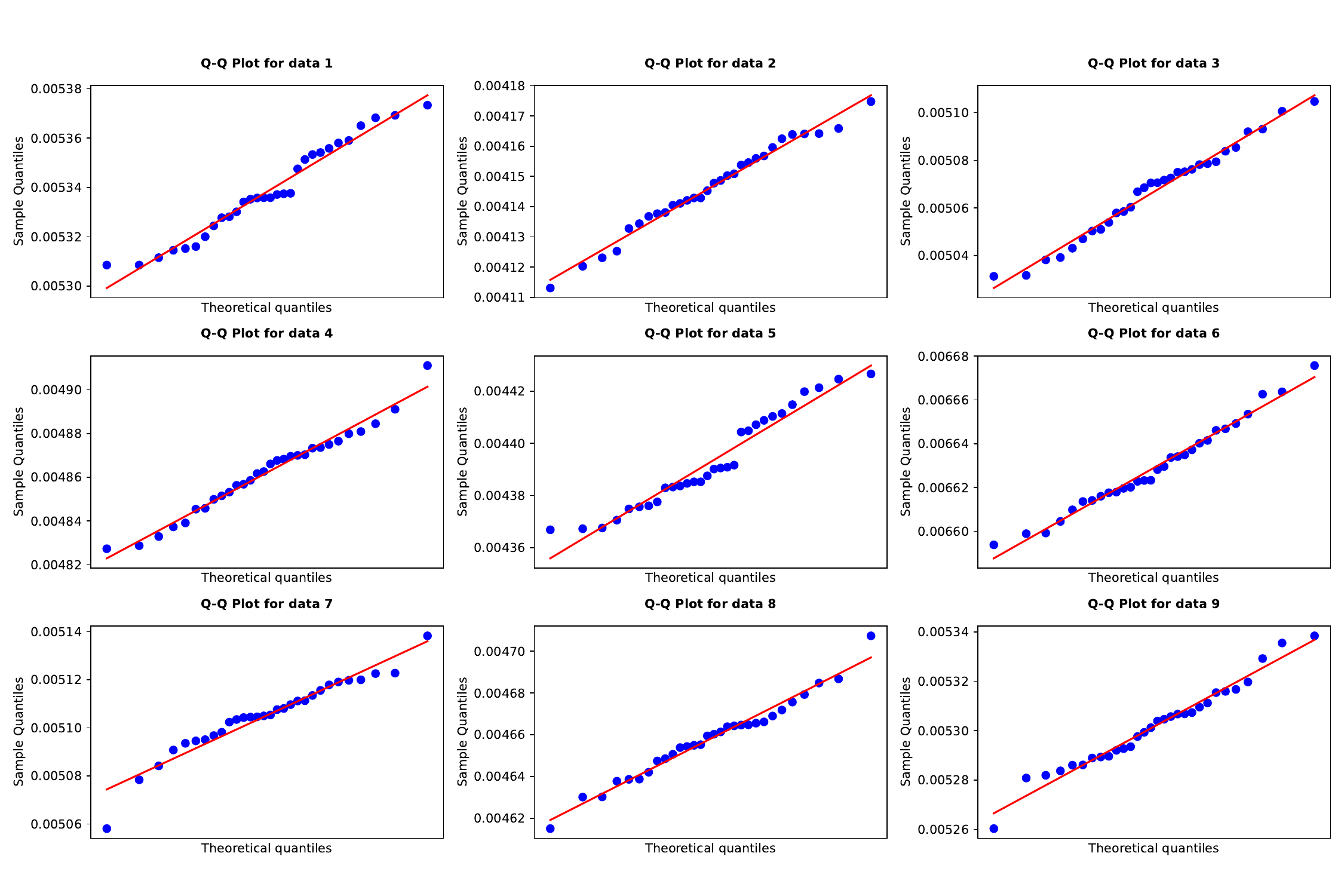}
\end{minipage}
}
\subfloat{
\begin{minipage}[b]{.32\linewidth}
\centering
\includegraphics[width=\linewidth]{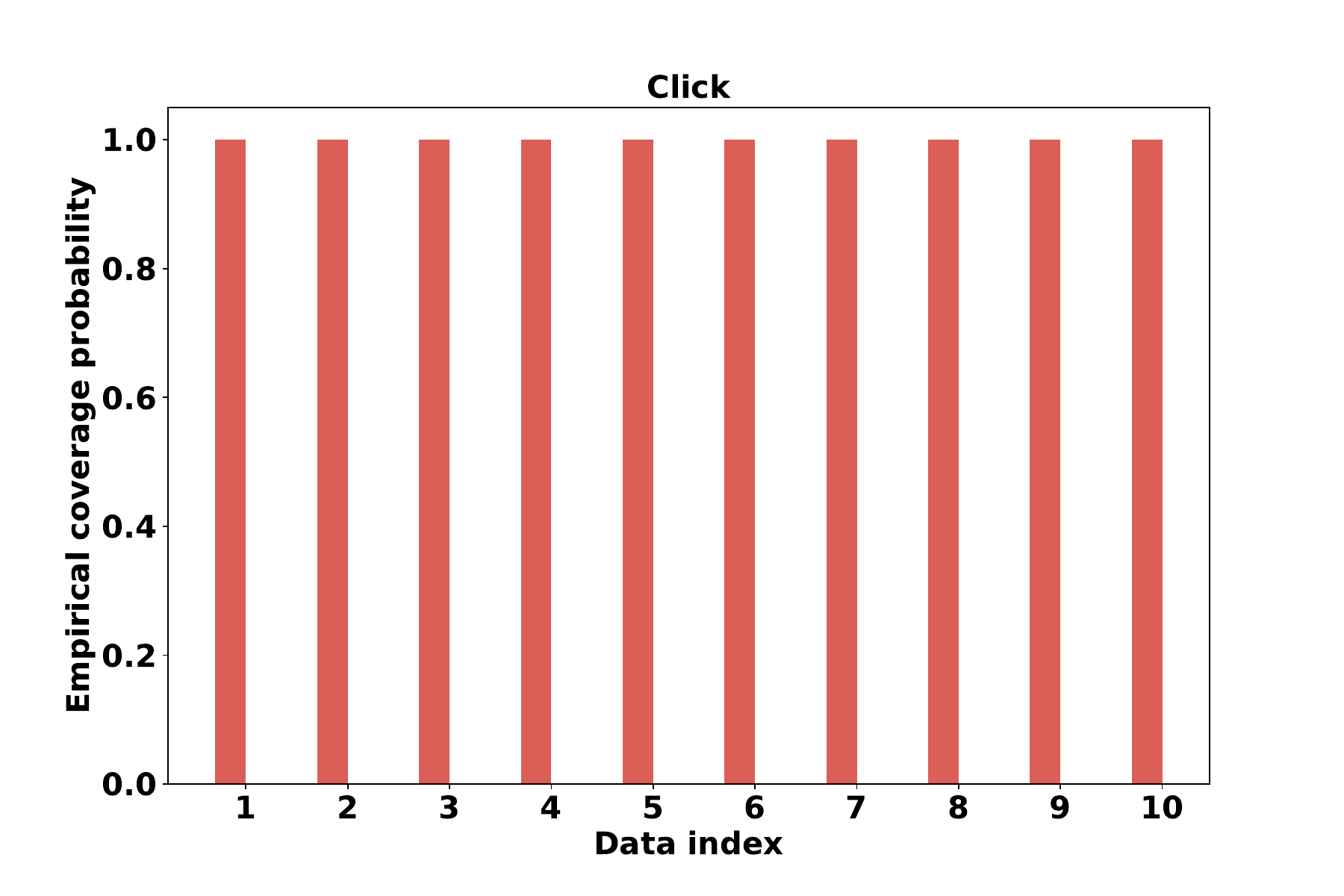}
\end{minipage}
}

\subfloat{
\begin{minipage}[b]{.32\linewidth}
\centering
\includegraphics[width=\linewidth]{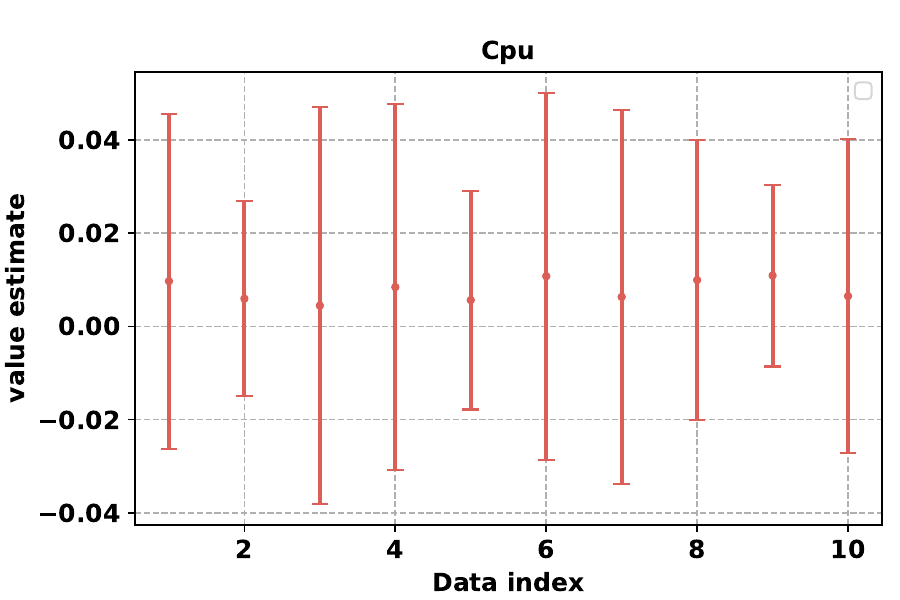}
\end{minipage}
}
\subfloat{
\begin{minipage}[b]{.30\linewidth}
\centering
\includegraphics[width=\linewidth]{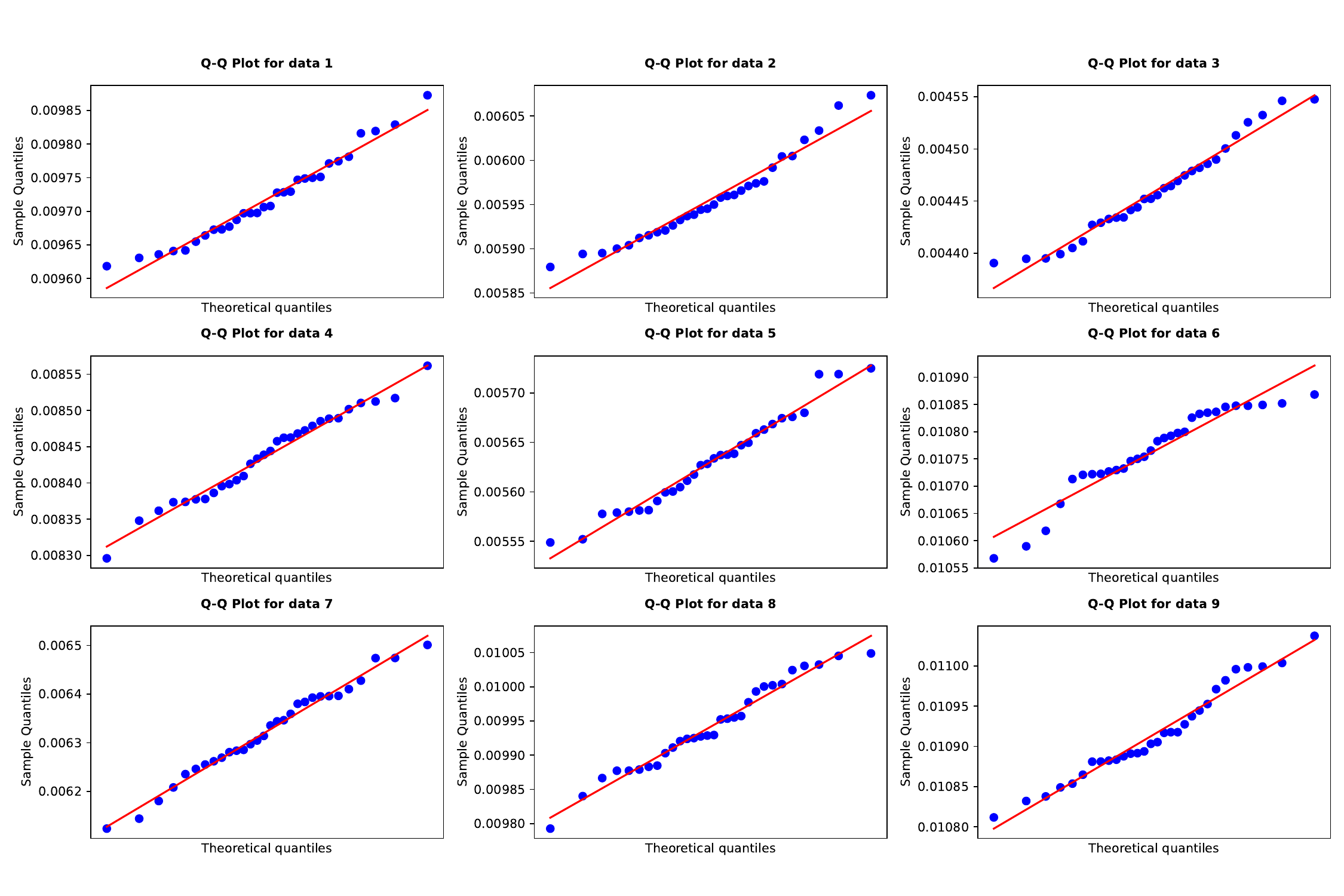}
\end{minipage}
}
\subfloat{
\begin{minipage}[b]{.32\linewidth}
\centering
\includegraphics[width=\linewidth]{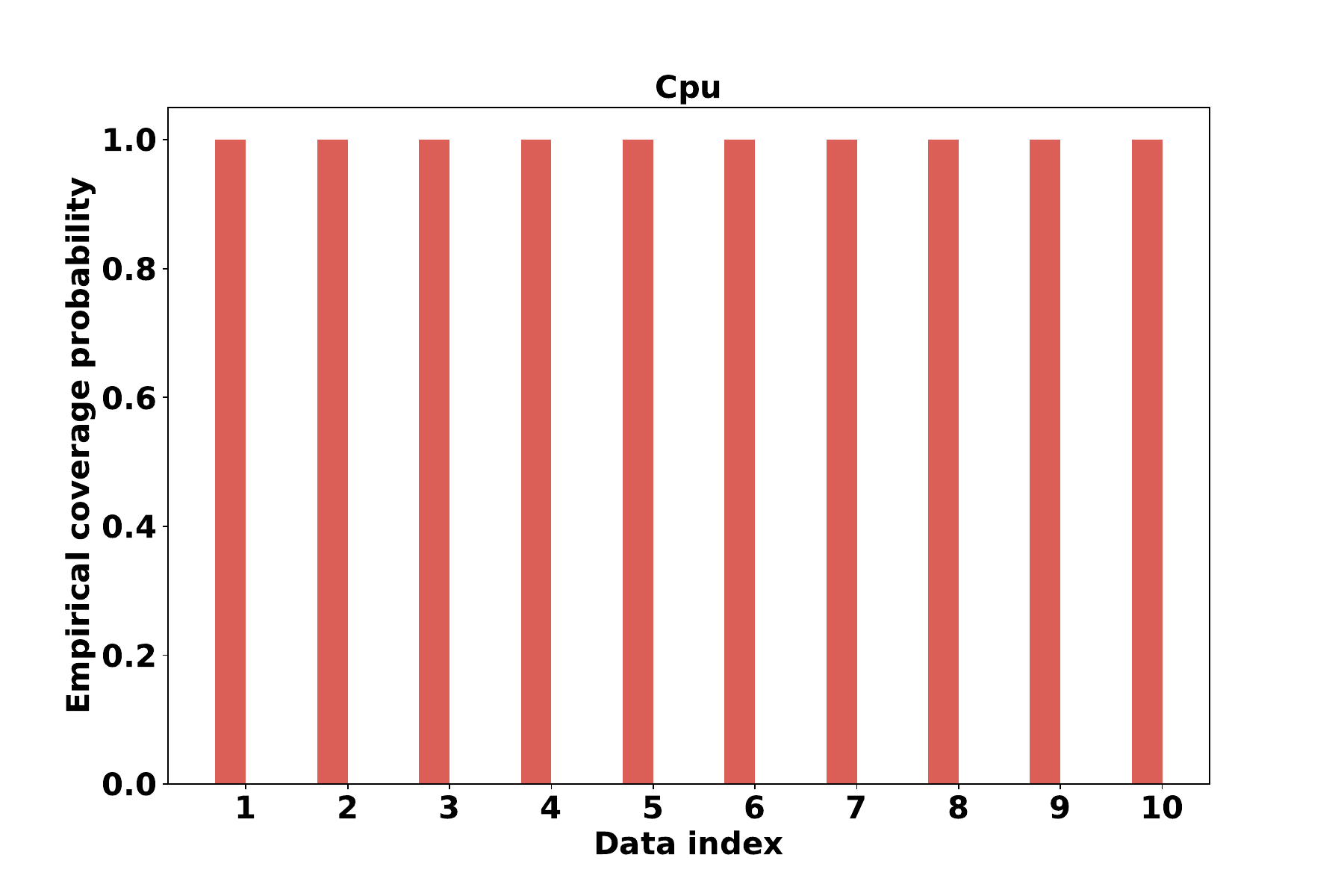}
\end{minipage}
}

\subfloat{
\begin{minipage}[b]{.32\linewidth}
\centering
\includegraphics[width=\linewidth]{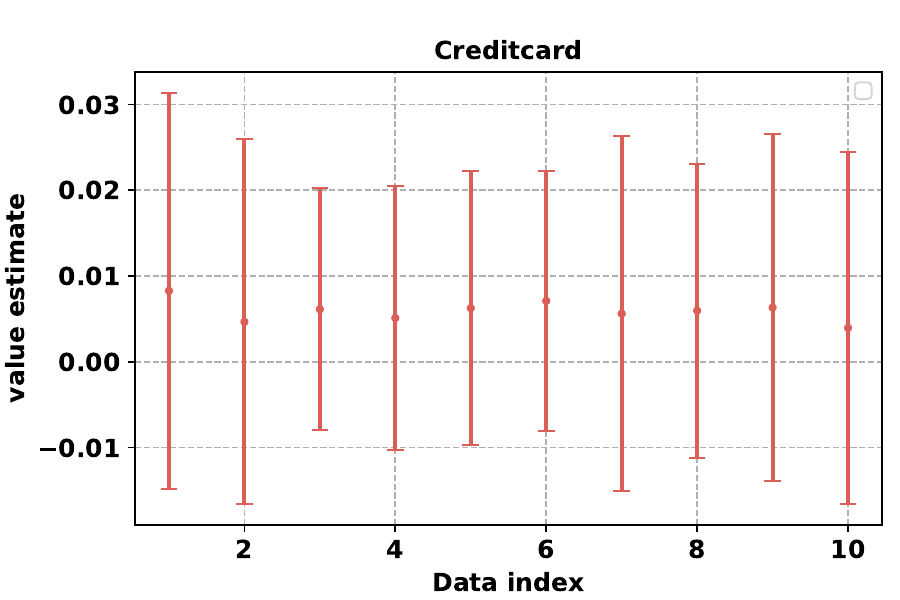}
\end{minipage}
}
\subfloat{
\begin{minipage}[b]{.30\linewidth}
\centering
\includegraphics[width=\linewidth]{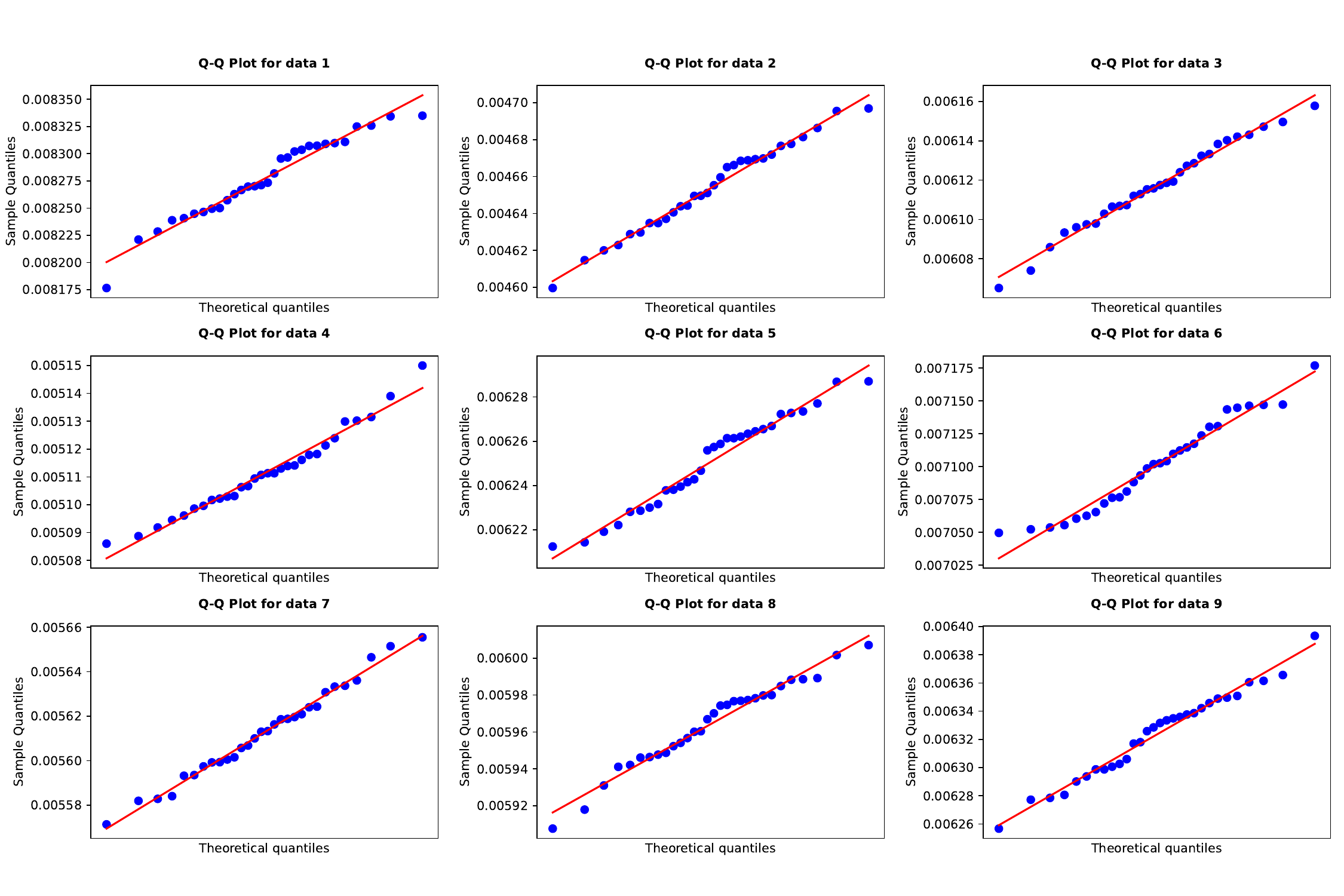}
\end{minipage}
}
\subfloat{
\begin{minipage}[b]{.32\linewidth}
\centering
\includegraphics[width=\linewidth]{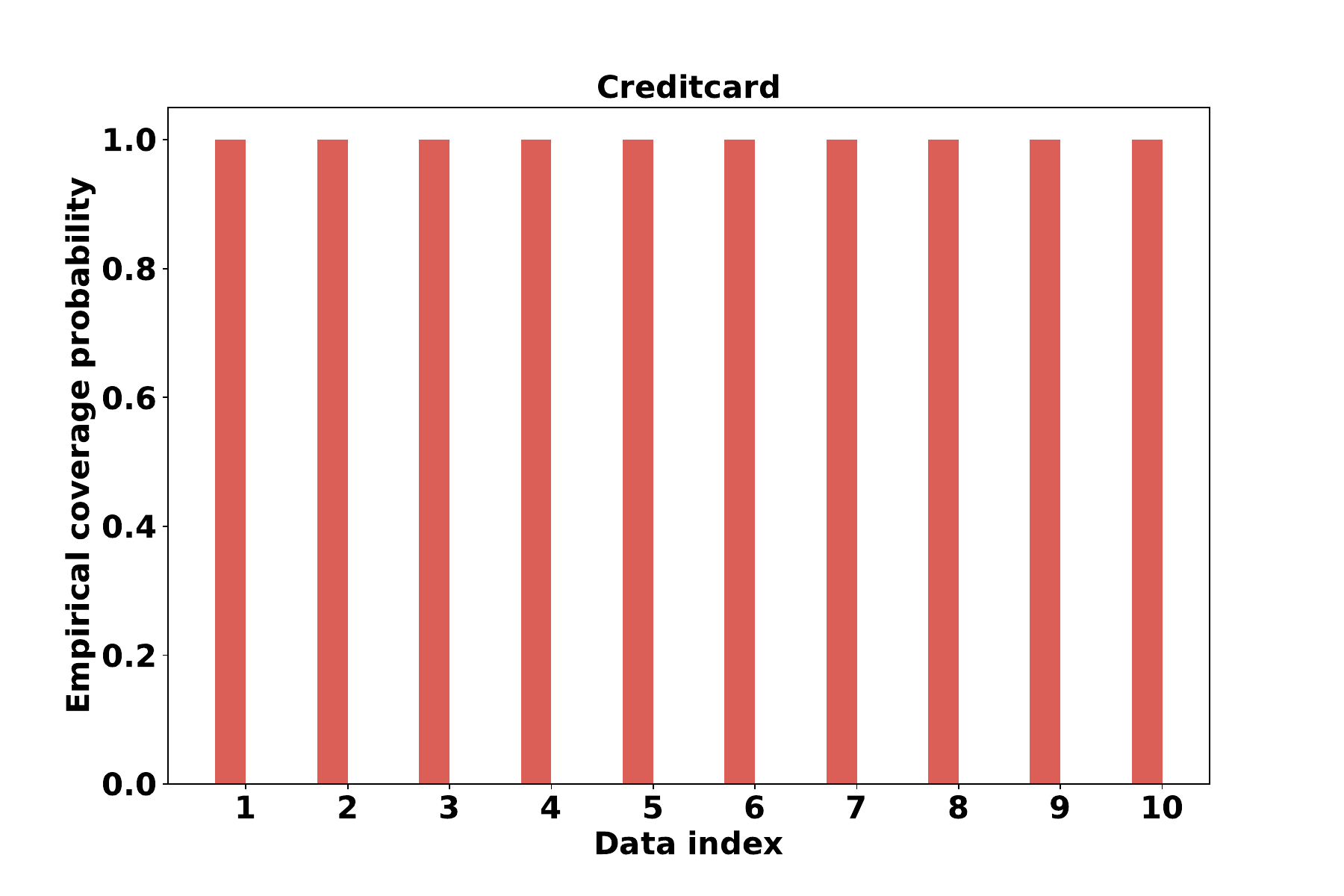}
\end{minipage}
}
\caption{Analysis of estimated values for Apsfail, 2Dplanes, Click, Cpu and Creditcard datasets with a training set size of 100 and $\gamma=2/3$. The first column shows the confidence interval graphs for the datasets. The second column presents the Q-Q plots for the estimated values, and the third column illustrates the empirical coverage probabilities, demonstrating nearly 100\% coverage.}
\label{othersets1}
\end{figure*}

\begin{figure*}[htbp]
\centering
\subfloat{
\begin{minipage}[b]{.32\linewidth}
\centering
\includegraphics[width=\linewidth]{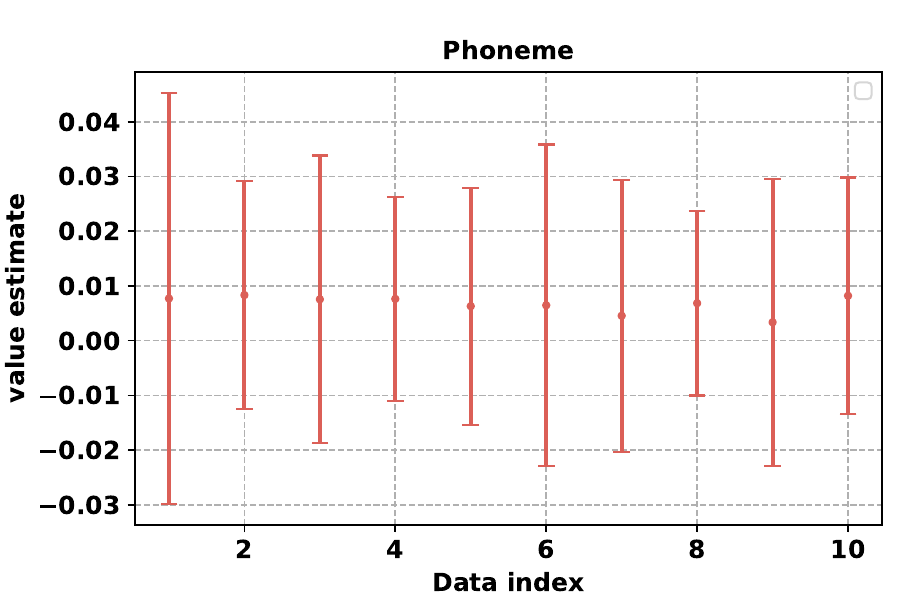}
\end{minipage}
}
\subfloat{
\begin{minipage}[b]{.30\linewidth}
\centering
\includegraphics[width=\linewidth]{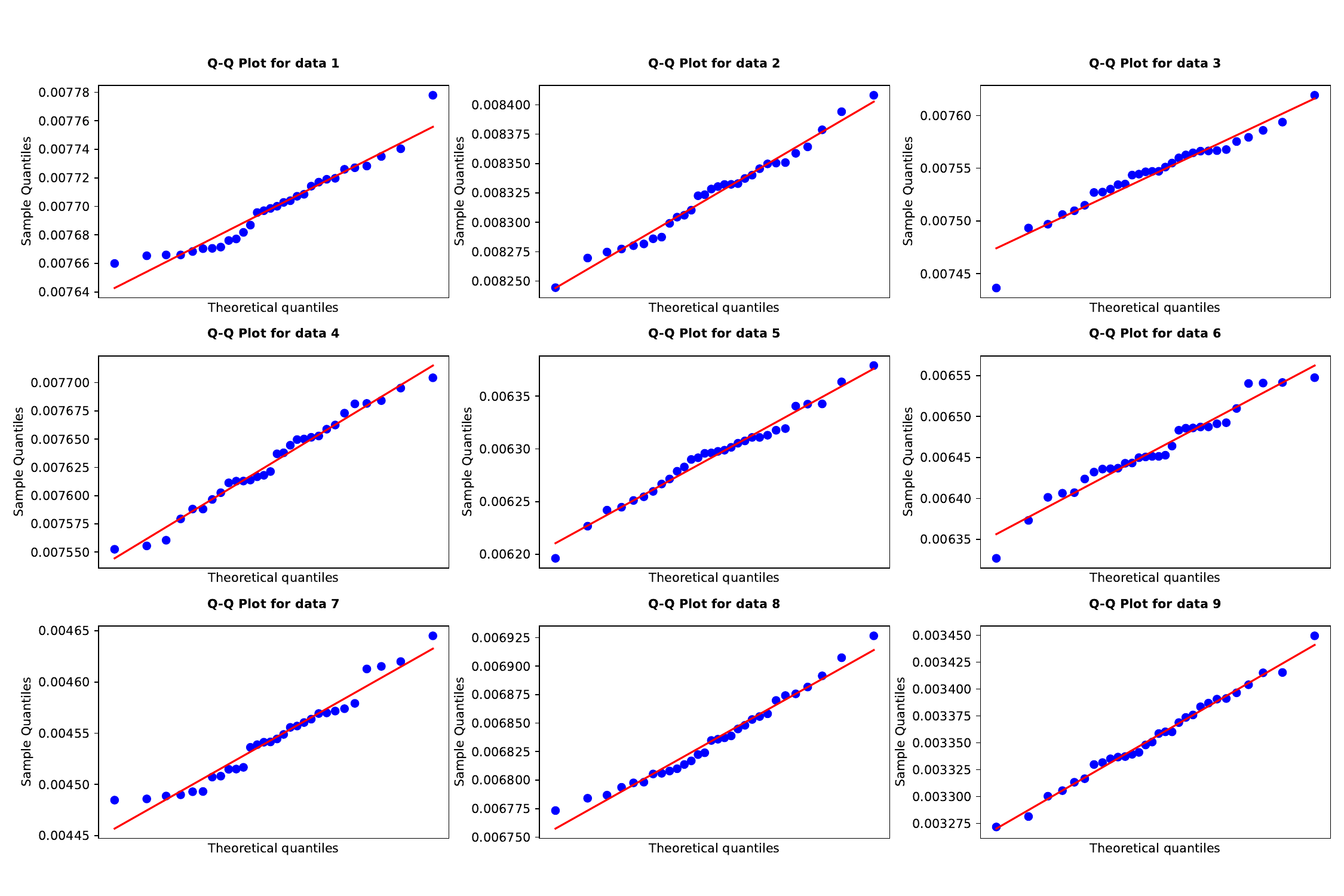}
\end{minipage}
}
\subfloat{
\begin{minipage}[b]{.32\linewidth}
\centering
\includegraphics[width=\linewidth]{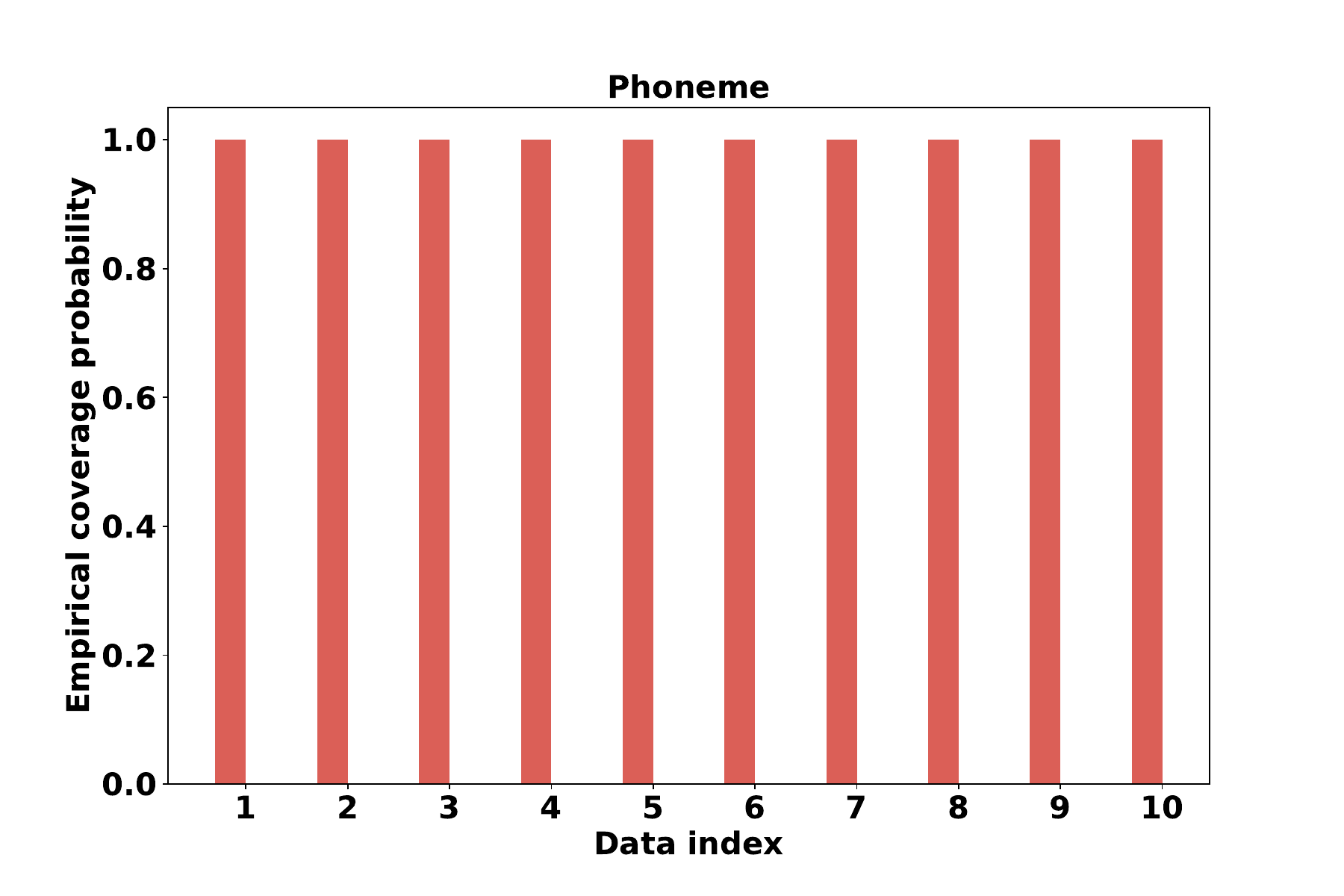}
\end{minipage}
}

\subfloat{
\begin{minipage}[b]{.32\linewidth}
\centering
\includegraphics[width=\linewidth]{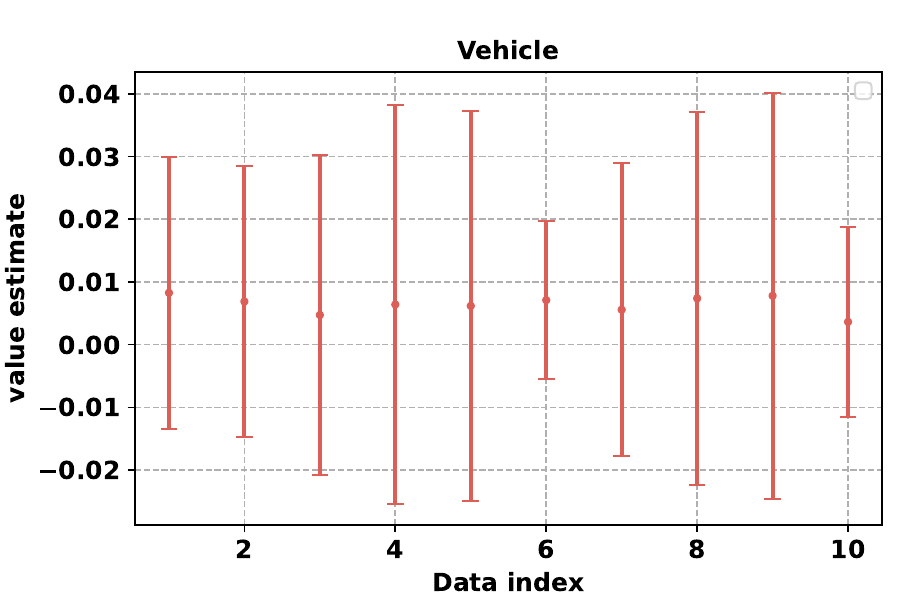}
\end{minipage}
}
\subfloat{
\begin{minipage}[b]{.30\linewidth}
\centering
\includegraphics[width=\linewidth]{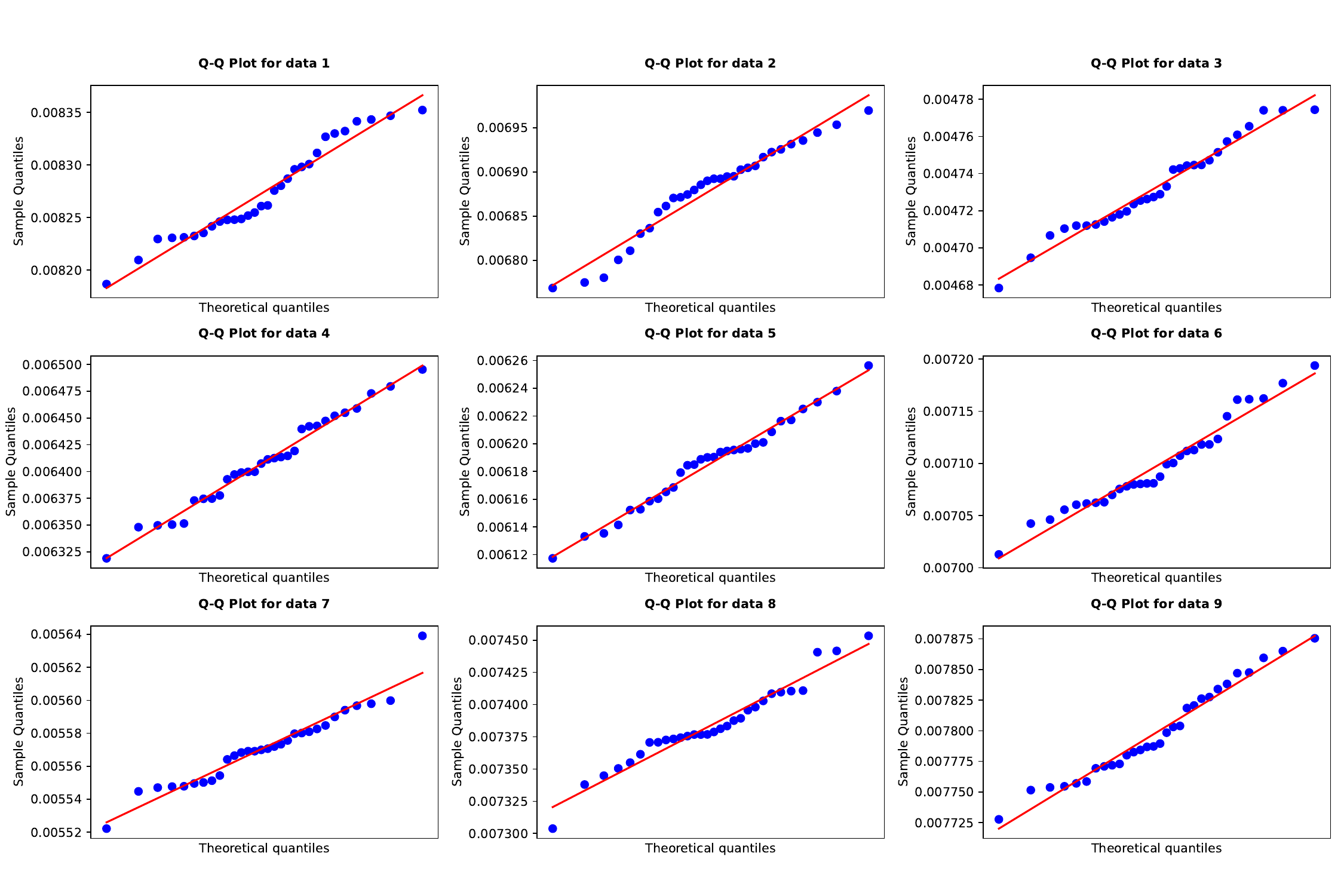}
\end{minipage}
}
\subfloat{
\begin{minipage}[b]{.32\linewidth}
\centering
\includegraphics[width=\linewidth]{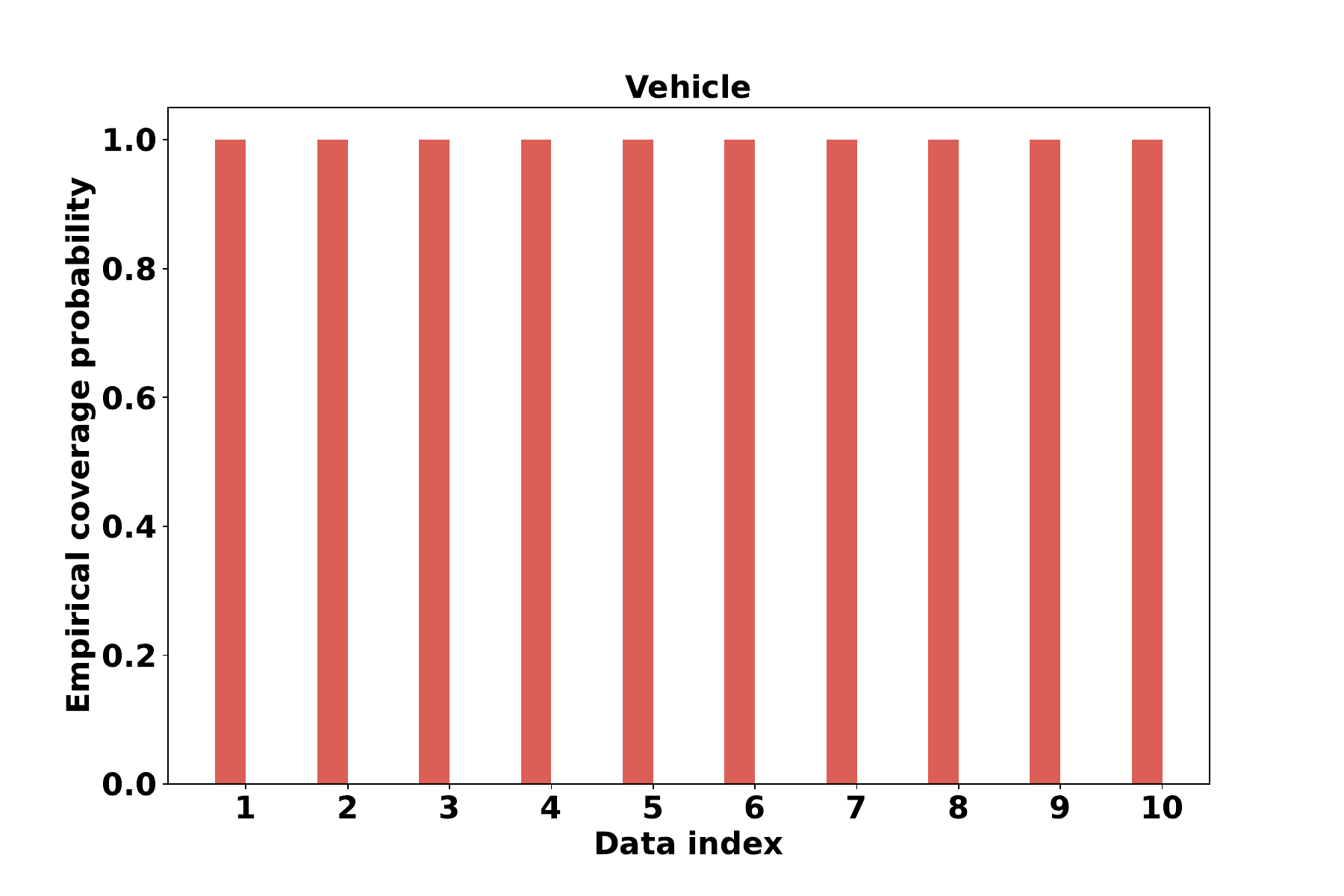}
\end{minipage}
}

\subfloat{
\begin{minipage}[b]{.32\linewidth}
\centering
\includegraphics[width=\linewidth]{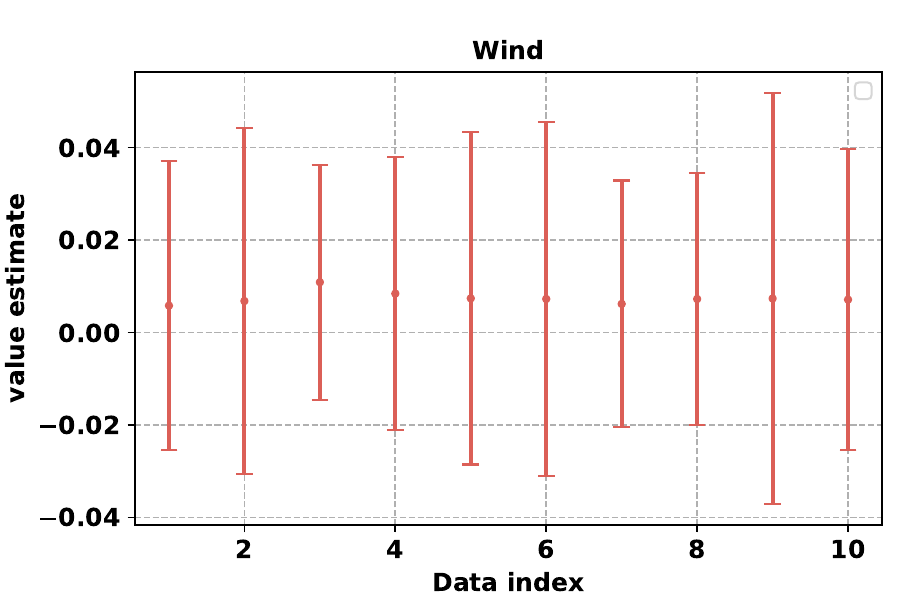}
\end{minipage}
}
\subfloat{
\begin{minipage}[b]{.30\linewidth}
\centering
\includegraphics[width=\linewidth]{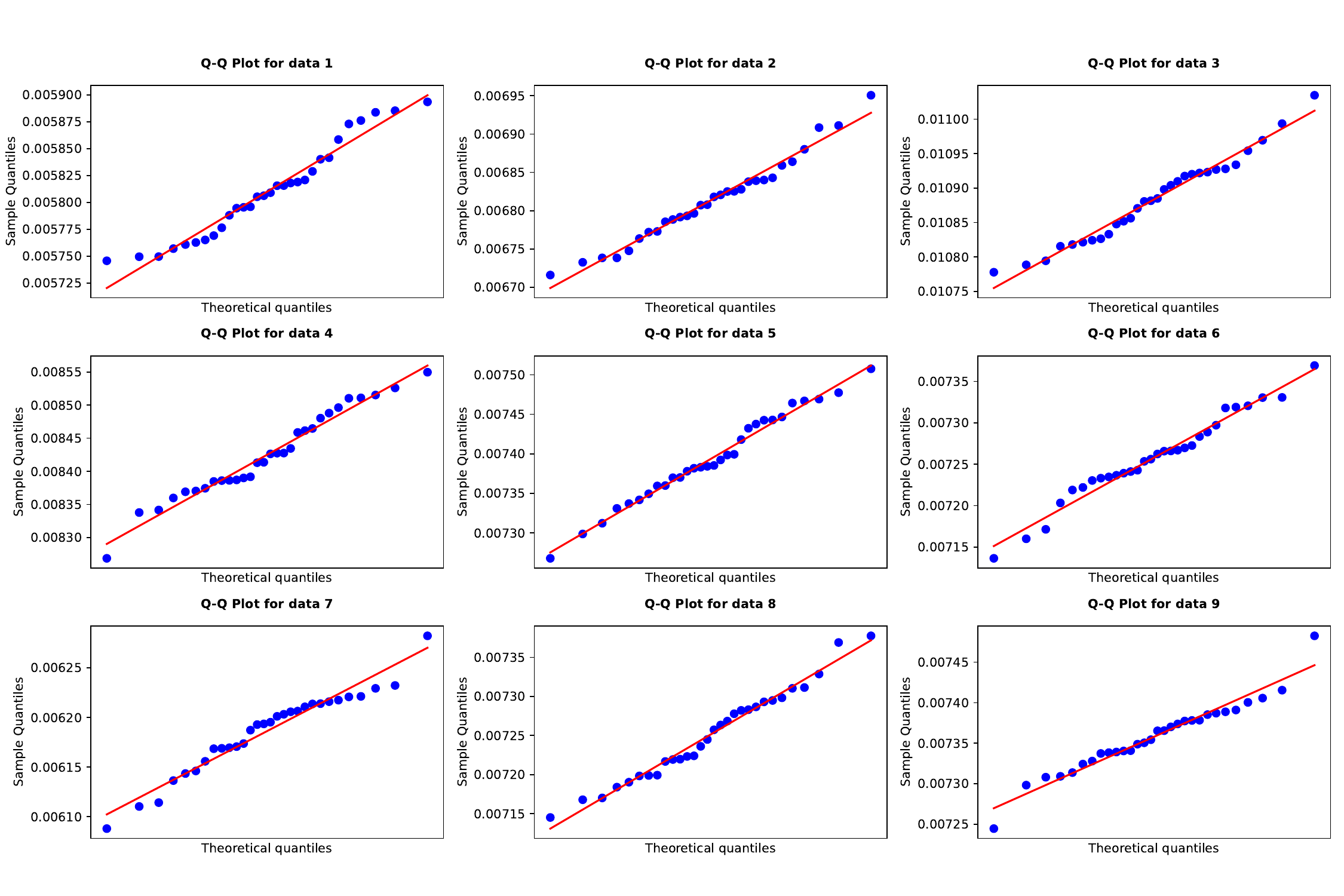}
\end{minipage}
}
\subfloat{
\begin{minipage}[b]{.32\linewidth}
\centering
\includegraphics[width=\linewidth]{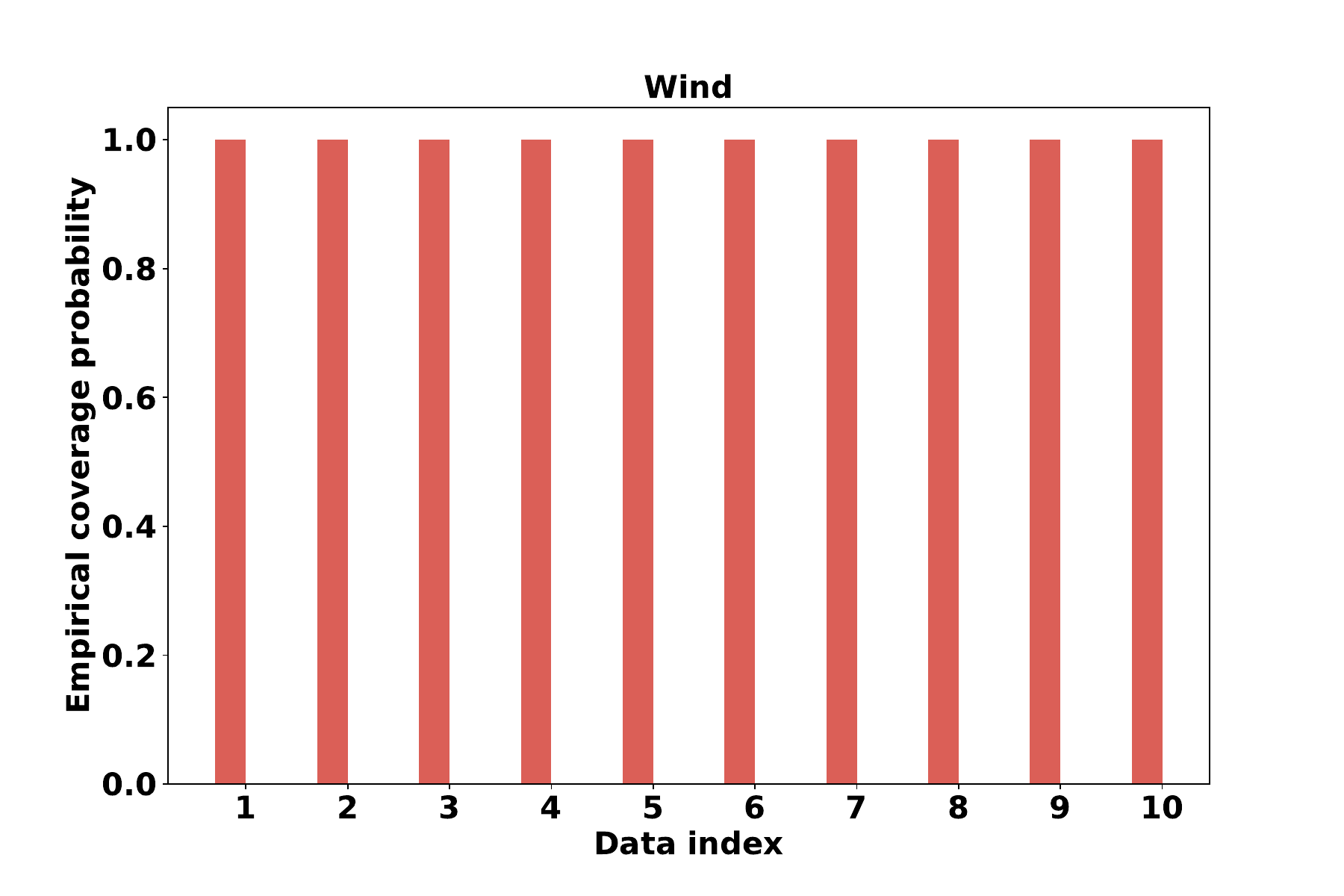}
\end{minipage}
}
\caption{Continued analysis of estimated values for Phoneme, Vehicle and Wind datasets with a training set size of 100 and $\gamma=2/3$. The first column displays the confidence interval graphs for the datasets. The second column provides the Q-Q plots for the estimated values, and the third column depicts the empirical coverage probabilities, confirming almost 100\% coverage.}
\label{othersets2} 
\end{figure*}

\end{document}